\documentclass[10pt,twocolumn,letterpaper]{article}

\usepackage[dvipsnames]{xcolor}
\usepackage{moresize}
\usepackage{graphicx}
\usepackage{booktabs}
\usepackage{colortbl}
\usepackage{tabularx}
\usepackage{rotating}
\usepackage{subcaption}
\usepackage{multirow}
\usepackage{tabularray}
\usepackage{listings}
\usepackage{makecell}
\usepackage{setspace}

\usepackage{wacv}              %

\usepackage{graphicx}
\usepackage{amsmath}
\usepackage{amssymb}
\usepackage{booktabs}
\usepackage[accsupp]{axessibility} %

\usepackage[pagebackref,breaklinks,colorlinks]{hyperref}

\usepackage[capitalize]{cleveref}
\crefname{section}{Sec.}{Secs.}
\Crefname{section}{Section}{Sections}
\Crefname{table}{Table}{Tables}
\crefname{table}{Tab.}{Tabs.}

\newcommand{\datasetname}{\emph{WAFFLE}\xspace}
\newcommand{\acronym}{\emph{WAFFLE} (\emph{\textbf{W}ikipedi\textbf{A}-\textbf{F}ueled \textbf{FL}oorplan \textbf{E}nsemble})\xspace}

\newcommand{\biglogo}{\makebox[0cm][r]{\raisebox{-8pt}{\includegraphics[height=0.7cm]{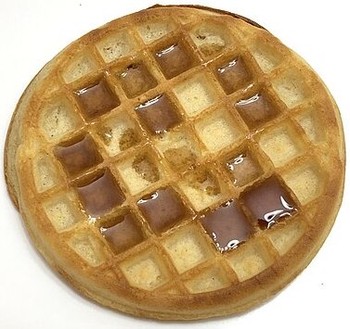}}}\xspace}
\newcommand{\smalllogo}{\includegraphics[height=0.4cm]{figures/waffle.jpeg}\xspace}

\newbox\jsavebox
\newcommand{\jsubfig}[2]{%
	\sbox\jsavebox{#1}%
	\parbox[t]{\wd\jsavebox}{\centering\usebox\jsavebox\\#2}%
	}

\newcommand{\jsubfigcent}[2]{%
	\sbox\jsavebox{#1}%
	\parbox[c]{\wd\jsavebox}{\centering\usebox\jsavebox\\#2}%
	}

\newcommand{\jsubfigright}[2]{%
  \sbox\jsavebox{#1}%
  \parbox[t]{\wd\jsavebox}{\raggedright\usebox\jsavebox\\#2}%
}
\newcommand{\whitetxt}[1]{{\color{white}#1}\normalfont}

\newcommand{\modified}[1]{{\textcolor{black}{#1}}}

\def\wacvPaperID{242} %
\def\confName{WACV}
\def\confYear{2025}

\title{\biglogo \datasetname{}: Multimodal Floorplan Understanding in the Wild}

\date{}
\author{{\centering Keren Ganon$^{*1}$ \: Morris Alper$^{*1}$ \: Rachel Mikulinsky$^{1}$ \: Hadar Averbuch-Elor$^{1,2}$ 
        }
\\
{\parbox{0.9\textwidth}{\centering
$^1$Tel Aviv University \quad
        $^2$Cornell University  
       }
}
\\
\\
{\parbox{\textwidth}{\centering
\small{\url{https://tau-vailab.github.io/WAFFLE}}       }
}
}
\begin{document}

\maketitle

\begin{abstract}
Buildings are a central feature of human culture and require significant work to design, build, and maintain. As such, the fundamental element defining their structure -- the \emph{floorplan} -- has increasingly become an object of computational analysis. Existing works on automatic floorplan understanding are extremely limited in scope, often focusing on a single semantic category and region (e.g. apartments from a single country). This contrasts with the wide variety of shapes and sizes of real-world buildings which reflect their diverse purposes.
In this work, we introduce \datasetname{}, a novel multimodal floorplan understanding dataset of nearly 20K floorplan images and metadata curated from Internet data spanning diverse building types, locations, and data formats. By using a large language model and multimodal foundation models, we curate and extract semantic information from these images and their accompanying noisy metadata.
We show that \datasetname{} serves as a challenging benchmark for prior computational methods, while enabling progress on new floorplan understanding tasks.
We will publicly release \datasetname{} along with our code and trained models, providing the research community with a new foundation for learning the semantics of buildings.

\vspace{-13pt}
\end{abstract}

\def\thefootnote{*}\footnotetext{Denotes equal contribution}
\section{Introduction}

\label{sec:intro}

\begin{figure}
    \centering
    \jsubfig{\fbox{\includegraphics[width=8cm]{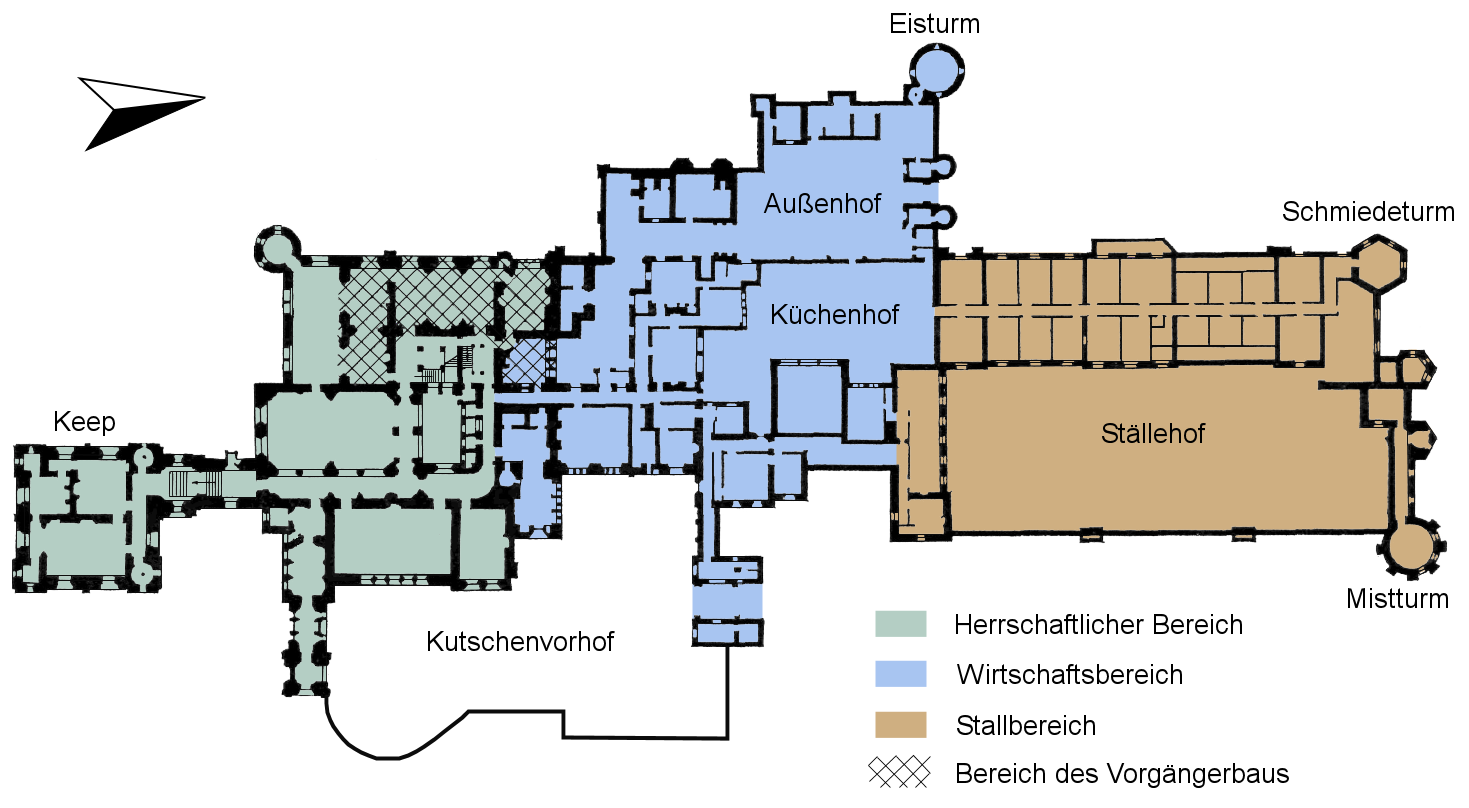}}}{}
    \hfill
    \jsubfig{\fbox{\includegraphics[height=3.1cm]{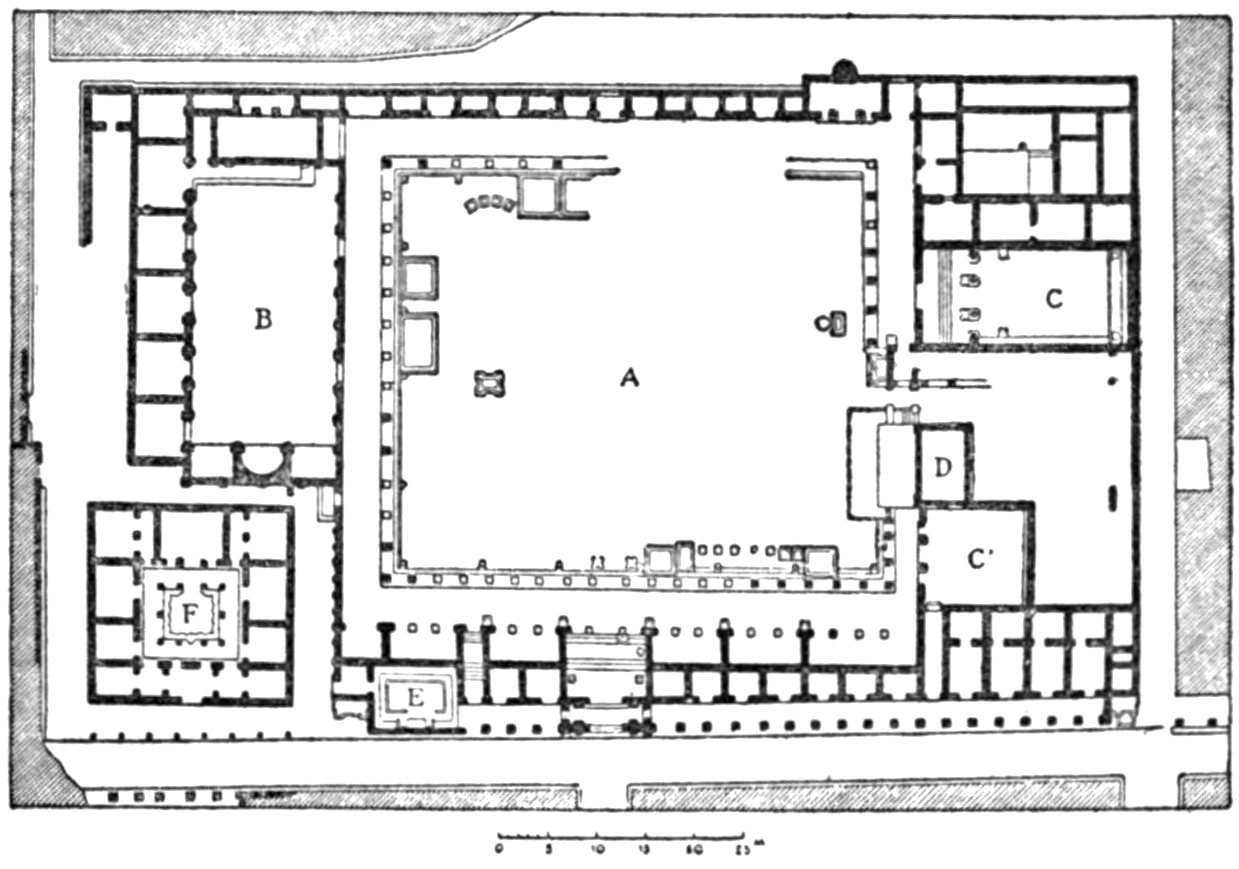}}}{}   
    \hfill
    \jsubfig{\fbox{\includegraphics[height=3.1cm]{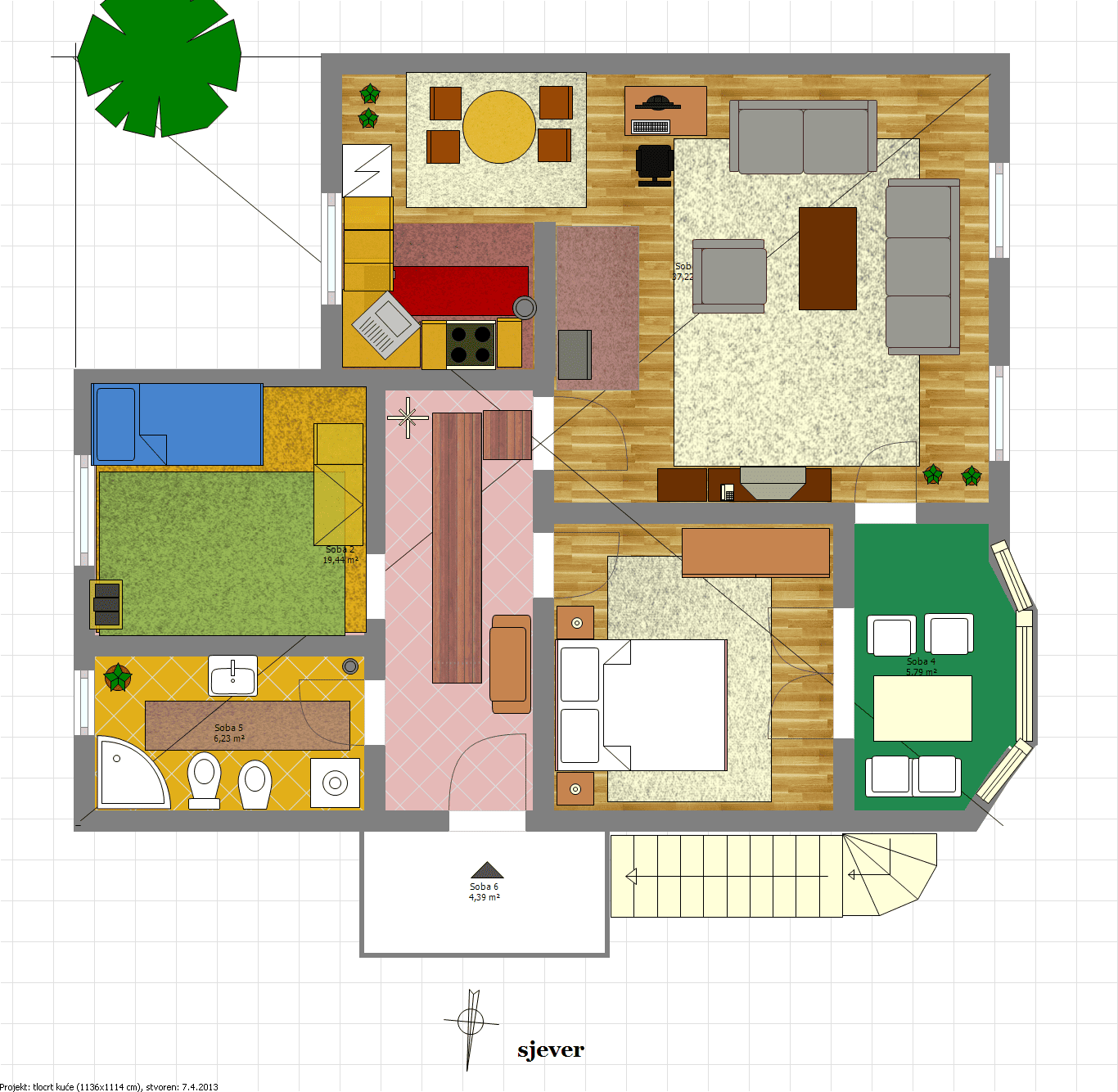}}}{}
    \caption{
   What can we understand from looking at these images? For instance, do we have a sense of what type of buildings these floorplans depict? Floorplans provide multimodal cues over the semantics and structure of buildings; however, they are often opaque for non-professionals, particularly for images lacking textual descriptions (such as the bottom images). We propose \datasetname{}, a new multimodal dataset depicting floorplan images associated with rich textual descriptions. Our dataset allows for understanding \emph{in-the-wild} floorplan imagery illustrating a wide array of building types.
    For example, a vision-and-language model finetuned on our data can correctly predict the building types for the examples depicted above (answers are provided below\setcounter{footnote}{0}\protect\footnotemark{}). 
    }
    \label{fig:teaser}
    \vspace{-15pt}
\end{figure}

\emph{``Life is chaotic, dangerous, and surprising. Buildings should reflect that."} \newline  \emph{---Frank Gehry}\\

Buildings come in all shapes and sizes, from the tiny cottages dotting the English countryside to the imposing galleries of the temple of Angkor Wat.
The diverse architectural designs of buildings have been influenced by their purposes, geographical locations, and changing trends throughout history.
Recent years have seen a growing interest in the development of computational tools for architecture, which promise to aid experts engaged in the design and maintenance of buildings. 
Of particular interest is the automatic analysis of floorplans, the most fundamental element defining the structure of buildings which communicate rich schematic and layout information.
\footnotetext{From left to right: castle, temple, residential building. These samples (depicting the \emph{Penrhyn Castle} in Wales, the \emph{Forum at Timgad} in Algeria and a house floorplan in Bosnia and Herzegovina, respectively) were taken from the \datasetname{} test set.}

Prior works have tapped into the vast visual knowledge encoded by floorplans for various applications, such as 3D reconstruction~\cite{martin20143d} and
floorplan-guided building navigation~\cite{wang2015lost,narasimhan2020seeing}. %
However, prior data-driven techniques operating on floorplans mostly focus on extremely limited semantic domains (e.g. apartments) and geographical locations (often a single country), failing to cover the diversity needed for automatic understanding of floorplans in an unconstrained setting.

In this work, we introduce \acronym{}, a multimodal floorplan understanding dataset comprised of diverse imagery spanning a variety of building types, geographical regions, historical eras, and data formats (as illustrated in Figure \ref{fig:teaser}), along with comprehensive textual data. \datasetname{} is derived from freely-available Internet images and metadata from the Wikimedia Commons platform.
To turn noisy Internet data into this curated dataset with rich semantic annotations, we leverage state-of-the-art foundation models, using large language models (LLMs) and vision-language models (VLMs) to perform curation tasks with little or no supervision. This includes %
a decomposition of floorplans into visual elements; and structuring textual metadata, code and OCR detections with LLMs.
By combining these powerful tools, we build a new dataset for floorplan understanding with rich and diverse semantics.

In addition to serving as a challenging benchmark for prior work, we show the utility of this data for various building understanding tasks that were not feasible with previous datasets. By using high-level and localized semantic labels along with floorplan images in \datasetname{}, we learn to predict building semantics and use them to generate floorplan images with the correct building type, along with optional conditioning on structural configurations. Grounded labels within images also provide supervision to segment areas corresponding to domain-specific architectural terms.
As shown by these applications, \datasetname{} opens the door for semantic understanding and generation of buildings in a diverse, real-world setting.

\section{Related Works}

\label{sec:rw}

\begin{table}[t]
\centering
\def\arraystretch{1.2}
\setlength{\tabcolsep}{1.8pt}
\begin{tabularx}{0.99\linewidth}{p{1.7cm}XXXXXXXX}
& {\small\rotatebox[origin=l]{60}{Rent3D++}} & {\small\rotatebox[origin=l]{60}{CubiCasa5K}} & {\small\rotatebox[origin=l]{60}{RPLAN}} & {\small\rotatebox[origin=l]{60}{SD}} & {\small\rotatebox[origin=l]{60}{R2V}} &  {\small\rotatebox[origin=l]{60}{FloorPlanCAD}} & {\small\rotatebox[origin=l]
{60}{\textbf{\datasetname}}} \\
\midrule
\small Building types &
\multicolumn{5}{c}{\cellcolor{Gray!25}{--- \small Residential buildings ---}} & {\small  ${\sim}4^{\diamond}$} &
\small \textbf{${>}1K$} \\
\small Countries & \small UK & \small Fin. & \small Asia & \small Switz. & \small Japan & \small ?$^{\dag}$ & \small \textbf{${>}100$} \\
\small \#Categories$^*$ & ${\sim}$\small 15 & ${\sim}$\small 80 & ${\sim}$\small 13 & \small 91 & \small 22 & \small 35 & \small \textbf{$\infty^{\ddag}$} \\
\small Real images? & $\checkmark$ & $\times$ & $\times$ & $\checkmark$ & \checkmark & \checkmark & \textbf{$\checkmark$} \\
\bottomrule
\end{tabularx}
\vspace{-5pt}
{\begin{flushleft}
    \scriptsize $^*$Number of unique annotation values for labeled grounded regions or objects.
\end{flushleft}}
\vspace{-17pt}
{\begin{flushleft}
    \scriptsize $^\dag$Unspecified data source
\end{flushleft}}
\vspace{-17pt}
{\begin{flushleft}
    \scriptsize $^\ddag$Free text, on a subset of images
\end{flushleft}}
\vspace{-17pt}
{\begin{flushleft}
    \scriptsize $^\diamond$Contains floorplans of 100 buildings spanning residential buildings, schools, hospitals, and shopping malls
\end{flushleft}}
\vspace{-10pt}
\caption{A comparison between \datasetname{} and other floorplan datasets. SD above stands for Swiss Dwellings. We can see that, in contrast to our proposed \datasetname{} dataset, most existing datasets focus on a single building type in a specific area in the world, and consider a small, closed list of annotation values. %
}
\vspace{-10pt}
\label{tab:comparison}
\end{table}

\noindent \textbf{Floorplans in Computer Vision}.
Floorplans are a fundamental element of architectural design; as such, automatic understanding and generation of floorplans has drawn significant interest from the research community.

Several works aim to reconstruct floorplans, either from 3D scans~\cite{liu2018floornet,yue2023connecting}, RGB panoramas~\cite{cabral2014piecewise,yang2019dula,shabani2021extreme}, room layouts~\cite{hosseini2023puzzlefusion} or combined modalities, such as sparse views and room-connectivity graphs~\cite{gueze2023floor}. 
Prior works also investigate the problem of alignment between floorplans and 3D point clouds depicting scenes~\cite{kaminsky2009alignment}. Martin \etal~\cite{martin20143d} leverage floorplans of large-scale scenes to produce a unified reconstruction from disconnected 3D point clouds. 
Floorplans have also been utilized for navigation tasks. 
Several works predict position over a given floorplan, for a single image~\cite{wang2015lost} or video sequences~\cite{chu2015you} depicting regions of the environment. Narasimhan \etal~\cite{narasimhan2020seeing} train an agent to navigate in new environments by predicting corresponding labeled floorplans.

Some works specifically target recognition of semantic elements over both rasterized~\cite{dodge2017parsing,zeng2019deep} and vectorized~\cite{yang2023vectorfloorseg} floorplan representations, as well as applying this to perform raster-to-vector conversion~\cite{liu2017raster, kalervo2019cubicasa5k, lv2021residential}. In our work, we are interested in understanding Internet imagery of diverse data types such as raster graphics and photographs or scans of real floorplans. In contrast to prior work that mostly focuses on a fixed set of semantic elements in residential apartments, such as walls, bathrooms, closets, and so on, we are interested in acquiring higher-level reasoning over a wide array of building types. %

The problem of synthesizing novel floorplans, and other types of 2D layouts such as documents~\cite{zheng2019content,patil2020read}, has also received considerable interest (see the recent survey by Weber \etal~\cite{weber2022automated} for a comprehensive review). Earlier works generate floorplans from high-level constraints, such as room adjacencies~\cite{hua2016irregular,merrell2010computer}. Later works are able to generate novel floorplans in more challenging settings, \eg only given their boundaries~\cite{wu2019data,hu2020graph2plan}. In our work, we show that SOTA text-to-image generation tools can be fine-tuned for generating floorplans of diverse building types, not only residential buildings, as explored by prior methods. %

\medskip 
\noindent \textbf{Floorplan Datasets}.
Prior datasets containing floorplan data are limited in structural and semantic diversity, typically being limited to residential building types such as apartments from specific geographic locations, often mined from real estate listings.
For example, Rent3D++~\cite{vidanapathirana2021plan2scene} contains floorplans of 215 apartments located in London, and CubiCasa5K~\cite{kalervo2019cubicasa5k} contains floorplans of 5K Finnish apartments.
The RPLAN~\cite{wu2019data} dataset contains 80K floorplans of apartments in Asia, further limited by various size and structural requirements (e.g., having only 3--9 rooms with specific proportions relative to the living room).
The Swiss Dwellings~\cite{standfestswiss} includes floorplan data for 42K Swiss apartments, and the Modified Swiss Dwellings~\cite{msd} dataset provides a filtered subset of this data with additional access graph information for floorplan auto-completion learning.
The R2V~\cite{liu2017raster} dataset introduces 815 Japanese residential building floorplans.

Additionally, \datasetname{} differs substantially from prior works with regards to the sourcing and curation of data. Datasets of real floorplans, such as those previously-mentioned, are constructed with tedious manual annotation. For example, specialists spent over 1K hours in the construction of FloorPlanCAD~\cite{Fan_2021_ICCV} to provide annotations of 30 categories (such as door, window, bed, etc.).  
Annotations may also derive from other input types rather than being direct annotations of floorplans; for instance, the Zillow Indoor Dataset~\cite{cruz2021zillow} generates floorplans with user assistance from 360$^\circ$ panoramas, yielding plans for 1,524 homes after over 1.5K hours of manual annotation labor. To bypass such manual procedures, other works
generate synthetic floorplans using predefined constraints~\cite{delalandre2010generation}. 
By contrast, \datasetname{} contains diverse Internet imagery of floorplans, including both original digital images and scans captured in the wild, and is curated with a fully automatic pipeline. See Table \ref{tab:comparison} for a comparison of the most related datasets with our proposed \datasetname{} dataset.

Finally, there are also large-scale datasets of landmark-centric image collections, such as Google Landmarks~\cite{noh2017large,weyand2020google} and WikiScenes~\cite{wu2021towers}. Along with photographs and similar imagery of these landmarks, such collections may include schematic data such as floorplans. While prior works focus on the natural imagery in these collections for tasks such as image recognition, retrieval, and 3D reconstruction, we specifically leverage the schematic diagrams found in such collections for layout generation and understanding.

\section{ \smalllogo \datasetname{}: Internet Floorplans Dataset}

\begin{figure}
\jsubfigright{\includegraphics[height=3.12cm]{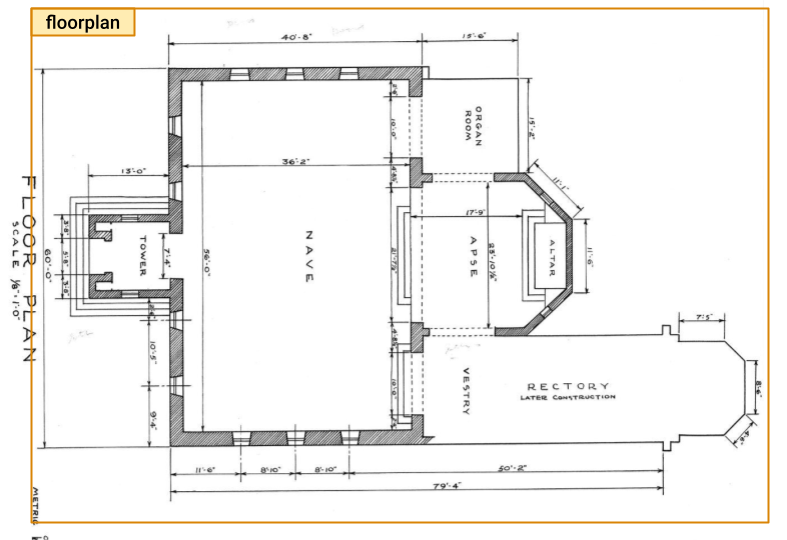}}{
\ssmall \textbf{Building name}: St. Paul's Episcopal Church \\
\textbf{Building type}: church \\
\textbf{Country}: United States of America \\
\textbf{Grounded architectural features}: \{organ room rector, chapel, vestry, altar, tower, nave, apse\}}
\hfill
\jsubfigright{
\includegraphics[height=3.12cm]{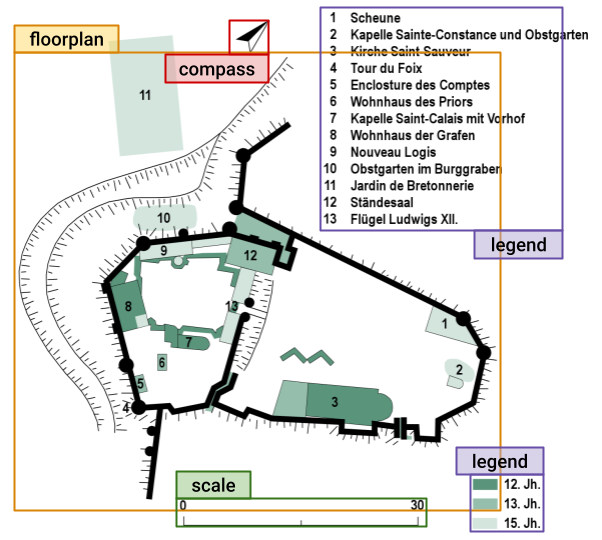}}{\ssmall
\textbf{Building name}: Château de Blois \\
\textbf{Building type}: castle \\
\textbf{Country}: France \\
\textbf{Grounded legend}: \{1: barn, 2:chapel, 3: saint savior church, 6: prior's house, 8: residence of counts, 9: new home, 10: orchard in moat, 11: bretonry garden, 12: booth hall\}
}
\vspace{-5pt}
\caption{
  \textbf{Samples from \datasetname{}}. Above, we show images paired with their structured data, including the building name and type, country of origin, and their grounded architectural features. We also visualize the detected layout components (floorplan, legend, compass, and scale, as relevant) overlaid on top of the images.}
\label{fig:samples_v2}
\end{figure}

In this section, we introduce \acronym{}, a new dataset of 18,556 floorplans, derived from Wikimedia Commons\footnote{\url{https://commons.wikimedia.org}} and associated textual descriptions available on Wikipedia. \datasetname{} contains floorplan images with paired structured metadata containing overall semantic information and spatially-grounded legends. %
Samples from our dataset are provided in Figure \ref{fig:samples_v2}. We provide an interactive viewer of samples from the \datasetname{} dataset, and additional details and statistics of our dataset, in the supplementary material. 
We proceed to describe the curation process and contents of \datasetname{}.

\subsection{Data Collection} \label{sec:collection}

Images and metadata in Wikimedia Commons data are ordered by hierarchical categories (\emph{WikiCategories}). To find relevant data, we recursively scrape the WikiCategories \texttt{Floor plans} and \texttt{Architectural drawings}, extracting images and metadata from Wikimedia Commons and the text of linked Wikipedia articles. As many images contain valuable textual information (e.g. hints to the location of origin, legend labels, etc.), we also extract text from the images using the Google Vision API\footnote{\url{https://cloud.google.com/vision?hl=en}} for optical character recognition (OCR). Finally, we decompose images into constituent items by fine-tuning the detection model DETR~\cite{carion2020end} on a small subset of labeled examples to predict bounding boxes for common layout components (floorplans, legend boxes, compass, and scale icons). %

The raw data includes a significant amount of noise along with floorplans, including similar topics such as maps and cross-sectional blueprints as well as other unrelated data. %
Therefore, we filter this data as follows:

\medskip
\noindent
\textbf{Text-based filtering (LLM).}
We perform an initial text-only filtering stage by processing our images' textual metadata with an LLM to extract structured information. We provide the LLM with a prompt containing image metadata and ask it to categorize the image in multiple-choice format, providing it with a closed set of possible categories. These include positive categories such as \emph{floorplan} and \emph{building} as well as some negative categories (not floorplans) such as \emph{map} and \emph{city}.

\medskip
\noindent
\textbf{Image-based filtering (CLIP).} We use CLIP~\cite{radford2021learning} image embeddings to filter for images likely to be floorplans. Firstly, as the WikiCategory \texttt{Architectural drawings} contains many non-floorplan images, we train a linear classifier on a balanced sample of items from the two WikiCategories and select images that are closer to those in the \texttt{Floor plans} WikiCategory. Moreover, we filter all images by comparing them with CLIP text prompt embeddings, following the use of CLIP for zero-shot classification. We compare to multiple prompts such as \emph{A map}, \emph{A picture of people}, and \emph{A floorplan}, aggregating scores for positive and negative classes and filtering out images with low scores. Finally, we train a binary classifier using high-scoring images and negative examples to adjust the zero-shot CLIP classifications for increased recall.

\medskip
\noindent

This step results in a final dataset of nearly 20K images. Each image is accompanied by the following raw data extracted from its Wikimedia Commons page and linked pages: the image file name, its Wikimedia Commons page content (including a textual description), a list of linked WikiCategories, the contents of linked Wikipedia pages (if present), OCR detections in the image, and bounding boxes of constituent layout components.

\subsection{LLM-Driven Structured pGT Generation}
\label{sec:llm-section}

Our raw data contains significant grounded information about each image in diverse formats, which we wish to systematically organize and structure for use in downstream tasks. To this aim, we harness the capabilities of large language models (LLMs) for distilling essential information from diverse textual data.
In particular, we extract the following information (also illustrated in Figure \ref{fig:samples_v2}) by prompting Llama-2 \cite{touvron2023llama} with an instruction and relevant metadata fields: building name, building type (i.e. \emph{church}, \emph{hotel}, \emph{museum} etc.), location information (country, state, city), and a list of architectural features that are grounded in the image.

In general, the raw metadata contains considerable and diverse noise, involving multilingual content and multiple written representations of identical entities (e.g. \emph{Notre Dame Cathedral} vs. \emph{Notre-Dame de Paris}). To control for the source language, we employ prompts that instruct the LLM to respond in English and request translations when necessary.
For linking representations of identical entities (also known as \emph{record linkage}), we employ LinkTransformer~\cite{arora2023linktransformer} clustering along with various textual heuristics. %
We provide additional details, including prompts used, in the supplementary material, and proceed to describe our method for grounding architectural features in floorplans.

\begin{figure}
  \centering
\jsubfig{{\includegraphics[height=2cm]{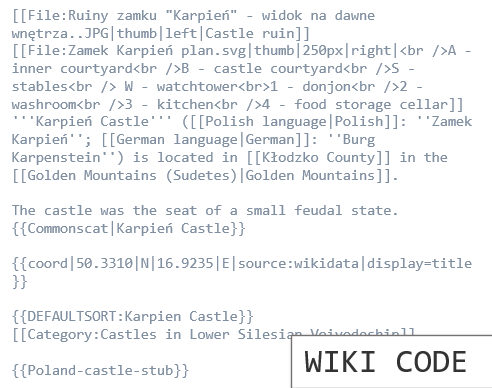}}}{Raw Data}
\hfill
\jsubfig{{\includegraphics[height=2cm]{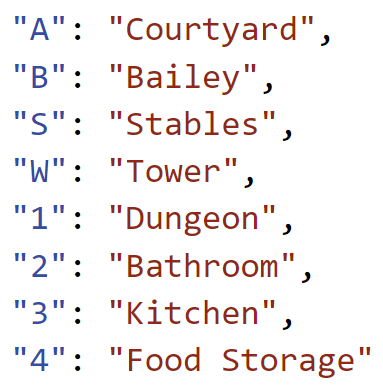}}}{Legend}
\hfill
\jsubfig{{\includegraphics[height=2cm]{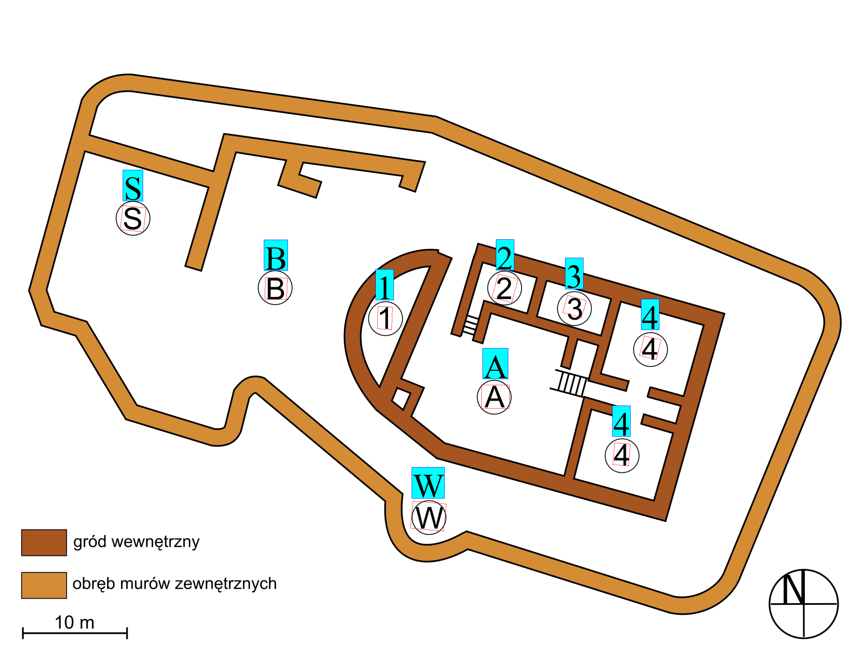}}}{Image}
\vspace{-5pt}
\caption{We automatically extract legends and architectural features from the image raw data (illustrated on the left, either the image metadata or OCR detections) by prompting LLMs. We associate the keys with text detected in the image, yielding grounded regions associated with semantics.}
  \label{fig:legend_grounding}
\vspace{-10pt}
\end{figure}

\medskip
\noindent \textbf{Architectural Feature Extraction and Grounding.}
Many floorplan images indicate architectural information either directly with text on the relevant region, or indirectly using a legend.
To identify legends and architectural information marked directly on the floorplan, we examine the bounding boxes of floorplan and legend detections (using the model described in Section \ref{sec:collection}) and select OCR detections within these areas. We also extract additional legend information from image metadata by prompting the LLM with an instruction including page content from the image's Wikimedia Commons page or the code surrounding the image in its linked Wikipedia pages (as legends often appear in these locations). 
We further structure the legend outputs using regular expressions to identify key-value pairs.
Finally, we link the legend keys and architectural features to the regions in the floorplan images coinciding with OCR detections, thus providing grounding for the semantic values of the image. See Figure \ref{fig:legend_grounding} for an example.

\subsection{Dataset Statistics} \label{sec:stats}
Our dataset contains nearly 20K images with accompanying metadata, in a range of formats. \modified{In particular, we note that our dataset contains over 1K vectorized floorplans}. Additionally, our dataset contains more than 1K building types spread over more than 100 countries across the world, and over 11K different Grounded Architectural Features (GAFs) across almost 3K grounded images. We split into train and test sets (18,259 and 297 images respectively) by selecting according to country (train: 50 countries; test: 57 countries), thus ensuring disjointedness with regards to buildings and preventing data leakage.

\medskip
\noindent \textbf{\modified{Data Quality Validation.}}
We manually inspect the test set images, removing images that do not contain a valid floorplan. Based on this validation, we find that 89\% are indeed relevant floorplan images. We find this level of noise acceptable for training models on in-the-wild data, while the manual filtering assures a clean test set for evaluation. \modified{In addition, we manually inspect the quality of our generated pGTs. We find that 89\% of the building names, 85\% of the building types and 96\% of the countries of origin are accurately labeled (considering 100 random data samples).}%

\section{Experiments}

In this section, we perform several experiments applying our dataset to both discriminative and generative building understanding tasks. For all tasks, we use the the train-test split outlined in Section \ref{sec:stats}. Please refer to the supplementary material for further training details. 

\begin{table}[t]
  \centering
  \setlength{\tabcolsep}{5pt}
  \begin{tabular}{lcccccc}
    & R@1 & R@5 & R@8 & R@16  & MRR \\
    \toprule
    CLIP & 1.5\% &  7.6\% & 10.3\% & 19.7\% & 0.07\\
    CLIP$_{FT}$ & \textbf{11.8\%} & \textbf{34.1\%} & \textbf{40.0\%} & \textbf{52.9\%} &   \textbf{0.23}\\
    \bottomrule
  \end{tabular}
  \vspace{-7pt}
  \caption{Results on CLIP retrieval of building types, for CLIP before and after fine-tuning on our dataset. We report Recall@k (R@k) for $k \in \{1, 5, 8, 16\}$ and Mean Reciprocal Rank (MRR) for these models, evaluated on our test set. As seen above, fine-tuning on \datasetname{} significantly improves retrieval metrics.}
\label{tab:clip}
\end{table}

\subsection{Building Type Understanding}\label{sec:FTCLIP}
\noindent \textbf{Task description.} We test the ability to predict building-level semantics from a floorplan, similarly to a human who might look at a floorplan and make an educated guess as to what type of building it depicts. To learn this understanding, we fine-tune CLIP with a contrastive objective on paired images and building type pseudo-labels from \datasetname{}. Our fine-tuned model (CLIP$_{FT}$) is expected to adjust CLIP to assign floorplan image embeddings close to those of relevant building types, allowing for subsequent retrieval or classification with floorplan images as input. We test the extent to which this understanding has been learned in practice with standard retrieval metrics, evaluating Recall@k for $k \in \{1, 5, 8, 16\}$ and Mean Reciprocal Rank (MRR).

\medskip \noindent \textbf{Results.} Results for fine-tuning CLIP for building type understanding are shown in Table \ref{tab:clip}. As is seen there, CLIP$_{FT}$ significantly outperforms the base model in retrieving the correct building type pseudo-labels, hence showing a better understanding of their global semantics.

\begin{figure}
    \centering
\rotatebox{90}{\hspace{-8pt}{\footnotesize Nave}}
\jsubfigcent{{\includegraphics[width=1.81cm]{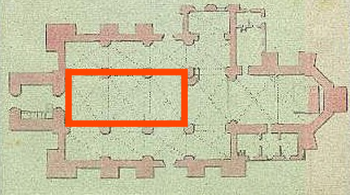}}}{}
\hfill
\hspace{1.81cm}
\hfill
\jsubfigcent{{\includegraphics[width=1.81cm]{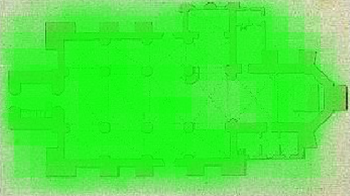}}}{ }
\hfill
\jsubfigcent{{\includegraphics[width=1.81cm]{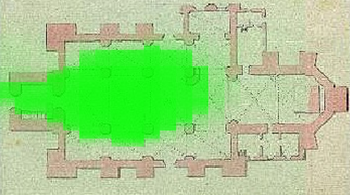}}}{ }
    \\
        \rotatebox{90}{\hspace{-7pt}{\footnotesize Court}}
\jsubfigcent{{\includegraphics[width=1.81cm]{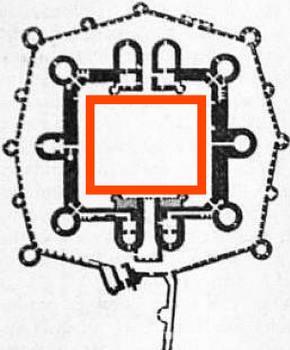}}}{}
\hfill
\hspace{1.81cm}
\hfill
\jsubfigcent{{\includegraphics[width=1.81cm]{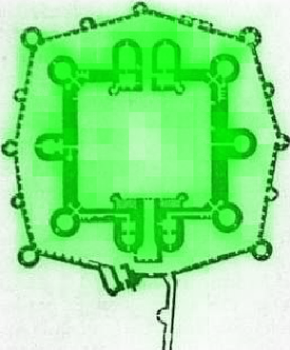}}}{}
\hfill
\jsubfigcent{{\includegraphics[width=1.81cm]{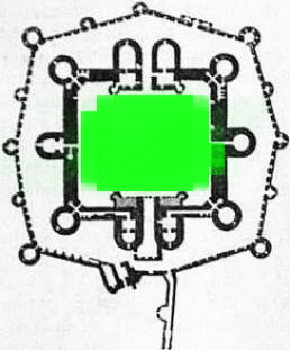}}}{} 
\\
\rotatebox{90}{\hspace{-6pt}{\footnotesize Kitchen}}
\jsubfigcent{{\includegraphics[width=1.81cm]{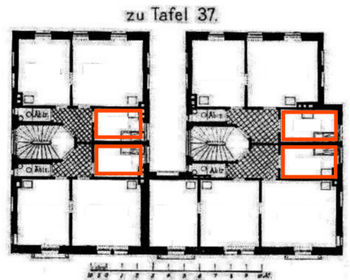}}}{GT}
\hfill
\jsubfigcent{{\includegraphics[width=1.81cm]{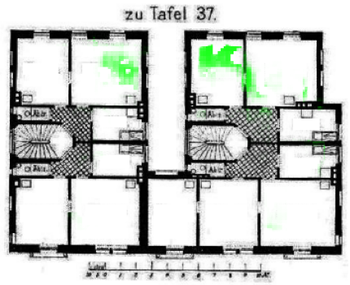}}}{CC5K$^*$}
\hfill
\jsubfigcent{{\includegraphics[width=1.81cm]{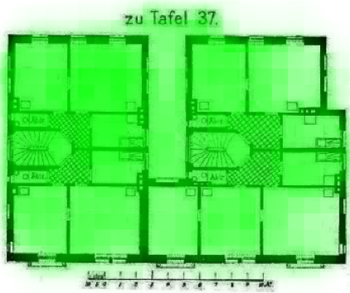}}}{CLIPSeg}
\hfill
\jsubfigcent{{\includegraphics[width=1.81cm]{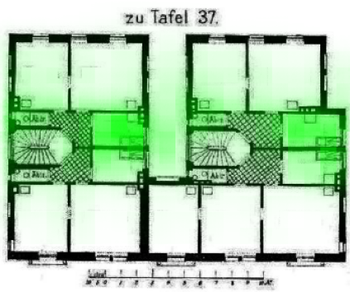}}}{Ours}
\\
 \vspace{-5pt}
\caption{Comparison of open-vocabulary segmentation probability map results. We show the input images in the first column, with the corresponding GT regions in red. $^*$Note that CC5K is a closed-vocabulary model designed for residential floorplan understanding, and therefore we cannot compare to it over additional building types (such as castles and cathedrals illustrated above). 
In addition to improving on the base CLIPSeg segmentation model, we outperform the strongly-supervised CC5K, suggesting that this model cannot generalize well beyond its training set distribution.
}
\vspace{-12pt}
    \label{fig:segmentation_comparison}
\end{figure}

\begin{table}[t]
  \centering
  \begin{tabular}{lcccc}
  \setlength{\tabcolsep}{5pt}
    & CC5K$^*$ & CLIPSeg & Ours \\
    \toprule
    AP & 0.138 & {0.157}  & {\textbf{0.226}} \\
    \modified{mIoU} & \modified{0.057} & \modified{0.066}  & \modified{\textbf{0.131}} \\
    \bottomrule
  \end{tabular}
 \vspace{-7pt}
  \captionof{table}{Open-Vocabulary Floorplan Segmentation Evaluation. We compare against a pretrained CLIPSeg model and against a closed-vocabulary segmentation model (CC5K). As illustrated above, our method improves localization across all evaluation metrics. $^*$Evaluated only over a subset of residential buildings.}
  \label{tab:segementation_metrics}
  \end{table}

\subsection{Open-Vocabulary Floorplan Segmentation}
\noindent \textbf{Task description.} To model localized semantics within floorplans, we use the GAFs in \datasetname{} to fine-tune a text-driven segmentation model. We adopt the open-vocabulary text-guided segmentation model CLIPSeg~\cite{luddecke2022image} and perform fine-tuning on the subset of these grounded images.

To provide supervision, we use the values of the GAFs as input text prompts for the segmentation model and the OCR bounding box regions of the associated grounded values as segmentation targets. This yields partial ground truth supervision; for a text query, we use OCR bounding box regions corresponding to text labels that semantically match the query (implemented via text embedding similarity) as positive targets and the remaining bounding box regions as negative targets. To prevent leakage from the written text in the images, we perform inpainting with Stable Diffusion~\cite{rombach2022highresolution} to replace the contents of the OCR bounding boxes.
As our inpainting process may cause artifacts, for evaluation purposes we manually select images that do not contain GAFs.
\modified{We follow prior work~\cite{luddecke2022image} and report mean Intersection over Union (mIoU) and Average Precision (AP). The mIoU metric requires a threshold, which we empirically set to 0.25. AP is a threshold-agnostic metric  that measures the area under the recall-precision curve, quantifying to what extent it can discriminate between correct and erroneous matches.}
In addition to comparing against the pretrained CLIPSeg model, we compare against the closed-vocabulary segmentation model provided by CubiCasa5K (CC5K)~\cite{kalervo2019cubicasa5k} over a subset of residential buildings in our test set (evaluating semantic regions which this model was trained on).

\medskip \noindent \textbf{Results.} \modified{Quantitative results are reported in Table \ref{tab:segementation_metrics}, showing a clear boost in performance across both metrics. This is further reflected in our qualitative results in Figures \ref{fig:segmentation_comparison}. %
In addition, the results on residential buildings of the strongly-supervised residential floorplan understanding model~\cite{kalervo2019cubicasa5k} yields inferior performance, likely because the latter model uses supervision from a specific geographical region and style alone (a limitation of existing datasets, as we describe in Section \ref{sec:rw}). Overall, both metrics show that there is much room for improvements with future techniques leveraging our data for segmentation-related tasks.}

\begin{table}[t]
  \centering
  \setlength{\tabcolsep}{5pt}
  \begin{tabular}{lcccccc}
     & Walls & Doors &  Windows & Interior & BG\\
    \toprule
    Precision & 0.737 & 0.201 & 0.339 & 0.799 & 0.697\\
    Recall & 0.590 & 0.163 & 0.334 & 0.521 & 0.912\\
    IoU & 0.488 & 0.099 & 0.202 & 0.461 &0.653\\
    \bottomrule
  \end{tabular}
  \vspace{-7pt}
\captionof{table}{Benchmark for Semantic Segmentation Evaluation. We benchmark prior work, reporting performance over the  CubiCasa-5k \cite{kalervo2019cubicasa5k} segmentation model, on common grounded categories. Note that background is denoted as BG above. As illustrated, \datasetname{} serves as a challenging benchmark for existing work. 
}
\label{tab:benchmark_segmentation}
\vspace{-7pt}
\end{table}

\begin{figure*}
    \centering
    \setlength\tabcolsep{0.1pt}
    \begin{tabular}{c@{\hspace{0.1cm}}ccccccc}
        & School & Palace & Church & Castle & Hospital & Hotel & Library \\
        \rotatebox{90}{\whitetxt{xx}pretrained} &
        \includegraphics[height=2.3cm]{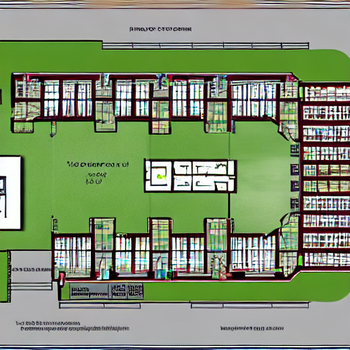} &
        \includegraphics[height=2.3cm]{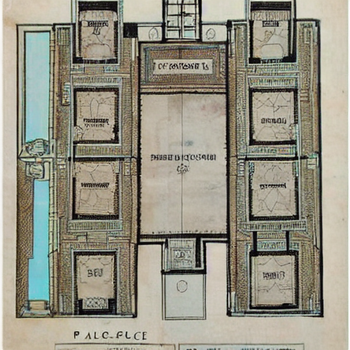} & \includegraphics[height=2.3cm]{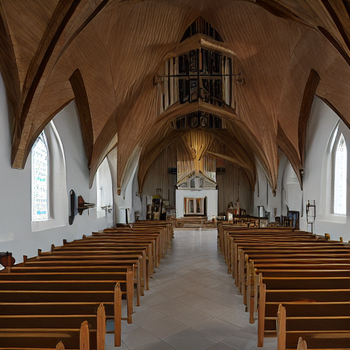} &
        \includegraphics[height=2.3cm]{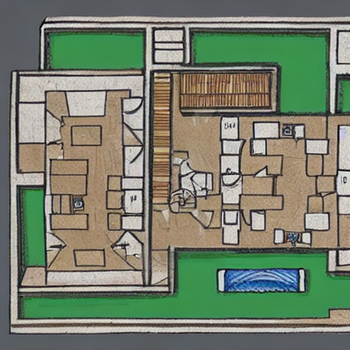} & \includegraphics[height=2.3cm]{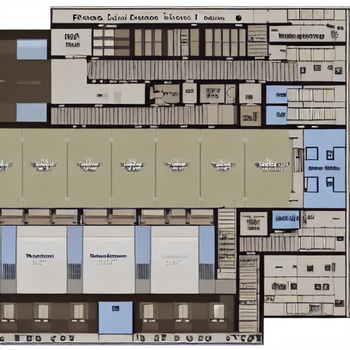} & \includegraphics[height=2.3cm]{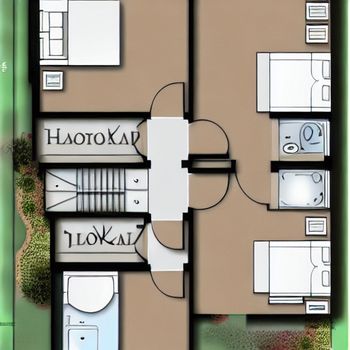} & \includegraphics[height=2.3cm]{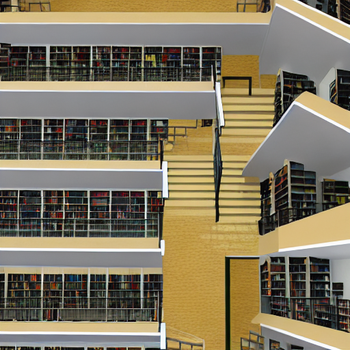} \\
        \addlinespace
        \multirow{3}{*}{\rotatebox[origin=c]{90}{$\longleftarrow$ fine-tuned $\longrightarrow$}} &
        \includegraphics[height=2.3cm]{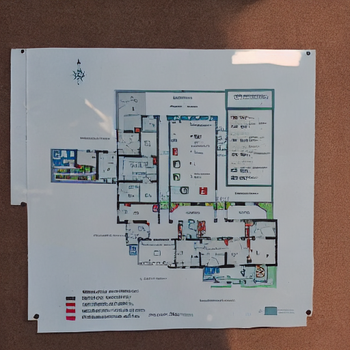} &
        \includegraphics[height=2.3cm]{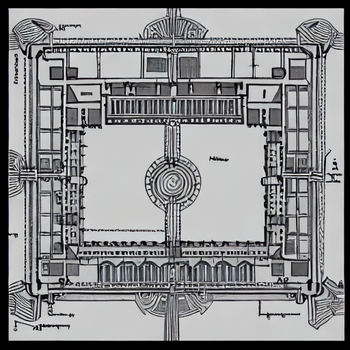} & \includegraphics[height=2.3cm]{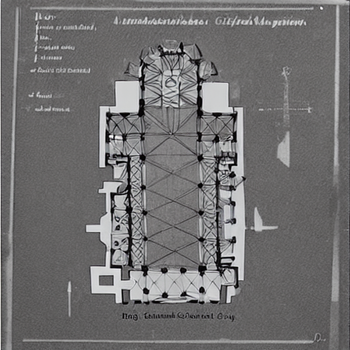} &
        \includegraphics[height=2.3cm]{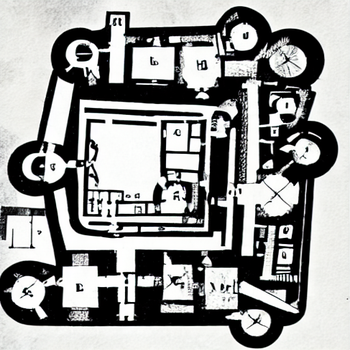} & \includegraphics[height=2.3cm]{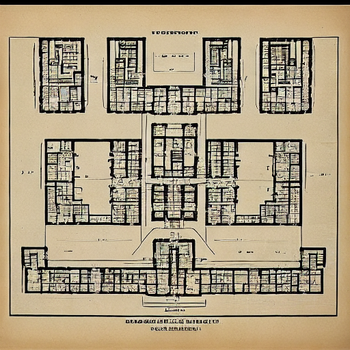} & \includegraphics[height=2.3cm]{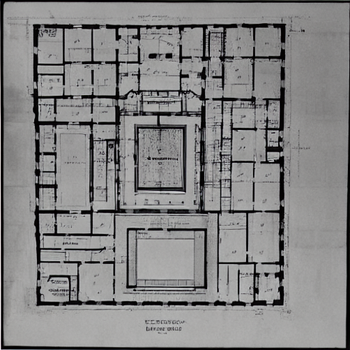} & \includegraphics[height=2.3cm]{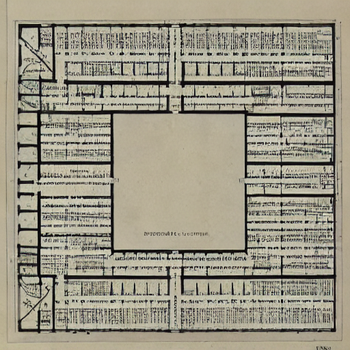} \\
        &
        \includegraphics[height=2.3cm]{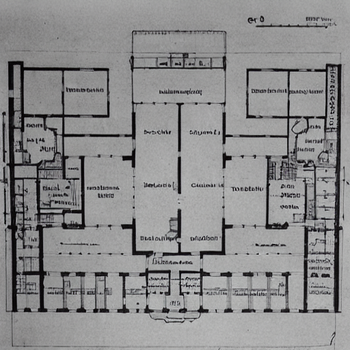} &
        \includegraphics[height=2.3cm]{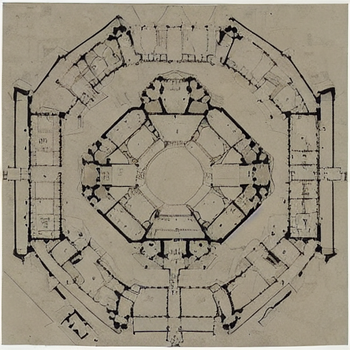} & \includegraphics[height=2.3cm]{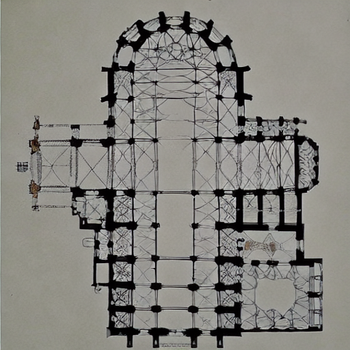} & \includegraphics[height=2.3cm]{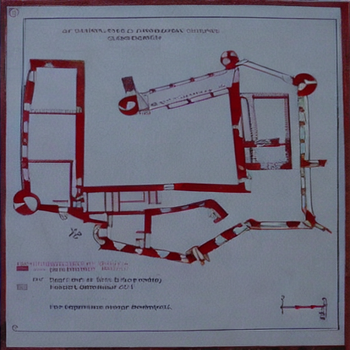} & \includegraphics[height=2.3cm]{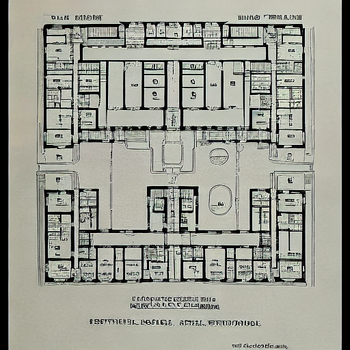} & \includegraphics[height=2.3cm]{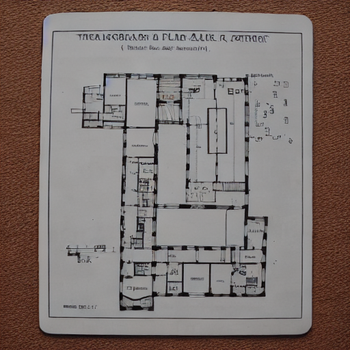} & \includegraphics[height=2.3cm]{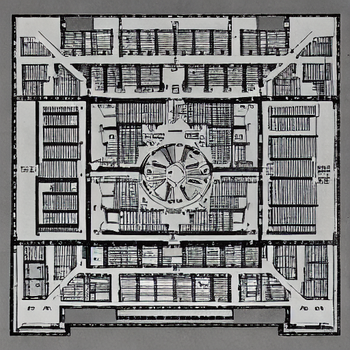}\\ %
        &
        \includegraphics[height=2.3cm]{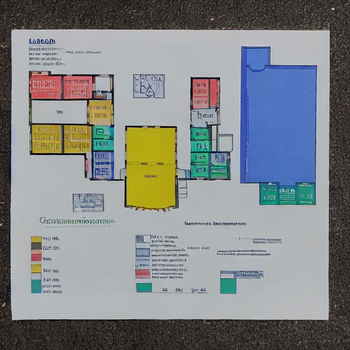} &
        \includegraphics[height=2.3cm]{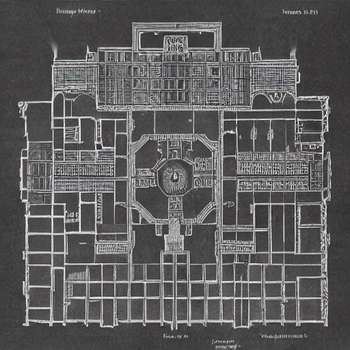} & \includegraphics[height=2.3cm]{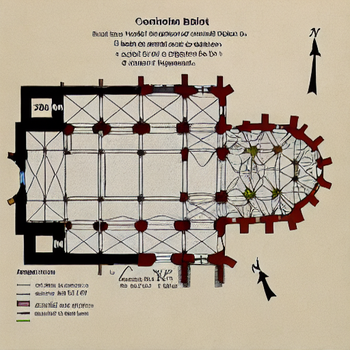} & \includegraphics[height=2.3cm]{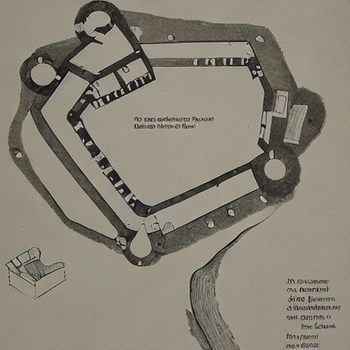} & \includegraphics[height=2.3cm]{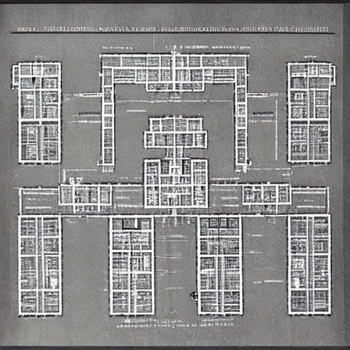} & \includegraphics[height=2.3cm]{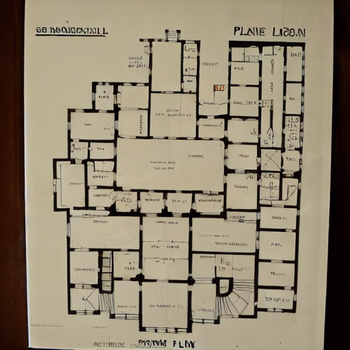} & \includegraphics[height=2.3cm]{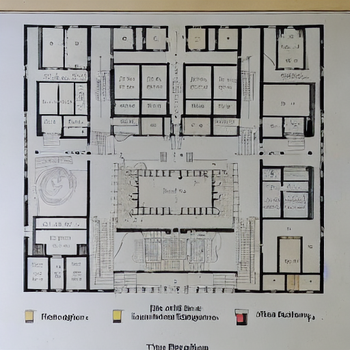} \\
     \end{tabular}
    \caption{Examples for generated floorplans for various building types, using the prompt ``A floor plan of a \texttt{<building\_type>}'' (corresponding types are shown on top). The first row shows samples from the pretrained SD model, and the bottom three show results from the model fine-tuned on \datasetname{}. As seen above, pretrained SD struggles at generating floorplans in general and often yields results that do not structurally resemble real floorplans. By contrast, our fine-tuned model can correctly generate fine-grained architectural structures, such as towers in castles or long corridors in libraries.  }
    \label{fig:generated_examples}
\end{figure*}

\begin{figure*}
    \centering
    \setlength\tabcolsep{0.1pt}
    \begin{tabular}{cc@{\hspace{5pt}}cc@{\hspace{5pt}}cccc}
        Input & & Mask & & Museum & Theater & School & Hotel \\
        \includegraphics[height=2.5cm]{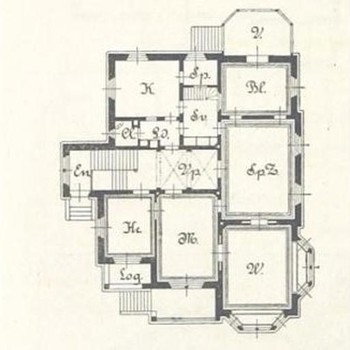} & & \includegraphics[height=2.5cm]{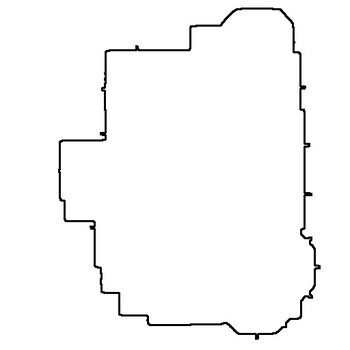} & &
        \includegraphics[height=2.5cm]{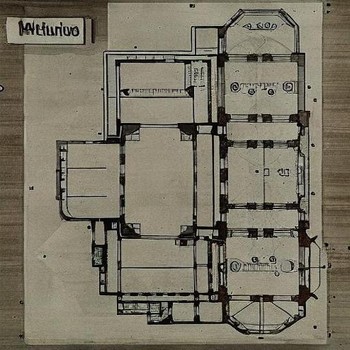} &
        \includegraphics[height=2.5cm]{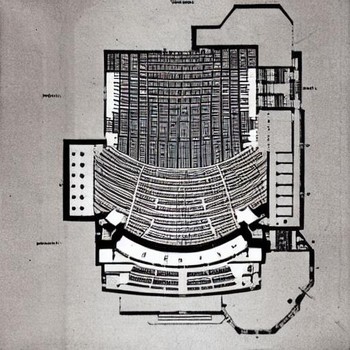} &
        \includegraphics[height=2.5cm]{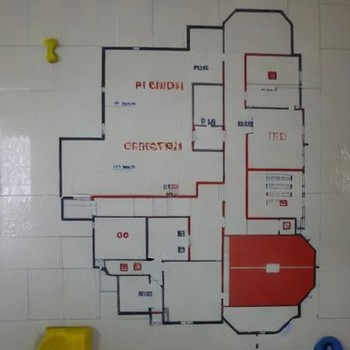} &
        \includegraphics[height=2.5cm]{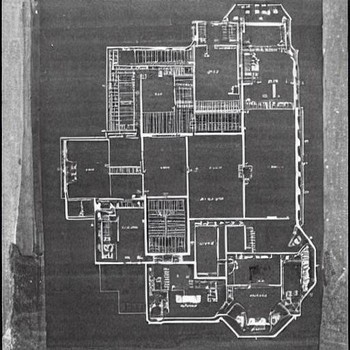} \\
        
        \includegraphics[height=2.5cm]{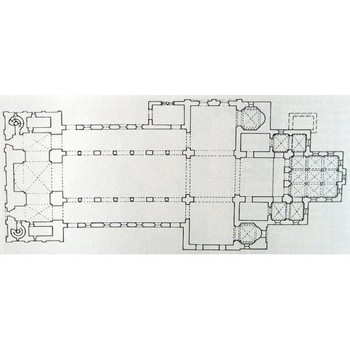} & & \includegraphics[height=2.5cm]{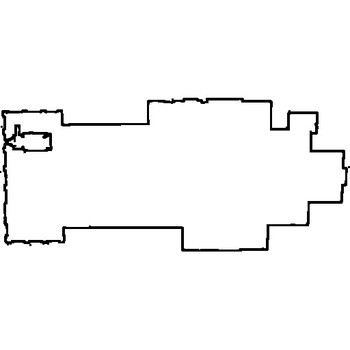} & &
        \includegraphics[height=2.5cm]{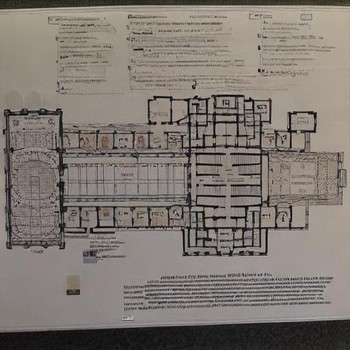} &
        \includegraphics[height=2.5cm]{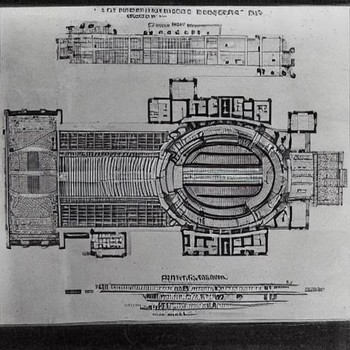} &
        \includegraphics[height=2.5cm]{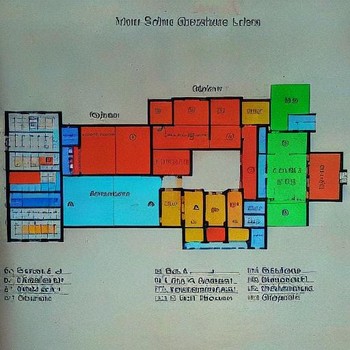} &
        \includegraphics[height=2.5cm]{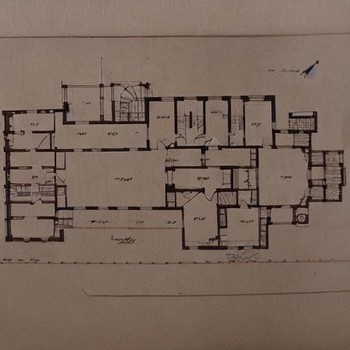} \\
        
        \includegraphics[height=2.5cm]{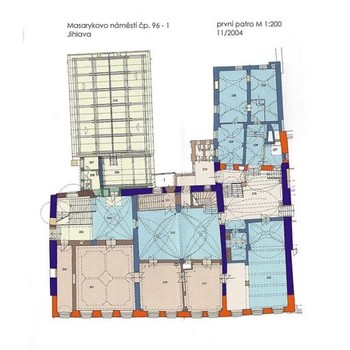} & & \includegraphics[height=2.5cm]{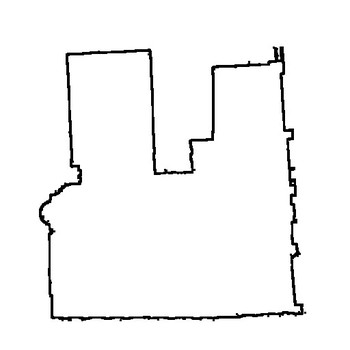} & &
        \includegraphics[height=2.5cm]{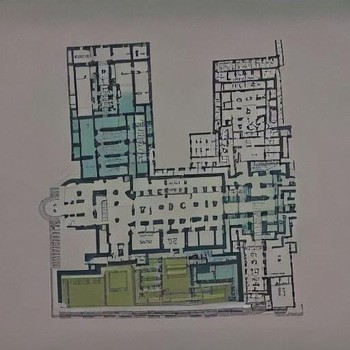} &
        \includegraphics[height=2.5cm]{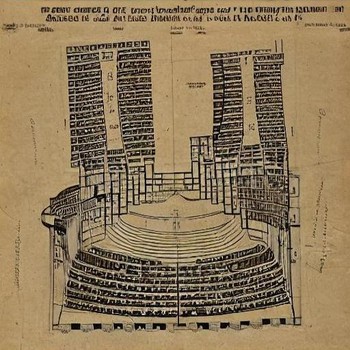} &
        \includegraphics[height=2.5cm]{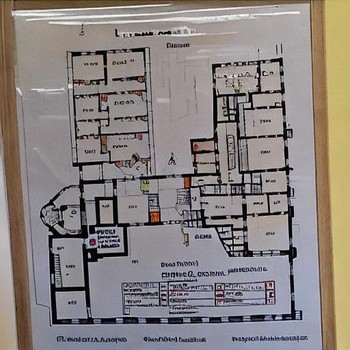} &
        \includegraphics[height=2.5cm]{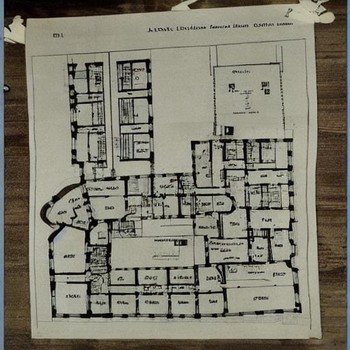}
    \end{tabular}
    \vspace{-10pt}
    \caption{Boundary-conditioned generation. The first column shows images in \datasetname{}, the second column shows automatically-extracted boundary masks, and the following columns show floorplan image generations conditioned on this generation with diverse building types provided as prompts. }
    \label{fig:controlnet_examples}
\end{figure*}

\begin{figure*}
    \centering
    \setlength\tabcolsep{0.1pt}
    \begin{tabular}{ccc@{\hspace{5pt}}ccc@{\hspace{5pt}}ccc@{\hspace{5pt}}cc}
        Cond. & School & & Cond. & Castle & & Cond. & Library & & Cond. & Cathedral \\
        \includegraphics[trim={0cm 1cm 0cm 1cm},clip,height=2cm]{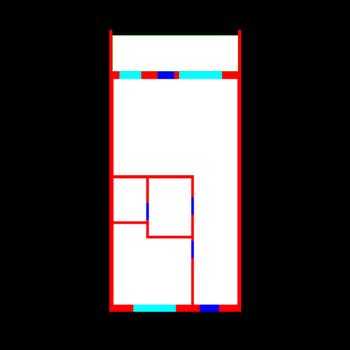} & \includegraphics[trim={0cm 1cm 0cm 1cm},clip,height=2cm]{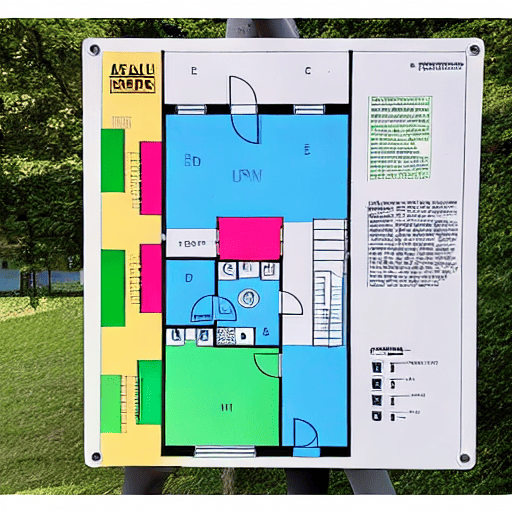} & &
        \includegraphics[height=2cm]{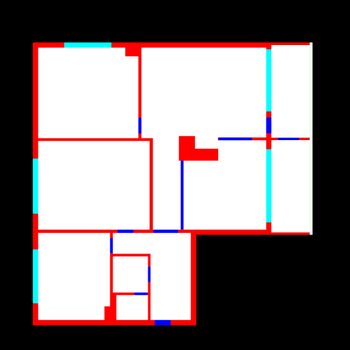} & \includegraphics[height=2cm]{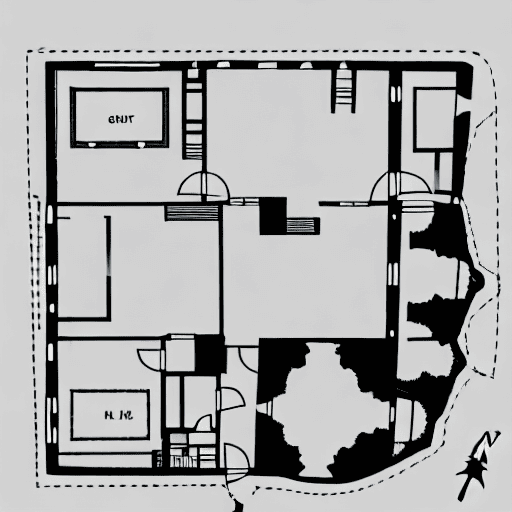} & &
        \includegraphics[trim={1cm 1cm 1cm 1cm},clip,height=2cm]{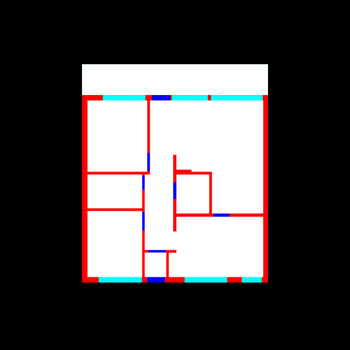} & \includegraphics[trim={1cm 1cm 1cm 1cm},clip,height=2cm]{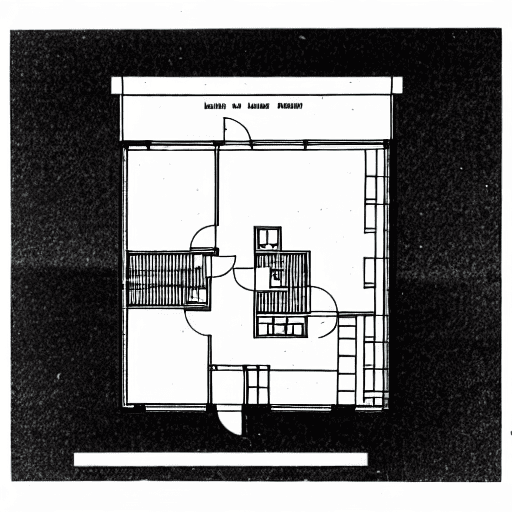} & & 
        \includegraphics[height=2cm]{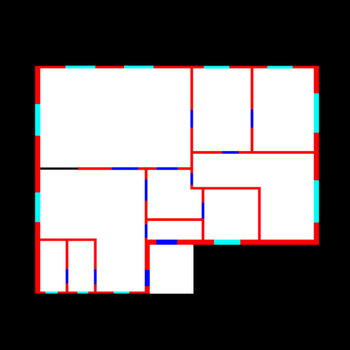} & \hspace{-3pt}\includegraphics[height=2cm]{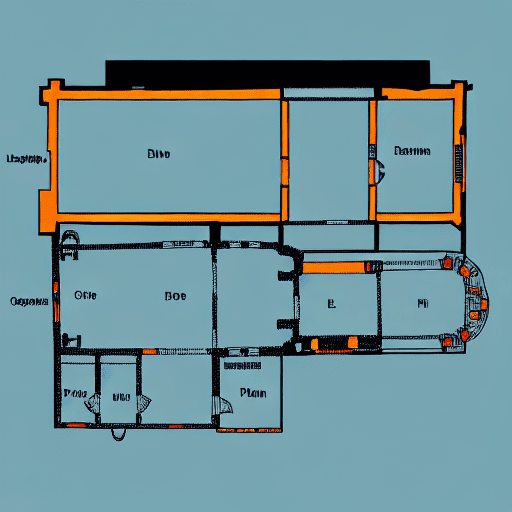}
     \end{tabular}
     \vspace{-5pt}
    \caption{Structure-conditioned generation. For each image pair, the first image displays a building layout condition, taken from the existing CubiCasa5K dataset, which defines foreground (white) and background (black) regions, walls (red), doors (blue), and windows (cyan). The second image shows a generation conditioned on this layout, using the ControlNet-based model described in Section \ref{sec:SCFG}. Our image data and metadata enable the generation of diverse building types with structural constraints, without requiring any pixel-level annotations of images in \datasetname{}. Notably, this succeeds even when the constraint is highly unusual for the corresponding building type, such as the condition above for \emph{cathedral} (as cathedrals are usually constructed in a cross shape).
    }
    \label{fig:structural_examples}
\end{figure*}
\vspace{-8pt}

\subsection{Benchmark for Semantic Segmentation}
\label{sec:benchmark}

\modified{Following prior work~\cite{liu2017raster,kalervo2019cubicasa5k,wu2019data}, we consider segmentation of rasterized floorplan images into fine-grained localized categories, as locating elements such as walls has applications to various downstream tasks. To provide a new benchmark for performance on the diverse floorplans in \datasetname{}, we manually annotate pixel-level segmentation maps for more than a hundred images over categories applicable to most building types: \emph{wall}, \emph{door}, \emph{window}, \emph{interior} and \emph{background}. As our dataset contains a variety of data types, we annotate SVG-formatted images, which can be easily manually annotated by region.}

\modified{We illustrate the utility of this benchmark by evaluating a standard existing model, namely the supervised segmentation model provided by CC5K~\cite{kalervo2019cubicasa5k}. We also evaluate a modern diffusion-based architecture trained with the same supervised data to predict wall locations as black-and-white images, to explore whether architectural modifications can yield improved performance. Further details of these models are provided in the supplementary material.}

\medskip \noindent \textbf{Results.}
\modified{Table \ref{tab:benchmark_segmentation} includes a quantitative evaluation of the existing model provided by CC5K on our benchmark. As illustrated in the table, our dataset provides a challenging benchmark for existing models, yielding low performance, particularly for more fine-grained categories, such as doors and windows. }
\modified{In addition to these results, we find that the modern diffusion architecture shows significantly better performance at localization of walls, generating binary maps with higher quantitative metric values (+1.2\% in precision, +36.4\% in recall and +29.5\% in IoU, in comparison to the values obtained on the wall category in Table \ref{tab:benchmark_segmentation}).} %
\modified{This additional experiment shows promise in using stronger architectures for improving localized knowledge on weakly supervised in-the-wild data to ultimately approach the goal of pixel-level localization within diverse floorplans. Qualitative results from both models, along with ground truth segmentations, are provided in the supplementary material.}

\subsection{Text-Conditioned Floorplan Generation} \label{sec:FTSD}
\noindent \textbf{Task description.} Inspired by the rich literature on automatic floorplan generation, we fine-tune a text-to-image generation model on paired images and pGT textual data from \datasetname{} for text-guided generation of floorplan images. We adopt the latent diffusion model Stable Diffusion~\cite{rombach2022highresolution} (SD), using prompts of the form ``A floor plan of a \texttt{<building\_type>}'' which use the building type pseudo-labels from our LLM-extracted data. We balance training samples across building names and types to avoid overfitting on common categories. %
\modified{We evaluate the realism of these generations using Fréchet Inception Distance (FID)~\cite{heusel2018gans} as well as Kernel Maximum Mean Discrepancy (KMMD), since FID can be unstable on small datasets~\cite{chong2020effectivelyunbiasedfidinception}. Similar to prior
work~\cite{noguchi2019imagegenerationsmalldatasets,wang2020minegan,chen2023s}, we measure KMMD on Inception features~\cite{szegedy2016rethinking}. To measure semantic correctness, we measure CLIP similarity (using pretrained CLIP) 
between generations and prompts. All metrics were calculated per building type, averaging over the most common 15 types.} 
\begin{table}[t]
  \centering
  \begin{tabular}{lccc}
    & FID $\downarrow$ & \modified{KMMD $\downarrow$} & CLIP Sim. $\uparrow$ \\
    \toprule
    SD & \modified{194.8} & \modified{0.10} & \modified{24.9} \\
    SD$_{FT}$ & \modified{\textbf{145.3}} & \modified{\textbf{0.07}} & \textbf{25.6} \\
    \bottomrule
  \end{tabular}
  \vspace{-7pt}
  \caption{Results on generated images, using a base and fine-tuned Stable Diffusion (SD) model. We compare the quality of the generated images (FID\modified{, KMMD}) and the similarity to the given prompt (CLIP Sim.). As illustrated above, SD fine-tuning improves both realism and semantic correctness of image generations.}
  \vspace{-7pt}
\label{tab:gen_metrics}
\end{table}

\medskip \noindent \textbf{Results.} Table \ref{tab:gen_metrics} summarizes quantitative metrics, comparing floorplan generation using the base SD model with our fine-tuned version. These provide evidence that our model generates more realistic floorplan images that better adhere to the given prompts. Supporting this, we provide examples of such generated floorplans in Figure \ref{fig:generated_examples}, observing the diversity and semantic layouts predicted by our model for various prompts. We note that our model correctly predicts the distinctive elements of each building type, such as the peripheral towers of castles and numerous side rooms for patient examinations in hospitals. Such distinctive elements are mostly not observed in the pretrained SD model, which generally struggles at generating floorplans. %
\modified{To further illustrate that our generated floorplans better convey the building type specified within the target text prompt, we conducted a user study. Given a pair of images, one generated with the pretrained model and one with our finetuned model, users were asked to select the image that best conveys the target text prompt. We find that 70.42\% of the time users prefer our generated images, in comparison to the generations of the pretrained model. %
Additional details regarding this study are provided in the supplementary.}

\subsection{Structure-Conditioned Floorplan Generation} \label{sec:SCFG}
\noindent \textbf{Task description.} Structural conditions for floorplan generation have attracted particular interest, as architects may wish to design floorplans given a fixed building shape or desired room configuration~\cite{hua2016irregular,merrell2010computer,wu2019data,hu2020graph2plan}. Unlike existing works that consider residential buildings exclusively, we operate on the diverse set of building types and configurations found in \datasetname{}, providing conditioning to our generative model SD$_{FT}$ by fine-tuning ControlNet~\cite{zhang2023adding} for conditional generation, combined with applying text prompts reflecting various building types. %

We are challenged by the fact that data in \datasetname{}, captured in-the-wild, does not contain localized structural annotations (locations of walls, doors, windows or other features) such as those painstakingly annotated in some existing datasets. Therefore, we leverage our data in an unsupervised manner to achieve conditioning. To condition on the desired building boundary, we approximate the outer shape for all images in the training set via an edge detection algorithm. To condition on more complex internal structures, we instead train ControlNet on image-condition pairs derived from the existing annotated CubiCasa5K dataset~\cite{kalervo2019cubicasa5k}. By using SD$_{FT}$ as the backbone for this ControlNet, reducing the conditioning scale (assigning less weight to the conditioning input during inference) and using a relatively high classifier-free guidance scale factor (assigning higher weight to the prompt condition), we fuse the ability of SD$_{FT}$ to generate diverse building types while incorporating the structural constraints derived from external annotated data of residential buildings.

\medskip \noindent \textbf{Results.} We provide structurally-conditioned generations in Figures \ref{fig:controlnet_examples}--\ref{fig:structural_examples} for various building types. For boundary conditioning, the condition shape is extracted from existing images in our dataset. For structure conditioning, the conditions are derived from the annotations in the external CubiCasa5K dataset, using categories relevant to the diverse buildings in \datasetname{}. These examples illustrate that our model is able to control the contents and style of the building according to the text prompt while adhering to the overall layout of the condition. This again demonstrates that the model has learned the distinct characteristics of each building type. In addition, we note that this succeeds even when the structural constraint is highly unusual for the paired building type, such as the cathedral in Figure \ref{fig:structural_examples} which deviates from the typical layout of a cathedral (usually constructed in a cross shape) in order to obey the condition.

\section{Conclusion}

We have presented the \datasetname{} dataset of diverse floorplans in the wild, curated from Internet data spanning diverse building types, geographic locations, and architectural features. To construct a large dataset of images with rich, structured metadata, we leverage SOTA LLMs and multimodal representations for filtering and extracting structure from noisy metadata, reflecting both the global semantics of buildings and localized semantics grounded in regions within floorplans. We show that this data can be used to train models for building understanding tasks, enabling progress on both discriminative and generative tasks which were previously not feasible.

While our dataset expands the scope of floorplan understanding to new unexplored tasks, it still has limitations. As we collect diverse images in-the-wild, our data naturally contains noise (mitigated by our data collection and cleaning pipeline) which could affect downstream performance. In addition, while our dataset covers a diverse set of building types, it leans towards historic and religious buildings, possibly introducing bias towards these semantic domains. We focus on 2D floorplan images, though we see promise in our approach and data for spurring further research in adjacent domains, such as 3D building generation and such as architectural diagram understanding in general.
In particular, although our work does not consider the 3D structure of buildings, we see promise in the use of our floorplans for aligning in-the-wild 3D point clouds or producing 3D-consistent room and building layouts. 
Finally, our work could provide a basis for navigation tasks which require indoor spatial understanding, such as indoor and household robotics. We envision future architectural understanding models that are enabled by datasets such as \datasetname{} will explore new challenging tasks such as visual question answering for floorplans, which could be enabled by our textual metadata and open-vocabulary architectural features.

{\small
\bibliographystyle{ieee_fullname}
\bibliography{egbib}
}

\clearpage

\setcounter{section}{0}
\def\thesection{\Alph{section}}
\appendix
{\LARGE\textbf{Appendix}}
\setcounter{page}{1}
\setcounter{section}{0}

\section{Interactive Visualization Tool}
Please see the attached HTML file (\texttt{waffle.html}) for an interactive visualization of data from the \datasetname{} dataset.

\section{Additional Dataset Details}

We proceed to describe the creation of our \datasetname{} dataset in the sections below, including details on curating, filtering, and generating pseudo-ground truth labels.

\subsection{Model Checkpoints and Settings}

We use Llama-2 \cite{touvron2023llama} for text-related tasks, and CLIP \cite{radford2021learning} for image-related tasks. \noindent For most text related tasks we use the \texttt{meta-llama/Llama-2-13b-chat-hf} model, and for legend extraction we use the \texttt{meta-llama/Llama-2-70b-chat-hf} model.
In both cases, we use the default sampling settings defined by the Hugging Face API. For CLIP, we use the \texttt{openai/clip-vit-base-patch32} model.

\subsection{Layout Component Detection}
As part of the data collection, we train a DETR \cite{carion2020end} object detection model to identify common floorplan layout components which will be later on used for the pGT extraction and in the segmentation experiment. We use the checkpoint \texttt{\small TahaDouaji/detr-doc-table-detection}\footnote{https://huggingface.co/TahaDouaji/detr-doc-table-detection} as the base model, and fine-tune it on ~200 manually annotated images with augmentations, using the following labels: \texttt{floorplan, legend, scale, compass}. We fine-tune for 1,300 iterations, a batch size of 4, and a $10^{-4}$ learning rate on one A5000 GPU, splitting our data into 80\% training images and 20\% test images.

\subsection{Data Filtering}

As described in the paper, we first scrape a set of images and metadata from Wikimedia Commons and proceed to filter to only select images of floorplans using a two-stage process: text-based filtering with an LLM, and image-based filtering with CLIP. All models used for these stages are described in our main paper.

\subsubsection{Text-based filtering (LLM)}
First, we query an LLM in order to obtain an initial categorization of our raw data. We ask it to choose what the image is most likely a depiction of out of a closed set of categories (i.e. multiple choice question format), marked positive (e.g. \textit{floorplan} or \textit{building}) or negative (e.g. \textit{map} or \textit{park}), and we filter out images categorized as negative categories. The full prompt is shown on the leftmost column of Figure \ref{tab:llm_prompts}, where options A and B are treated as positive and the rest are negative.
\begin{figure} {
\centering
\begin{tabular}{p{1.1cm}|p{6.45cm}}
\toprule
\small Prefixes &\cellcolor{cyan!25} \texttt{\ssmall"an illustration of ", "a drawing of ", "a sketch of ", "a picture of ",
    "a photo of ",
    "a document of ",
    "an image of ",
    "a visual representation of ",
    "a graphic of ",
    "a rendering of ",
    "a diagram of ",
    ""} \\
\midrule
\small Negative Prefixes &\cellcolor{orange!25}\texttt{\ssmall"a 3d simulation of ",
    "a 3d model of ",
    "a 3d rendering of "} \\
\midrule
\small Positive Suffixes & \cellcolor{cyan!25}\texttt{\ssmall"a floor plan",
    "an architectural layout",
    "a blueprint of a building",
    "a layout design"} \\
\midrule
\small Negative Suffixes &\cellcolor{orange!25} \texttt{\ssmall"a map",
    "a building",
    "people",
    "an aerial view",
    "a cityscape",
    "a landscape",
    "a topographic representation",
    "a satellite image",
    "geographical features",
    "a mechanical design",
    "an engineering sketch",
    "an abstract pattern",
    "wallpaper",
    "a Window plan",
    "a staircase plan"} \\
\bottomrule
\end{tabular}} 
\vspace{-5pt}
\caption{The prompts used for gathering CLIP scores. (\textcolor{cyan}{Prefixes} $\times$ \textcolor{cyan}{Positive Suffixes}) are used as positive prompts, and (\textcolor{orange}{(Prefixes + Negative Prefixes)} $\times$ \textcolor{orange}{Negative Suffixes}) are negative prompts.}  
 \label{tab:clip_prompts}  
\end{figure}

\subsubsection{Image-based filtering (CLIP)}
We proceed to use image-based filtering to yield our final dataset. This is composed of two sub-stages: first, we generate a smaller set of highly accurate images (a \emph{seed}); we then extend this seed to produce an enlarged dataset. These sub-stages are described below.

\medskip \noindent \textbf{Seed generation.} We start by creating a highly accurate seed of images (\emph{i.e.} containing floorplans) by aggressively filtering according to the CLIP normalized scores extracted over positive and negative prompts. We list the prompts used in Figure \ref{tab:clip_prompts}. As illustrated in the figure, negative prompts correspond to images that depict categories similar to floorplans, such as maps or satellite images. We sort the prompts by score, and add images to the set if it passes the following two tests: (i) All top five prompts contain \emph{floor plan}, and (ii) The sum of all prompts containing \emph{floor plan} is over 0.5. We empirically find that these tests allow for creating a highly accurate seed of 3,402 images.

\medskip \noindent \textbf{Dataset extension.} Next, we use this seed to bootstrap an image classifier, in order to enlarge the dataset.
We first use this seed to train a vision transformer binary classifier. We take 1K images from the seed as positive samples, and 1K images that were categorized as a negative category in the text-based filtering step as negative samples. We fine-tune a ViT model~\cite{dosovitskiy2020image} for 5 epochs with a batch size of 4 and a $2 * 10^{-4}$ learning rate on one A5000 GPU, splitting our data into 1,400 training images, 300 validation and 300 test images. 

To create our final dataset, we select images that pass the following two tests: (i) classifier threshold selected to filter out 10\% of data, and (ii) the sum of normalized scores on all positive prompts (described in Figure \ref{tab:clip_prompts}) is over 0.5.

Altogether this leads us to the final dataset of ${\sim}$19K floorplans. 
In addition to the manual validation over 100 random sampled images in the dataset, we also manualy inpect the entire test set, and remove all image that do not contain a valid floorplan. Based on this validation, we estimate that 89\% of images from our full dataset are indeed floorplans.

\subsection{LLM-Driven Structured pGT Generation}

Figure \ref{tab:llm_prompts} contains the prompts used for extracting the pseudo-ground truth (pGT) labels for our dataset. Note that some prompts use previously extracted pGTs as inputs, such as those for ``Building Type'' and ``Location Information''.

The architectural feature grounding process is split into two: legend extraction from the image metadata, and architectural information extraction from the image.

\begin{figure}
    \centering
    \includegraphics[width=1\linewidth]{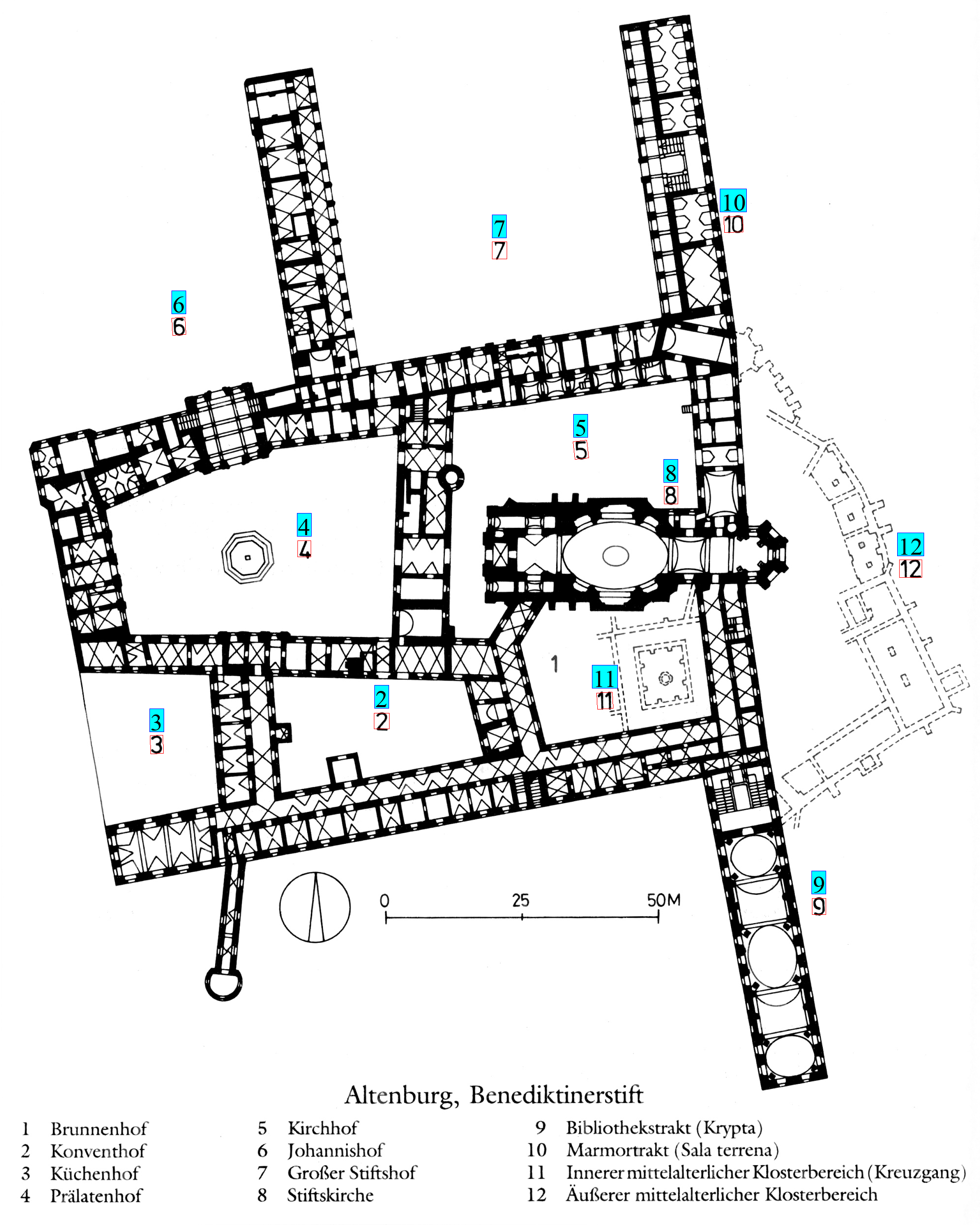}
    \vspace{-10pt}
    \caption{An example of an image which contains legend text, seen as rasterized text underneath the floorplan. Our legend key-grounding correctly detects the keys in the image and can successfully avoid incorrect grounding such as to the legend depicted below.}
    \label{fig:legend_in_image_example}
\end{figure}

\label{sec:legend_metadata}
\medskip \noindent \textbf{Legend Structuring from Metadata.} We divide the task of legend structuring into four sub-tasks: (i) Legend raw content extraction (the raw text containing key--value pairs), extracted using the prompts in Figure \ref{tab:legend_prompts}; (ii) Key--value identification (raw text structurization) using regular expressions on the raw text legend; (iii) Legend content simplification, using the prompt in Figure \ref{tab:legend_simplification_prompt}; and (iv) grounding the architectural features in the legend to the image, by marking the keys of the legend in the image, mapping between the legend values and the key locations. The last sub-task is obtained by searching for the keys in the image's OCR detected texts. Images can sometimes contain the full key--value legend itself in addition to the key markings (as seen in Figure \ref{fig:legend_in_image_example} for example). To avoid marking these as well, we leverage the multiple text granularities returned by the Google Vision API and filter out identified keys that are part of a sentence/paragraph, and exclude areas detected as 'legend' by our layout component detector.

\label{sec:arc_legend_image}
\medskip \noindent \textbf{Architectural Information from the Image.}
As mentioned in the main paper, we use the OCR detections within the relevant layout components as candidates to include interesting architectural information -- legends in \texttt{legend} areas, and architectural labels in \texttt{floorplan} areas.
Next, we send these candidates to the LLM (similarly to the metadata legend extraction process) using the prompts in Figure \ref{tab:legend_arc_ocr_prompts} to obtain a raw legend/list of architectural labels. The legends are formatted and grounded similarly to those extracted from the metadata. The architectural features are translated if they are not in English using
the Google Translate API
while maintaining the original text representation which is used to ground them to the image.

\begin{figure*}
    \centering  \includegraphics[width=\textwidth]{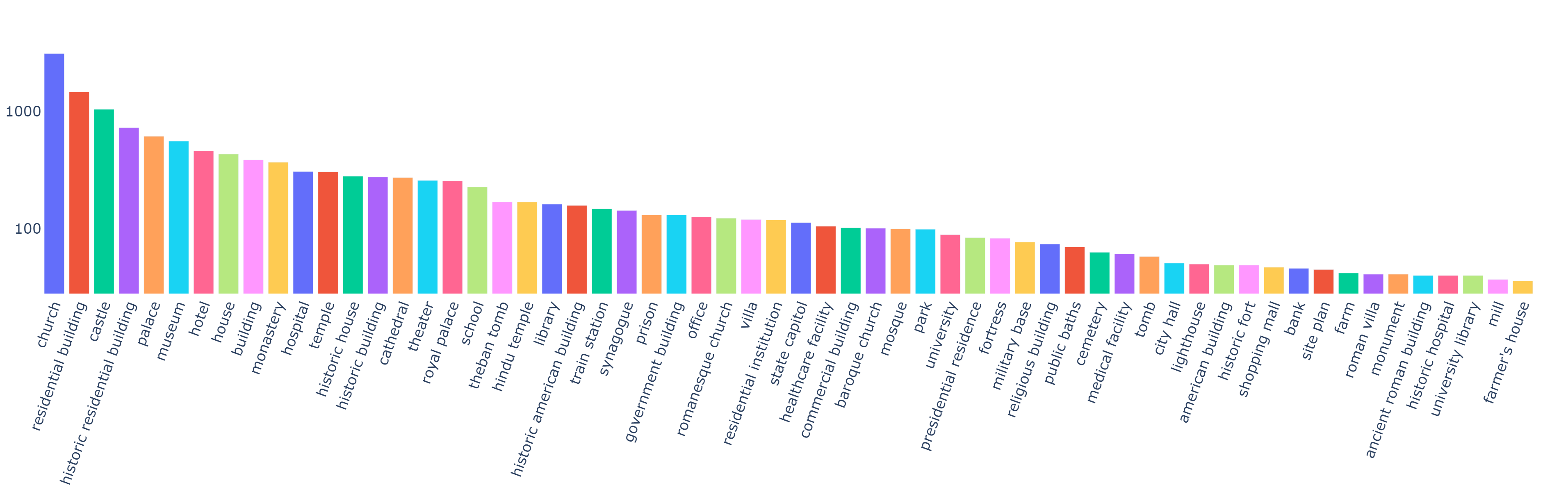}
    \caption{Distribution of common building types extracted automatically (log scale), illustrating the rich semantics captured in \datasetname{}.}
    \label{fig:statistics}
\end{figure*}

\begin{figure*}
    \centering  \includegraphics[width=\textwidth]{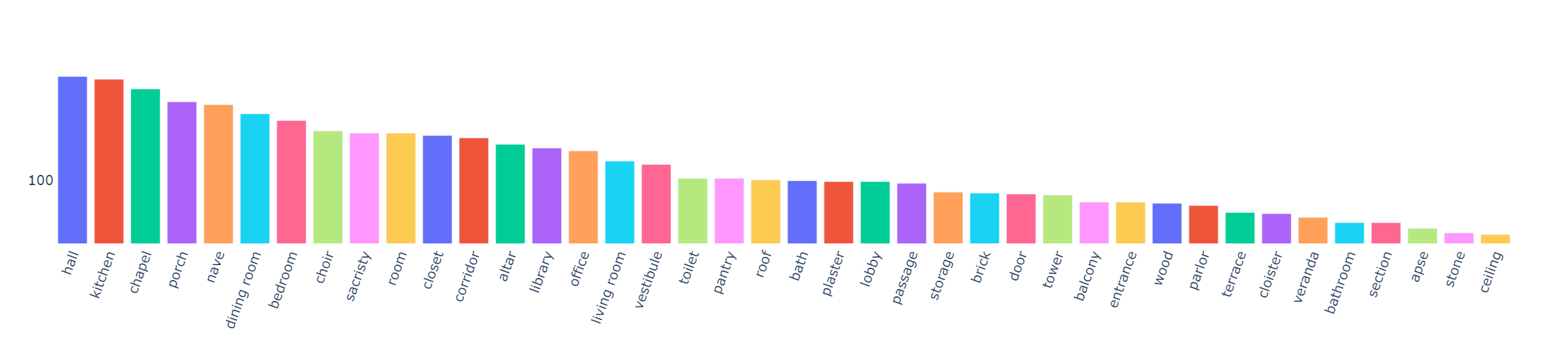}
    \caption{Distribution of the grounded architectural features (log scale), among almost 3K grounded images, 25K instances grounded, and 11K unique features.}
    \label{fig:grounded_arc_feats}
\end{figure*}

\begin{figure}
    \centering  \includegraphics[width=\linewidth]{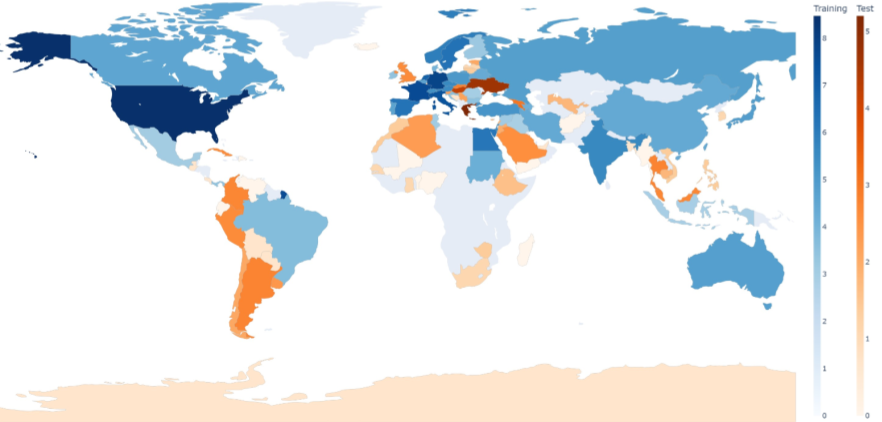}
    \caption{Number of samples per country (log scale) in \datasetname{}, showing the diversity of our dataset for both training and test splits. \color{blue}{Blue}: training data; \color{orange}{Orange}: test data.}
    \label{fig:countries}
\end{figure}

\begin{figure}
    \centering  \includegraphics[width=\linewidth]{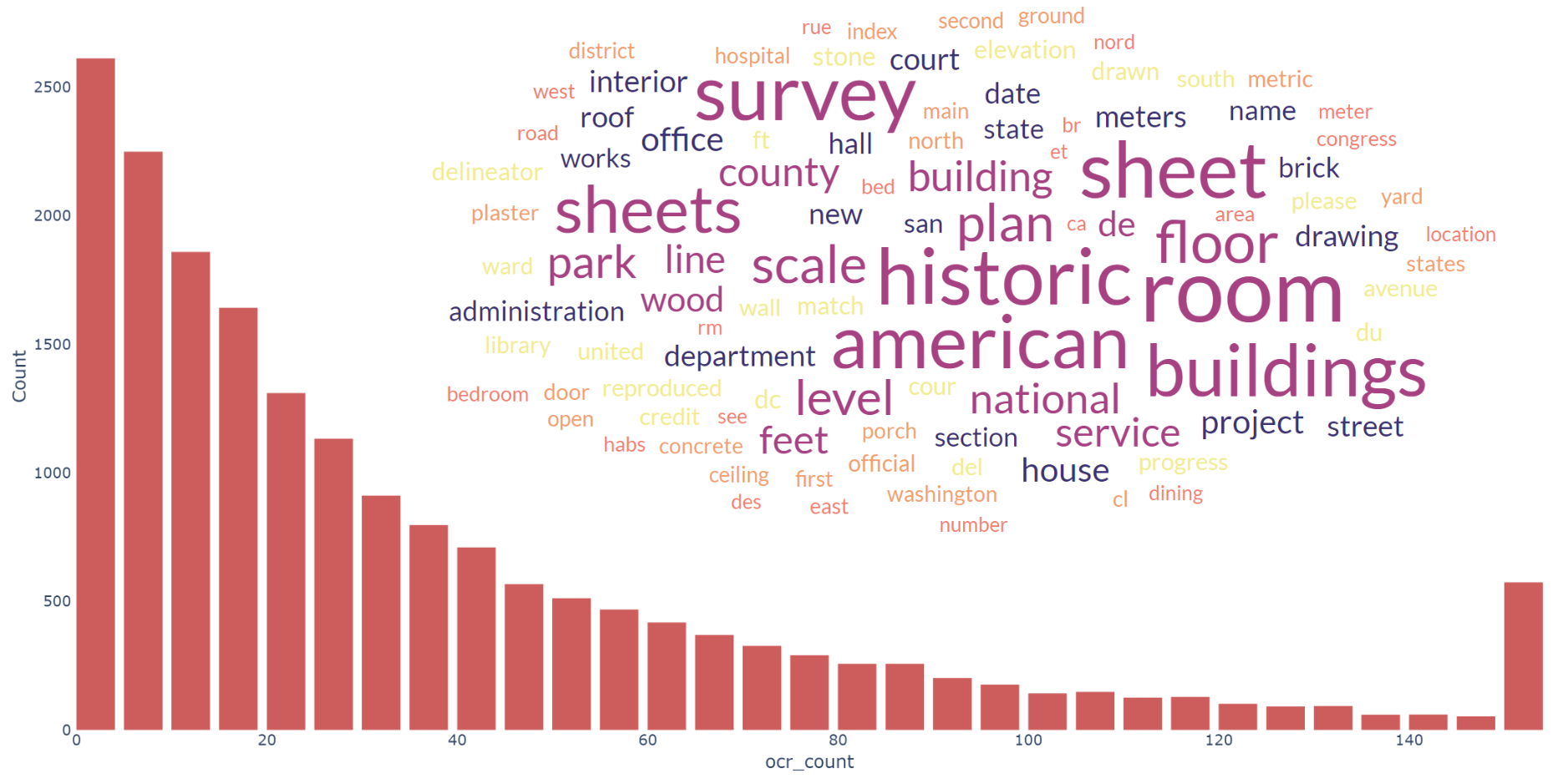}
    \caption{OCR words statistics. The bar plot depicts a histogram of the number of words detected in an image; the word map on the top right shows the most common words detected in our dataset. As illustrated above, the raw OCR data contains semantic information and also significant levels of noise, and thus it is challenging to operate over this data directly; hence motivating the need for extracting data from source external to the images (\emph{e.g.} linked Wikipedia pages).}
    \label{fig:word_cloud}
\end{figure}

\subsection{Dataset statistics}
Figure \ref{fig:countries} shows a visualization of the different countries in our dataset; Figure \ref{fig:statistics} shows a histogram of common building types in the dataset; Figure \ref{fig:word_cloud} shows a visualization of the distribution of words in images detected using OCR; and Figure \ref{fig:grounded_arc_feats} show a histogram of the common architectural features that are grounded in \datasetname{} images.

\section{Experimental Details}

\subsection{Building Type Understanding Task}
For the building type understanding task, we fine-tune CLIP on our training set. We train it for 5 epochs with batch size of 256, learning rate of $10^{-3}$ and Adam optimizer, on one A5000 GPU.

\subsection{Open-Vocabulary Floorplan Segmentation Task}
We learn image segmentation by fine-tuning CLIPSeg~\cite{luddecke2022image} (using base checkpoint \texttt{CIDAS/clipseg-rd64-refined}) on images with corresponding positive and negative segmentation maps.

\begin{figure}
  \centering
  \jsubfig{\includegraphics[width=6cm]{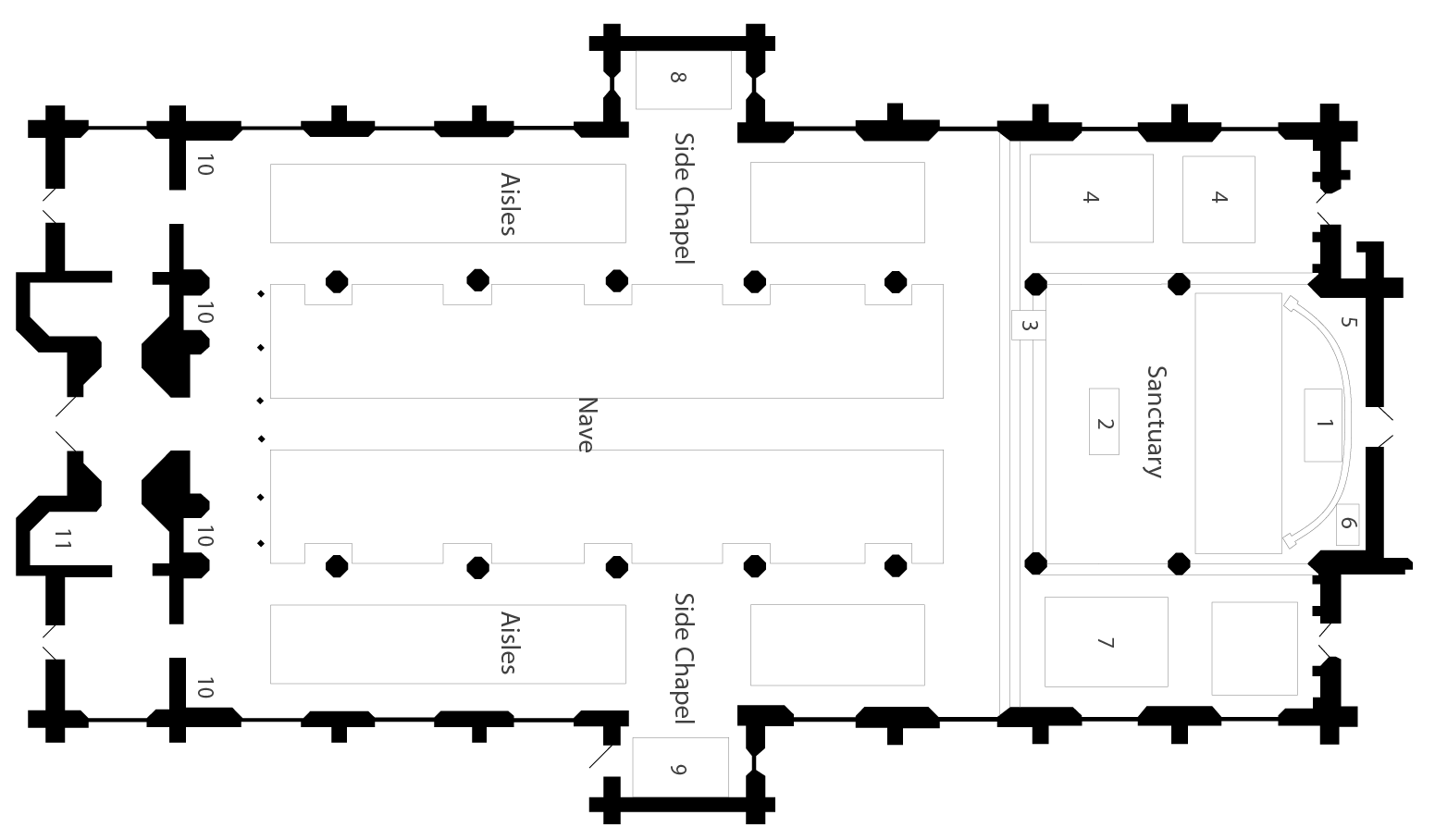}}{Original}
  \hfill
  \jsubfig{\includegraphics[width=6cm]{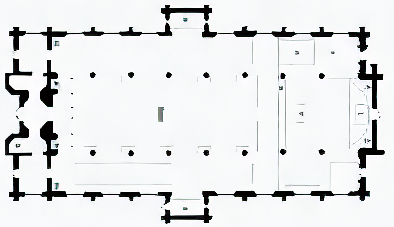}}{In-painted}
  \caption{An example of in-painting to preprocess data for the segmentation task. On the left, we show the original image which contains text indicating architectural features, including the \emph{Nave}, \emph{Side Chapels} and \emph{Aisles}. On the right, we show the in-painted version of the image, which succeeds in removing these texts to prevent leakage. We observe that this in-painting process may slightly modify the appearance of the image, but the floorplan's structure is mostly preserved.}
  \label{fig:eval_gt_segmentation_example}
\end{figure}

\medskip \noindent
\textbf{Training details.}
Training data for segmentation is created from images with grounded architectural features (GAFs), using those that occur over 10 times in our dataset in order to filter out noise.
To avoid leakage of information from rendered text, OCR detections are removed via in-painting as seen in Figure \ref{fig:eval_gt_segmentation_example}.
During training, we also apply augmentations such as cropping, resizing and noising to enlarge our training dataset, applied to both images and target segmentation maps as needed.

To identify positive and negative targets for a given image and GAF text, we use the text embedding model \texttt{\small paraphrase-multilingual-mpnet-base-v2} from SentenceTranformers\footnote{https://www.sbert.net/}, measuring semantic similarity between GAF texts via embedding cosine similarity. Pairs of features with high ($> 0.7$) similarity scores are marked as positive and those with low ($< 0.4$) similarity are marked as negative; the loss is calculated on these areas alone.

Our overall loss consists of the weighted sum of three losses: cross-entropy over the masked positive and negative areas ($\mathcal{L}_{ce}$), L1 regularization loss ($\mathcal{L}_{L1}$) and entropy loss (mean of binary entropies of pixel intensities on the whole image) ($\mathcal{L}_{e}$). Our total loss is $\mathcal{L}_{total} = \frac{1}{2}\mathcal{L}_{ce} + \frac{1}{2}\mathcal{L}_{L1} + \mathcal{L}_{e}$.

We fine-tune with the following settings: 20 epochs; batch size 1; on one A5000 GPU; with a $10^{-4}$ learning rate; with an Adam \cite{kingma2014adam} optimizer. 
 
\begin{table*}[t]
  \centering
  \setlength{\tabcolsep}{3.5pt}
  \begin{tabularx}{1\linewidth}{cccccccccccccccc}
    &
    {\small\rotatebox[origin=l]{90}{Building}} &
    {\small\rotatebox[origin=l]{90}{Castle}} &
    {\small\rotatebox[origin=l]{90}{Cathedral}} &
    {\small\rotatebox[origin=l]{90}{Church}} &
    {\small\rotatebox[origin=l]{90}{Historic building}} &
    {\small\rotatebox[origin=l]{90}{Hospital}} &
    {\small\rotatebox[origin=l]{90}{Hotel}} &
    {\small\rotatebox[origin=l]{90}{House}} &
    {\small\rotatebox[origin=l]{90}{Monastery}} &
    {\small\rotatebox[origin=l]{90}{Museum}} &
    {\small\rotatebox[origin=l]{90}{Palace}} &
    {\small\rotatebox[origin=l]{90}{Residential building}} &
    {\small\rotatebox[origin=l]{90}{Temple}} & {\small\rotatebox[origin=l]{90}{School}} & {\small\rotatebox[origin=l]{90}{Theater}} \\
    \midrule
    \small \textbf{FID} $\downarrow$ \\
     \small  SD$\;\;\,\,$ & \small 284.6 & \small 159.6 & \small 198.4 & \small 188.3 & \small 285.1 & \small 199.2 & \small 194.3 & \small 212.2 & \small 159.2 & \small 180.6 & \small 139.6 & \small 224.0 & \small \textbf{165.9} & \small 141.6 & \small \textbf{189.5}
     \\
     \small SD$_{FT}$ & \small \textbf{148.3} & \small \textbf{146.1} & \small \textbf{156.8} & \small \textbf{142.0} & \small \textbf{147.7} & \small \textbf{114.5} & \small \textbf{100.9} & \small \textbf{158.0} & \small \textbf{122.3} & \small \textbf{137.6} & \small \textbf{119.1} & \small \textbf{168.6} & \small 176.9 & \small \textbf{102.0} & \small 238.4
     \\
    \midrule
    \small \textbf{KMMD} $\downarrow$ \\
     \small  SD$\;\;\,\,$ & \small 0.13 & \small 0.06 & \small 0.11 & \small 0.09 & \small 0.16 & \small 0.09 & \small 0.13 & \small 0.12 & \small 0.08 & \small 0.07 & \small 0.05 & \small 0.13 & \small \textbf{0.07} & \small 0.06 & \small \textbf{0.08}
     \\
     \small SD$_{FT}$ & \small \textbf{0.07} & \small \textbf{0.05} & \small \textbf{0.06} & \small \textbf{0.04} & \small \textbf{0.11} & \small \textbf{0.05} & \small \textbf{0.05} & \small \textbf{0.08} & \small \textbf{0.03} & \small \textbf{0.05} & \small \textbf{0.04} & \small \textbf{0.11} & \small 0.09 & \small \textbf{0.03} & \small 0.17
     \\
    \midrule
    \small \textbf{CLIP Sim.} $\uparrow$ \\
     \small SD$\;\;\,\,$ & \small 25.3 & \small 25.2 & \small 24.4 & \small 24.2 & \small \textbf{25.3} & \small 24.0 & \small 24.2 & \small 25.8 & \small 25.9 & \small 25.2 & \small \textbf{25.5} & \small 25.1 & \small 24.7 & \small 24.6 & \small \textbf{24.6}
    \\
     \small SD$_{FT}$ & \small \textbf{25.9} & \small \textbf{25.6} & \small \textbf{25.6} & \small \textbf{25.4} & \small \textbf{25.3} & \small \textbf{24.5} & \small \textbf{25.1} & \small \textbf{26.7} & \small \textbf{26.5} & \small \textbf{26.1} & \small \textbf{25.5} & \small \textbf{26.0} & \small \textbf{25.6} & \small \textbf{25.8} & \small 24.5
    \\
    \bottomrule

  \end{tabularx}
  \caption{Quantitative results of results on floorplan image generation split by building type, comparing the quality of images generated with the pretrained model and our fine-tuned model.}
\label{tab:metrics_full}
\end{table*}

\medskip \noindent \textbf{Evaluation.}
We manually annotate 95 images for evaluation with 27 common GAFs. The most common building types in our evaluation set are churches, castles and residential buildings.

\subsection{Floorplan Generation Task}
For the generation task, we adopt the text to image example provided in Hugging Face\footnote{\url{https://huggingface.co/docs/diffusers/v0.18.2/en/training/text2image}}, by fine-tuning the Stable Diffusion (SD)~\cite{rombach2022highresolution} model \newline \texttt{CompVis/stable-diffusion-v1-4}.  We add a custom sampler to avoid over-sampling the same building; in particular, in each epoch we use only one sample out of all those corresponding to a given \texttt{<building\_type>} and \texttt{<building\_name>}. In addition, we resize the images to $512\times512$ keeping the original proportions of the image and adding padding as needed.
We train our model for 20K iterations with batch size of 4, a learning rate of $10^{-5}$ and Adam optimizer, on one A5000 GPU.

For the boundary-conditioned generation task (conditioned on the outer contour of the building), we first extract the outer edges for all images in the training set. We use the Canny edge detection algorithm as implemented in the OpenCV library\footnote{\url{https://docs.opencv.org/4.x/da/d22/tutorial_py_canny.html}}, extract only external contours\footnote{\url{https://docs.opencv.org/4.x/d9/d8b/tutorial_py_contours_hierarchy.html}}, and remove contours with small areas. Samples where edge detection fails (returns an empty mask) are excluded.
We then fine-tune\footnote{\url{https://huggingface.co/docs/diffusers/v0.18.2/en/training/controlnet}} ControlNet~\cite{zhang2023adding}. We initialize the SD part of the architecture with our fine-tuned SD from the previous paragraph and the shape-condition part with a pre-trained model trained on Canny edges masks (as this condition is similar to our task) (\texttt{lllyasviel/sd-controlnet-canny})\footnote{\url{https://huggingface.co/blog/controlnet}} We use the same custom sampler and resize as described above. We train the model for 15K iterations with a batch size of 4, a learning rate of $10^{-5}$ and Adam optimizer, on one A5000 GPU.

For structure-conditioned generation (conditioned on building layouts), we use our fine-tuned SD model from above and fine-tune ControlNet on top of it. Unlike the boundary-conditioned task, we fine-tuned ControlNet using external data from the CubiCasa5K (CC5K)~\cite{kalervo2019cubicasa5k} train set. As conditions, we convert the structured CC5K SVG data into images with pixel values representing the subset of categories relevant to our dataset: foreground (white), background (black), walls (red), doors (blue), and windows (cyan). The foreground category comprises all CC5K room categories that are not background; doors and windows are taken from the CC5K icon categories and overlaid on top of foreground/background/walls to produce the condition image (rather than being independent layers). We initialize the SD part of the architecture with our fine-tuned SD from the previous paragraph and the shape-condition part with the model's UNet weights. Images and conditions are resized using the method described above. We train the model for 20K iterations with a batch size of 4, a learning rate of $10^{-5}$ and Adam optimizer, on one A5000 GPU. During inference we use CFG scale 15.0 and condition scale 0.5, to fuse the style of the input prompt (learned from floorplans in \datasetname{}) and the structure condition.

\begin{figure}
    \centering
    \includegraphics[width=\linewidth]{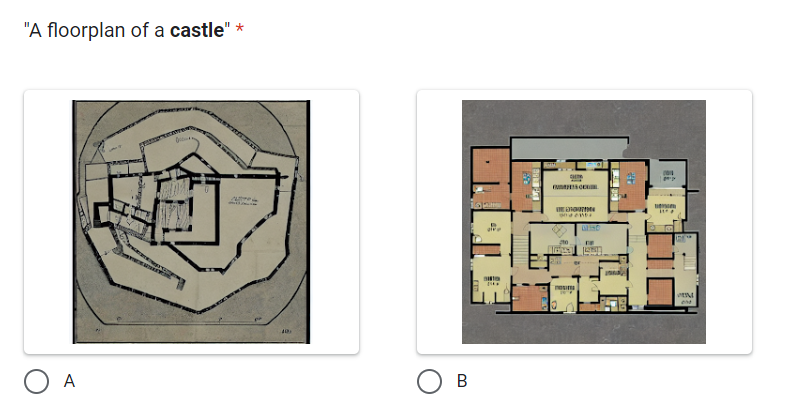}
    \caption{\modified{A sample question from our user study on text-conditioned floorplan generation}.}
    \label{fig:user_study}
\end{figure}

\medskip \noindent \textbf{\modified{User study.}}
\modified{Each study contained 36 randomly-generated image pairs, with text prompts mentioning various building types that were sampled from the 100 most common types. Overall, thirty one users participated in the study, resulting in a total of 1,116 image pairs (one generated from the pretrained model, and the other generated from the finetuned model) that were averaged for obtaining the final results reported in the main paper.}

\modified{Participants were provided with the following instructions: \textit{In this user study you will be presented with a series of pairs of images, generated by the prompt: "a floorplan of a $<$BUILDING\_TYPE$>$". For example, "a floorplan of a cathedral". For each pair,  please select the image that best conveys the text prompt (i.e., both looks like a \textbf{floorplan diagram}, and also looks like a plan of the specific \textbf{building type} mentioned in the prompt). If you are unsure, please make an educated guess to the best of your ability. Thank you for participating!}}

\modified{A sample question from our user study is illustrated in Figure \ref{fig:user_study}. %
All of the questions were forced-choice, and participants could only submit after answering all of of the questions.}

\subsection{Benchmark for Semantic Segmentation}

\begin{figure}
    \centering
\jsubfigcent{{\includegraphics[width=2.3cm]{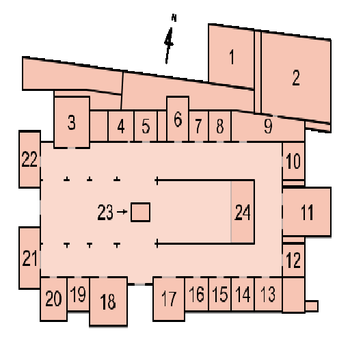}}}{}
\hfill
\jsubfigcent{{\includegraphics[width=2.3cm]{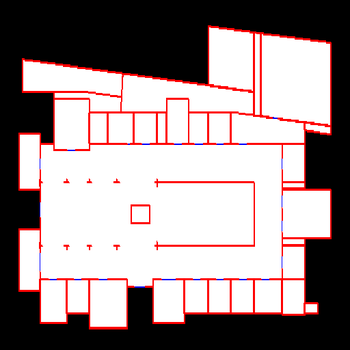}}}{}
\hfill
\jsubfigcent{{\includegraphics[width=2.3cm]{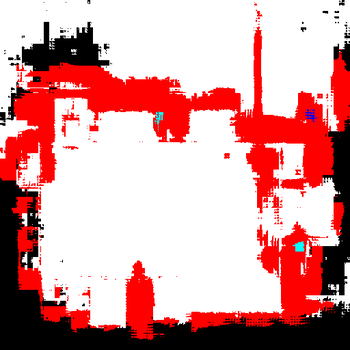}}}{}
\\
\jsubfigcent{{\includegraphics[width=2.3cm]{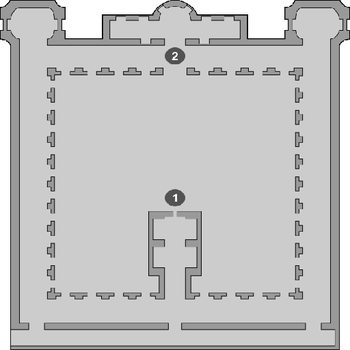}}}{}
\hfill
\jsubfigcent{{\includegraphics[width=2.3cm]{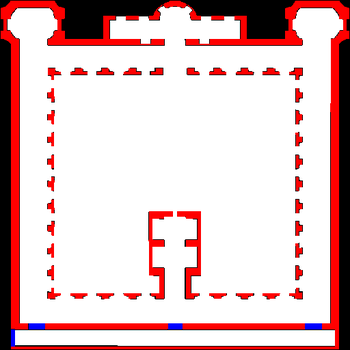}}}{}
\hfill
\jsubfigcent{{\includegraphics[width=2.3cm]{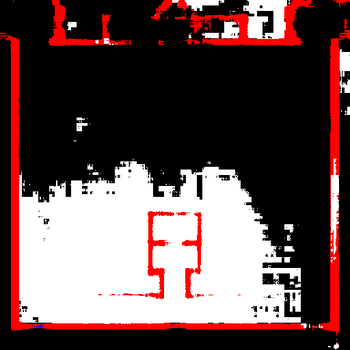}}}{}
\\
\jsubfigcent{{\includegraphics[width=2.3cm]{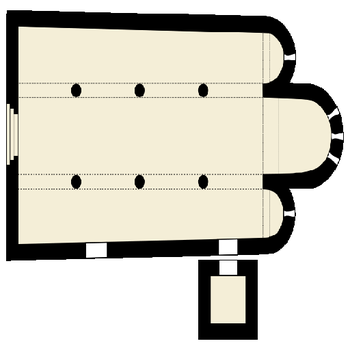}}}{Image}
\hfill
\jsubfigcent{{\includegraphics[width=2.3cm]{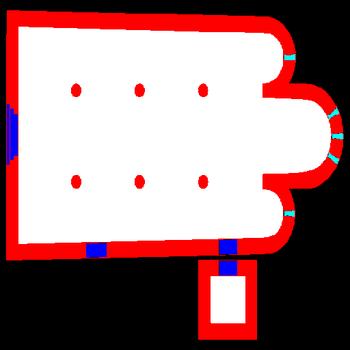}}}{GT}
\hfill
\jsubfigcent{{\includegraphics[width=2.3cm]{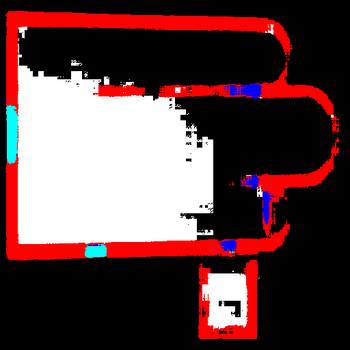}}}{CC5K}
 \vspace{-6pt}
\caption{Benchmark for semantic segmentation (over the {\color{red}{walls}}, {\color{blue}{doors}}, {\color{cyan}{windows}}, interior and background categories) on images from \datasetname{} using the strong supervised CC5K~\cite{kalervo2019cubicasa5k} pretrained model. We can see that our data serves as a challenging benchmark as the model struggles with more diverse and complex floorplans. %
}
    \label{fig:benchmark_segmentation}
\end{figure}

\begin{figure*}
    \centering
\jsubfigcent{{\includegraphics[width=2.7cm]{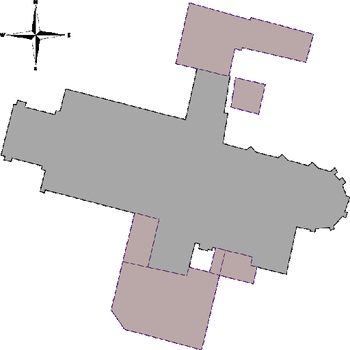}}}{}
\jsubfigcent{{\includegraphics[width=2.7cm]{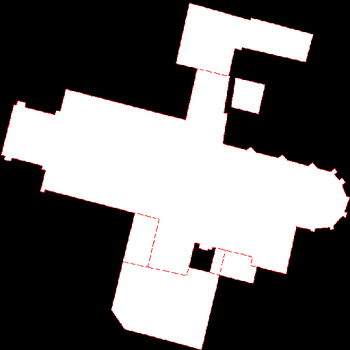}}}{}
\jsubfigcent{{\includegraphics[width=2.7cm]{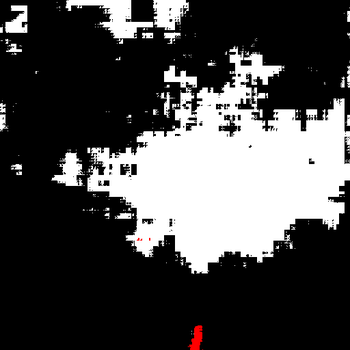}}}{}
\hfill
\jsubfigcent{{\includegraphics[width=2.7cm]{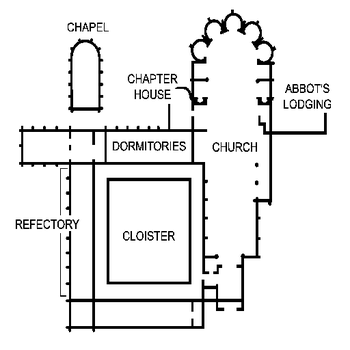}}}{}
\jsubfigcent{{\includegraphics[width=2.7cm]{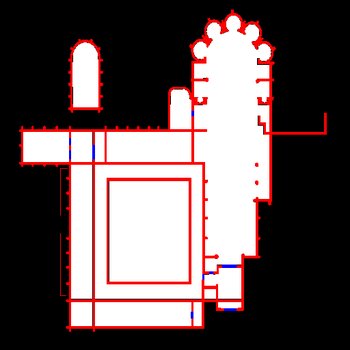}}}{}
\jsubfigcent{{\includegraphics[width=2.7cm]{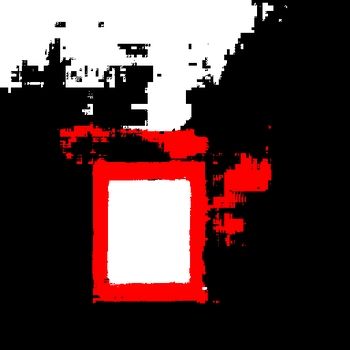}}}{}
\\
\jsubfigcent{{\includegraphics[width=2.7cm]{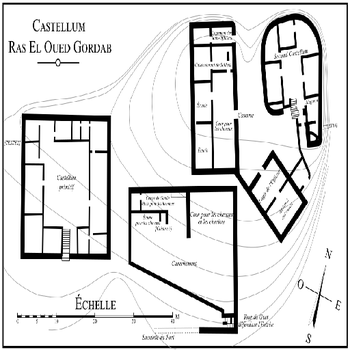}}}{}
\jsubfigcent{{\includegraphics[width=2.7cm]{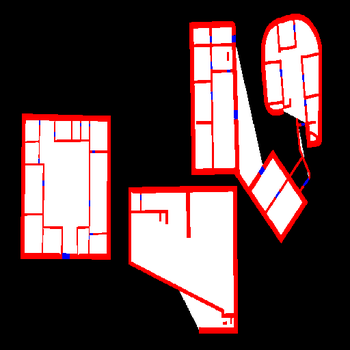}}}{}
\jsubfigcent{{\includegraphics[width=2.7cm]{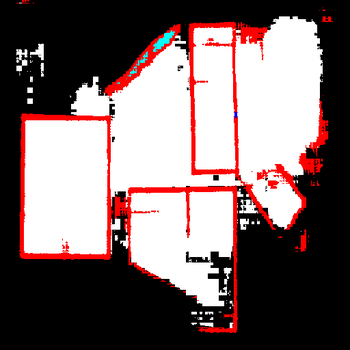}}}{}
\hfill
\jsubfigcent{{\includegraphics[width=2.7cm]{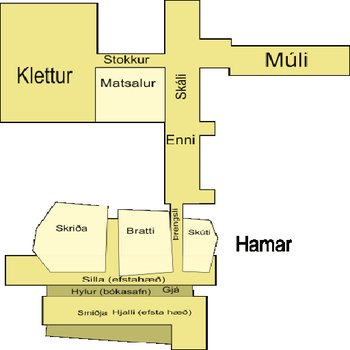}}}{}
\jsubfigcent{{\includegraphics[width=2.7cm]{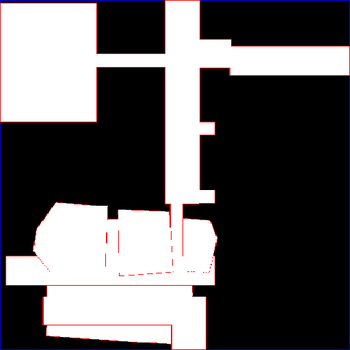}}}{}
\jsubfigcent{{\includegraphics[width=2.7cm]{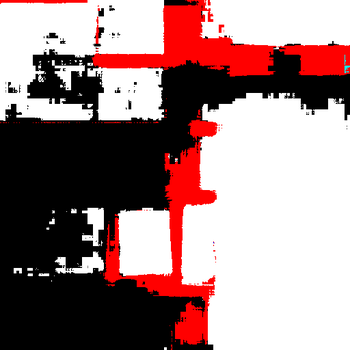}}}{}
\\
\jsubfigcent{{\includegraphics[width=2.7cm]{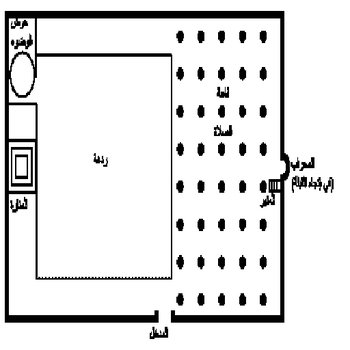}}}{Image}
\jsubfigcent{{\includegraphics[width=2.7cm]{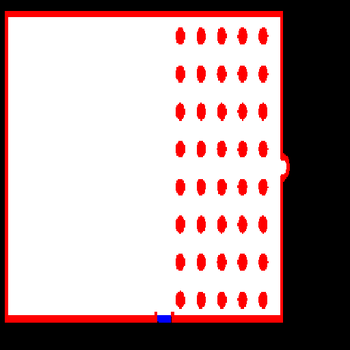}}}{GT}
\jsubfigcent{{\includegraphics[width=2.7cm]{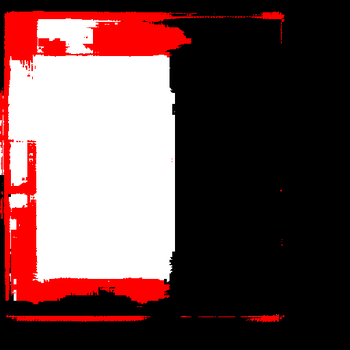}}}{CC5K}
\hfill
\jsubfigcent{{\includegraphics[width=2.7cm]{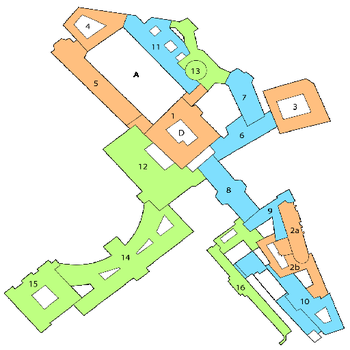}}}{Image}
\jsubfigcent{{\includegraphics[width=2.7cm]{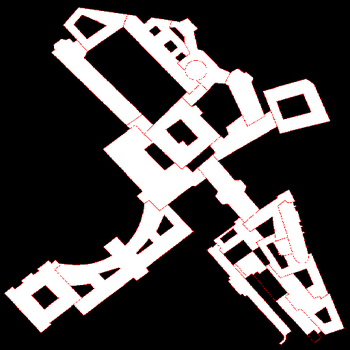}}}{GT}
\jsubfigcent{{\includegraphics[width=2.7cm]{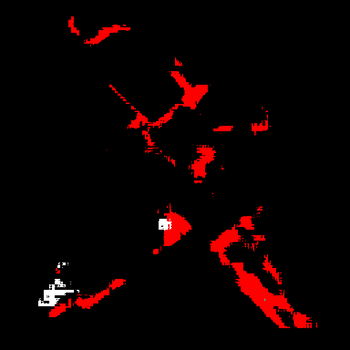}}}{CC5K}
 \vspace{-6pt}
\caption{Additional results on the semantic segmentation benchmark. Images are annotated according to {\color{red}{walls}}, {\color{blue}{doors}}, {\color{cyan}{windows}}, interior and background categories. On the right we show results obtained with CC5K \cite{kalervo2019cubicasa5k}.}
    \label{fig:supp_benchmark_segmentation}
\end{figure*}

We created a benchmark of 110 SVG images, containing wall, windows and door annotations. We included SVG images from our test set. To obtain additional SVG images, we also searched for SVGs that were filtered during the dataset filtering step. Then, we used Inkscape \footnote{https://inkscape.org/} which allowed us to easily annotate full SVG components at once instead of doing it pixel-wise. This made the manual annotation process less tedious and more accurate. 

\subsection{\modified{Wall Segmentation with a Diffusion Model}} \label{sec:supp_wall}

\modified{We apply a diffusion-based architecture to wall segmentation by training ControlNet~\cite{zhang2023adding} using CubiCasa5K~\cite{kalervo2019cubicasa5k} (CC5K) layout maps as the target image to generate and input images as conditions. In particular, we convert CC5K annotations into binary images by denoting walls with black pixels and use these as supervision for binary wall segmentation. We initialize with the \texttt{CompVis/stable-di\\ffusion-v1-4} checkpoint and train for 200K iterations on train items in the CC5K dataset which provides us with 4,200 pairs of images. Other training hyperparameters are the same as those used for ControlNet applied to other tasks as described above. During inference, we input an image (resized to the correct resolution) as a control, generate 25 images with random seeds (and guidance scale 1.0, CFG scale 7.5, 20 inference steps using the default PNDM scheduler). We discretize output pixel values to the closest valid layout color and then use pixel-wise mode values, thus reducing noise from the random nature of each individual generation.}
\label{sec:wall_detection}
\vspace{7pt}

\begin{table}[t]
  \centering
  \setlength{\tabcolsep}{5pt}
  \begin{tabular}{lcc}
     & ResNet & Diffusion\\
    \toprule
    Precision & 0.737 & \textbf{0.746}\\
    Recall & 0.590 & \textbf{0.805}\\
    IoU & 0.488 & \textbf{0.632}\\
    \bottomrule
  \end{tabular}
  \vspace{-7pt}
\captionof{table}{\modified{A comparison between an existing ResNet-based wall detection model (introduced in CC5K \cite{kalervo2019cubicasa5k}) and a Diffusion-model based one (detailed further in Section \ref{sec:wall_detection}), evaluated on our benchmark. We can see the Diffusion-based model outperforms the ResNet-based model across all metrics, suggesting that newer architectures show promise in improving localized knowledge of in-the-wild data, such the floorplans found in \datasetname{}.}
}
\label{tab:wall_detection}
\vspace{-7pt}
\end{table}

\begin{figure}
    \centering
\jsubfigcent{{\includegraphics[width=2cm]{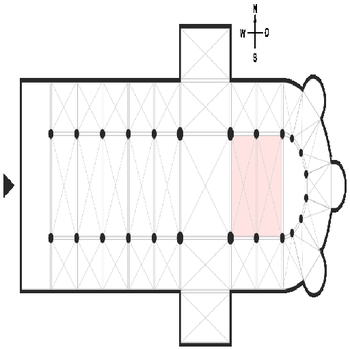}}}{}
\hfill
\jsubfigcent{{\includegraphics[width=2cm]{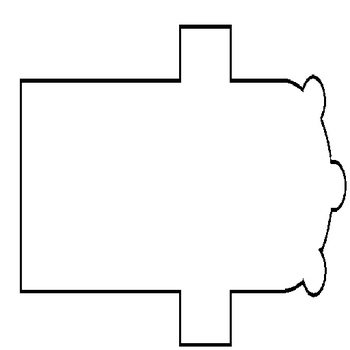}}}{}
\hfill
\jsubfigcent{{\includegraphics[width=2cm]{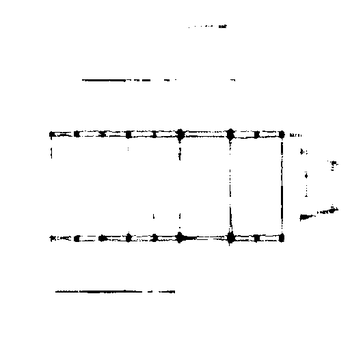}}}{}
\hfill
\jsubfigcent{{\includegraphics[width=2cm]{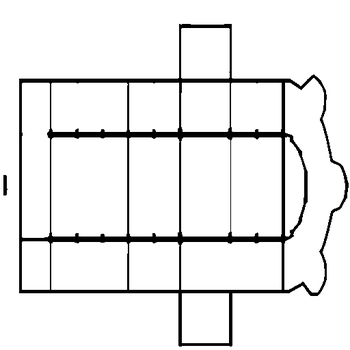}}}{}
\\
\jsubfigcent{{\includegraphics[width=2cm]{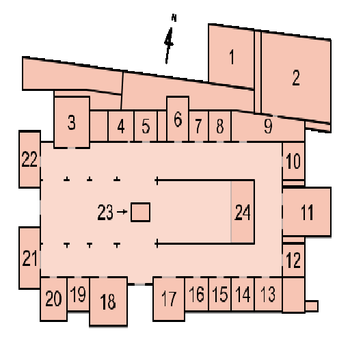}}}{}
\hfill
\jsubfigcent{{\includegraphics[width=2cm]{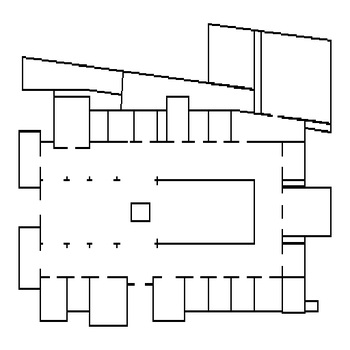}}}{}
\hfill
\jsubfigcent{{\includegraphics[width=2cm]{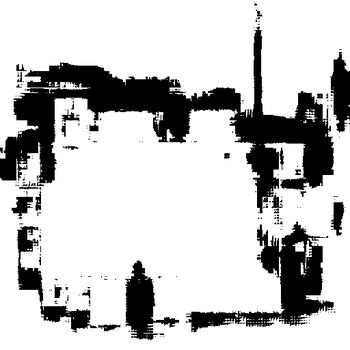}}}{}
\hfill
\jsubfigcent{{\includegraphics[width=2cm]{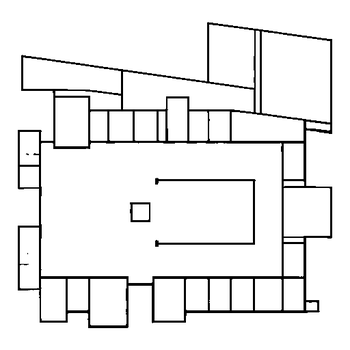}}}{}
\\
\jsubfigcent{{\includegraphics[width=2cm]{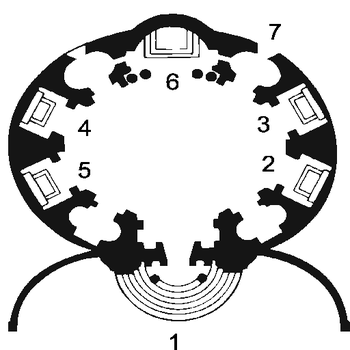}}}{Image \whitetxt{xxxxxxxxx}}
\hfill
\jsubfigcent{{\includegraphics[width=2cm]{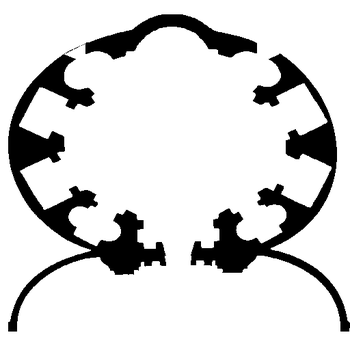}}}{GT \whitetxt{xxxxxxxxx}}
\hfill
\jsubfigcent{{\includegraphics[width=2cm]{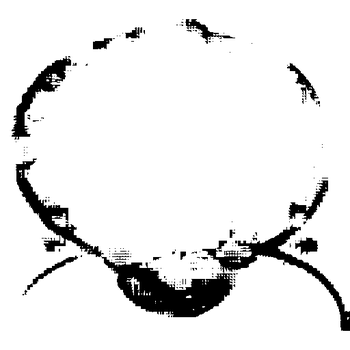}}}{Existing ResNet model}
\hfill
\jsubfigcent{{\includegraphics[width=2cm]{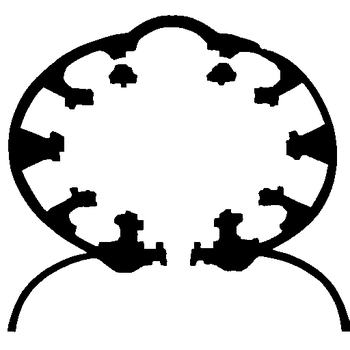}}}{Diffusion model}
\caption{\modified{Wall segmentation results, comparing the CubiCasa5K (CC5K) \cite{kalervo2019cubicasa5k} baseline segmentation model that uses a ResNet backbone to a modified architecture that uses a Diffusion model, as described in Section \ref{sec:wall_detection}. The Diffusion-based model yields refined wall predictions, as illustrated by the examples shown above.} }
    \label{fig:wall_detection}
\end{figure}

\begin{figure*}
    \centering
    \rotatebox{90}{\hspace{-13pt}Transept}
\jsubfigcent{{\includegraphics[width=2.7cm]{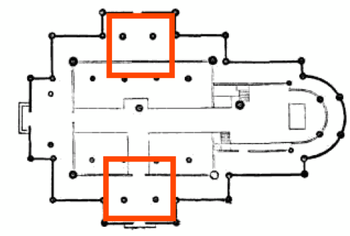}}}{}
\jsubfigcent{{\includegraphics[width=2.7cm]{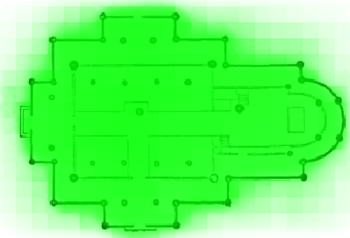}}}{}
\jsubfigcent{{\includegraphics[width=2.7cm]{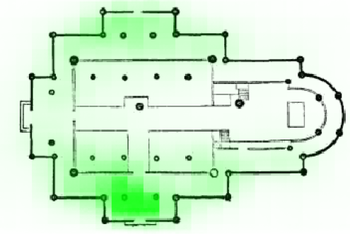}}}{} 
\hfill
    \rotatebox{90}{\hspace{-13pt}Narthex}
\jsubfigcent{{\includegraphics[width=2.7cm]{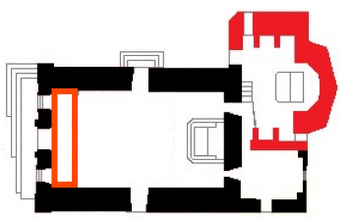}}}{}
\jsubfigcent{{\includegraphics[width=2.7cm]{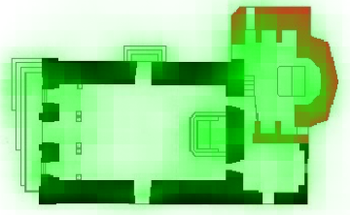}}}{}
\jsubfigcent{{\includegraphics[width=2.7cm]{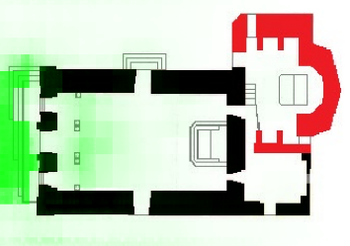}}}{} 
\\
    \rotatebox{90}{\hspace{-10pt}Tower}
\jsubfigcent{{\includegraphics[width=2.7cm]{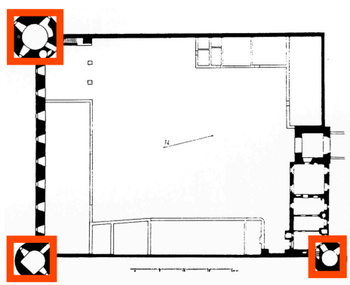}}}{}
\jsubfigcent{{\includegraphics[width=2.7cm]{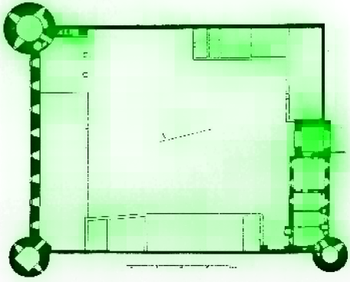}}}{}
\jsubfigcent{{\includegraphics[width=2.7cm]{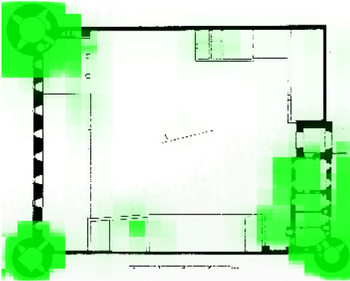}}}{}
\hfill
    \rotatebox{90}{\hspace{-10pt}Court}
\jsubfigcent{{\includegraphics[width=2.7cm]{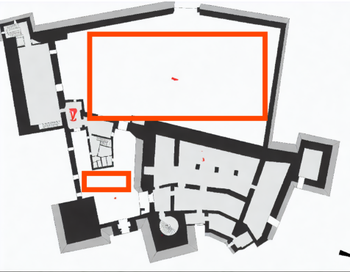}}}{}
\jsubfigcent{{\includegraphics[width=2.7cm]{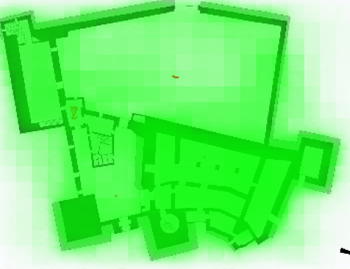}}}{}
\jsubfigcent{{\includegraphics[width=2.7cm]{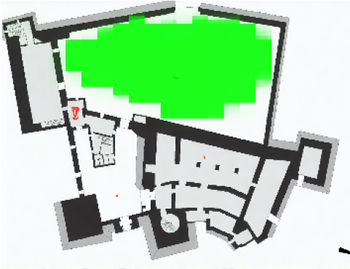}}}{}
\\
    \rotatebox{90}{\hspace{-8pt}Nave}
\jsubfigcent{{\includegraphics[width=2.7cm]{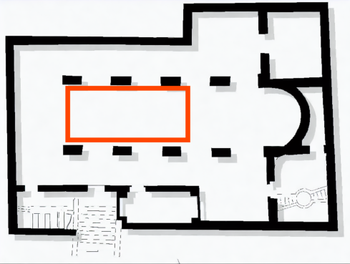}}}{\whitetxt{xxxxxxxxxxxxxxxxxx}}
\jsubfigcent{{\includegraphics[width=2.7cm]{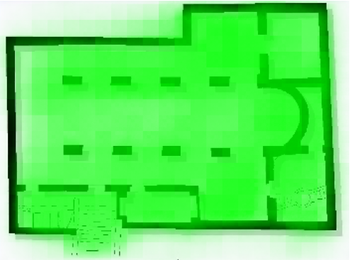}}}{\whitetxt{xxxxxxxxxxxxxxxxxx}}
\jsubfigcent{{\includegraphics[width=2.7cm]{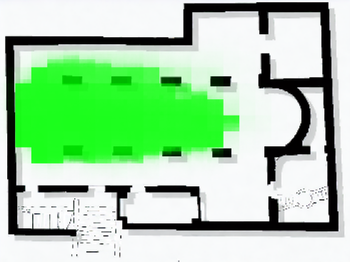}}}{\whitetxt{xxxxxxxxxxxxxxxxxx}}
\hfill
    \rotatebox{90}{\hspace{-9pt}Choir}
\jsubfigcent{{\includegraphics[width=2.7cm]{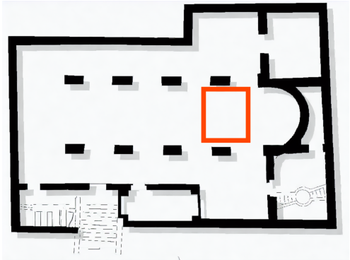}}}{}
\jsubfigcent{{\includegraphics[width=2.7cm]{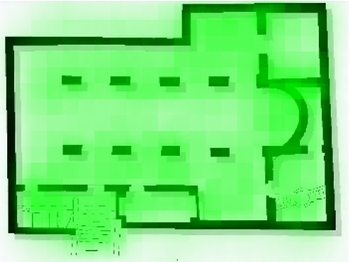}}}{}
\jsubfigcent{{\includegraphics[width=2.7cm]{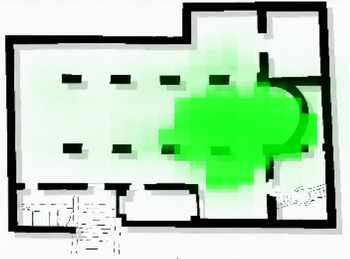}}}{}
\\
    \rotatebox{90}{\hspace{-8pt}Court}
\jsubfigcent{{\includegraphics[width=2.7cm]{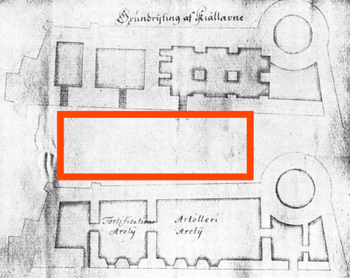}}}{\whitetxt{xxxxxxxxxxxxxxxxxx}}
\jsubfigcent{{\includegraphics[width=2.7cm]{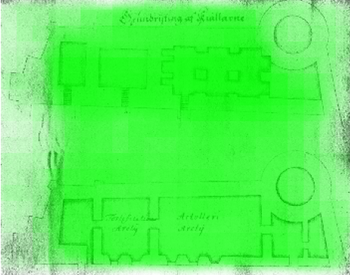}}}{\whitetxt{xxxxxxxxxxxxxxxxxx}}
\jsubfigcent{{\includegraphics[width=2.7cm]{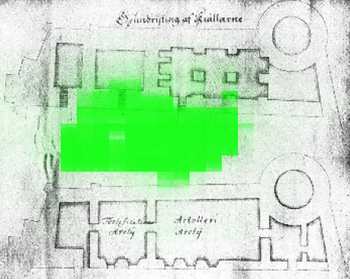}}}{\whitetxt{xxxxxxxxxxxxxxxxxx}}
\hfill
    \rotatebox{90}{\hspace{-9pt}Nave}
\jsubfigcent{{\includegraphics[width=2.7cm]{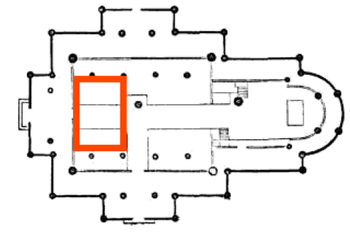}}}{}
\jsubfigcent{{\includegraphics[width=2.7cm]{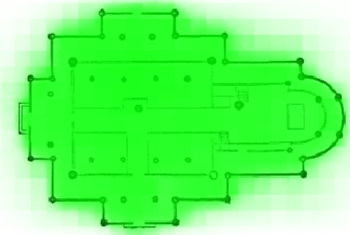}}}{}
\jsubfigcent{{\includegraphics[width=2.7cm]{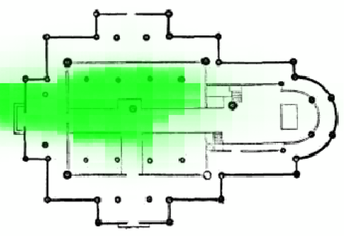}}}{}
\\
    \rotatebox{90}{\hspace{-8pt}Choir}
\jsubfigcent{{\includegraphics[width=2.7cm]{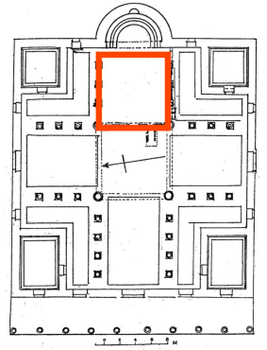}}}{\whitetxt{xxxxxxxxxxxxxxxxxx}}
\jsubfigcent{{\includegraphics[width=2.7cm]{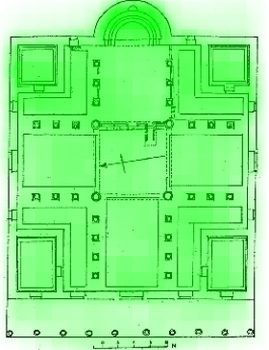}}}{\whitetxt{xxxxxxxxxxxxxxxxxx}}
\jsubfigcent{{\includegraphics[width=2.7cm]{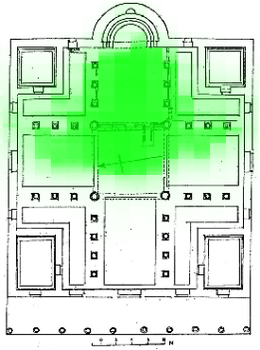}}}{\whitetxt{xxxxxxxxxxxxxxxxxx}}
\hfill
    \rotatebox{90}{\hspace{-9pt}Choir}
\jsubfigcent{{\includegraphics[width=2.7cm]{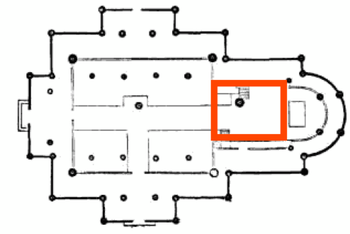}}}{}
\jsubfigcent{{\includegraphics[width=2.7cm]{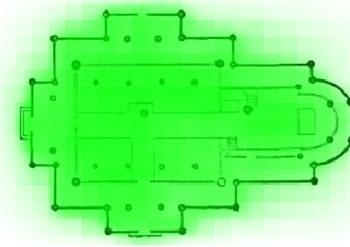}}}{}
\jsubfigcent{{\includegraphics[width=2.7cm]{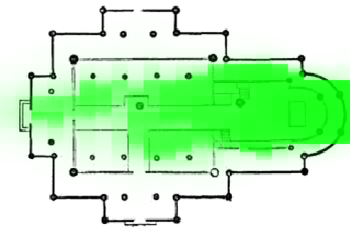}}}{}
\\
    \rotatebox{90}{\hspace{-3pt}Entrance}
\jsubfigcent{{\includegraphics[width=2.7cm]{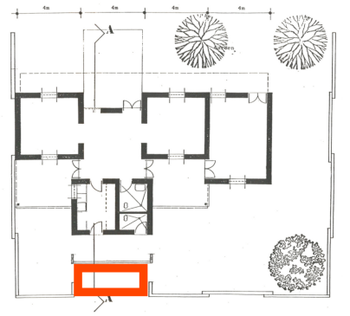}}}{\whitetxt{xxxxxxxxxxxxxxxxxx} \whitetxt{xxxxxxxxxxxxxxxxxx} GT}
\jsubfigcent{{\includegraphics[width=2.7cm]{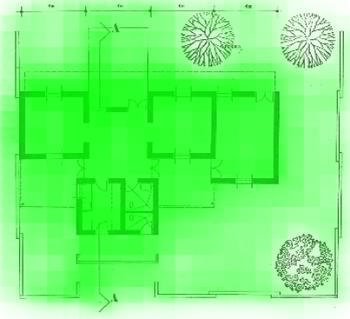}}}{\whitetxt{xxxxxxxxxxxxxxxxxx} \whitetxt{xxxxxxxxxxxxxxxxxx} CLIPSeg}
\jsubfigcent{{\includegraphics[width=2.7cm]{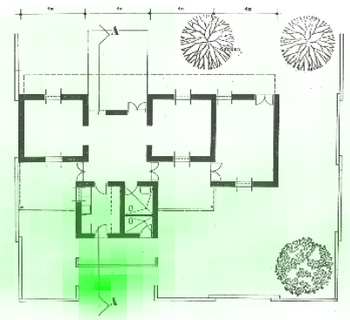}}}{\whitetxt{xxxxxxxxxxxxxxxxxx} \whitetxt{xxxxxxxxxxxxxxxxxx} Ours}
\hfill
    \rotatebox{90}{Court}
\jsubfigcent{{\includegraphics[width=2.7cm]{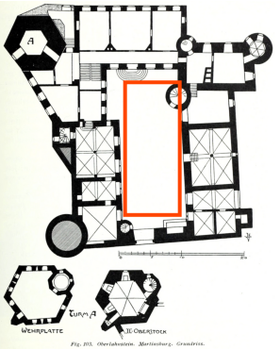}}}{GT}
\jsubfigcent{{\includegraphics[width=2.7cm]{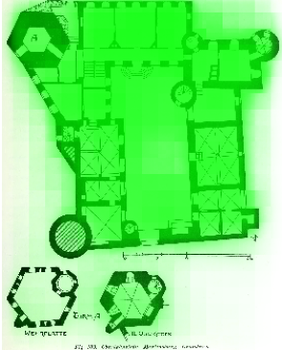}}}{CLIPSeg}
\jsubfigcent{{\includegraphics[width=2.7cm]{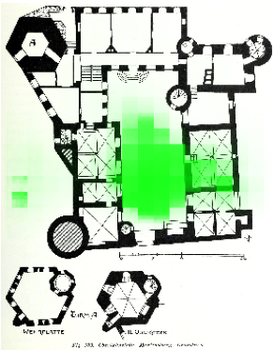}}}{Ours}

\caption{In each column: segmentation results on samples of our test set before (center) and after (right) fine-tuning on our data.}
    \label{fig:segmentation_examples_supp}
\end{figure*}

\begin{figure*}
    \centering
\rotatebox{90}{\hspace{-12pt}Living}
\rotatebox{90}{\hspace{-10pt}room}
\jsubfigcent{{\includegraphics[width=3.5cm]{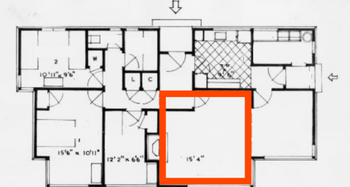}}}{}
\hfill
\jsubfigcent{{\includegraphics[width=3.5cm]{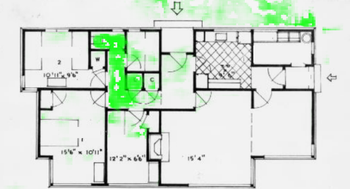}}}{}
\hfill
\jsubfigcent{{\includegraphics[width=3.5cm]{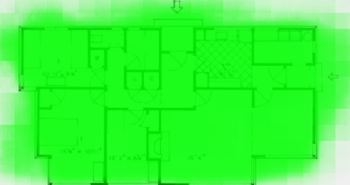}}}{}
\hfill
\jsubfigcent{{\includegraphics[width=3.5cm]{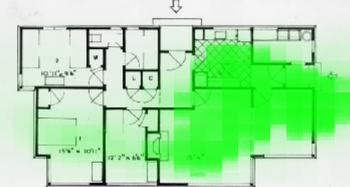}}}{}
\\
\rotatebox{90}{\hspace{-10pt}Kitchen}
\rotatebox{90}{\hspace{-30pt}\whitetxt{room}}
\jsubfigcent{{\includegraphics[width=3.5cm]{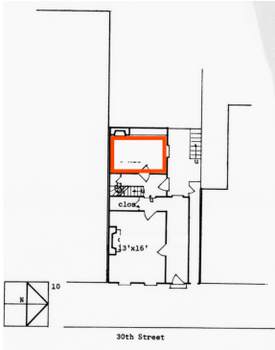}}}{}
\hfill
\jsubfigcent{{\includegraphics[width=3.5cm]{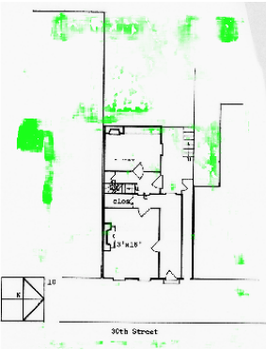}}}{}
\hfill
\jsubfigcent{{\includegraphics[width=3.5cm]{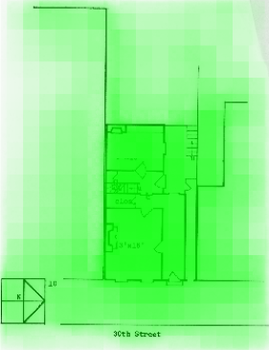}}}{}
\hfill
\jsubfigcent{{\includegraphics[width=3.5cm]{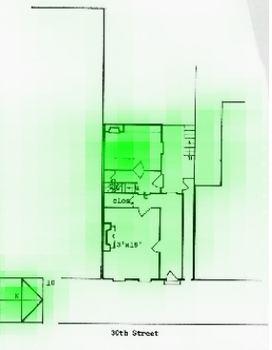}}}{}
\\
\rotatebox{90}{\hspace{-10pt}Porch$^*$}
\rotatebox{90}{\hspace{-30pt}\whitetxt{room}}
\jsubfigcent{{\includegraphics[width=3.5cm]{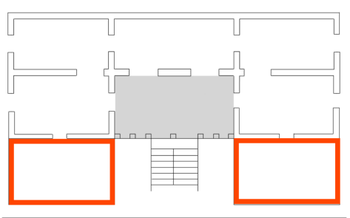}}}{}
\hfill
\jsubfigcent{{\includegraphics[width=3.5cm]{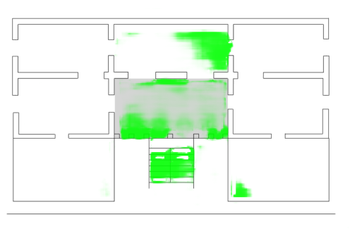}}}{}
\hfill
\jsubfigcent{{\includegraphics[width=3.5cm]{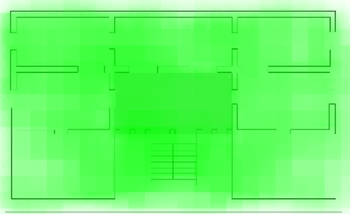}}}{}
\hfill
\jsubfigcent{{\includegraphics[width=3.5cm]{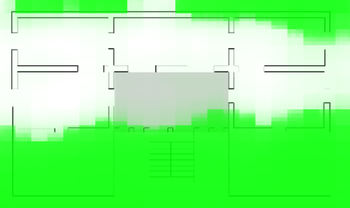}}}{}
\\
\rotatebox{90}{\hspace{-30pt}\whitetxt{room}}
\rotatebox{90}{\hspace{-30pt}\whitetxt{room}}
\jsubfigcent{{\includegraphics[width=3.5cm]{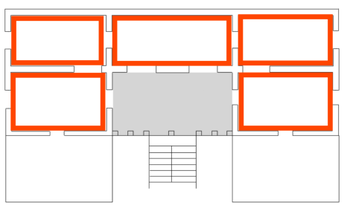}}}{}
\hfill
\jsubfigcent{{\includegraphics[width=3.5cm]{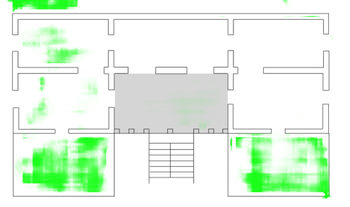}}}{}
\hfill
\jsubfigcent{{\includegraphics[width=3.5cm]{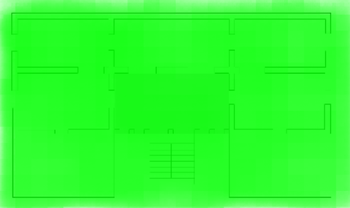}}}{}
\hfill
\jsubfigcent{{\includegraphics[width=3.5cm]{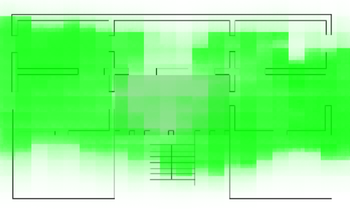}}}{}
\\
\rotatebox{90}{\hspace{-30pt}$\longleftarrow$Bedroom$\longrightarrow$}
\rotatebox{90}{\hspace{-30pt}\whitetxt{room}}
\jsubfigcent{{\includegraphics[width=3.5cm]{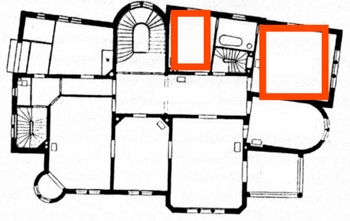}}}{}
\hfill
\jsubfigcent{{\includegraphics[width=3.5cm]{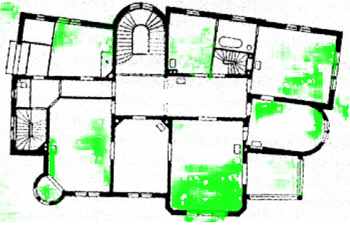}}}{}
\hfill
\jsubfigcent{{\includegraphics[width=3.5cm]{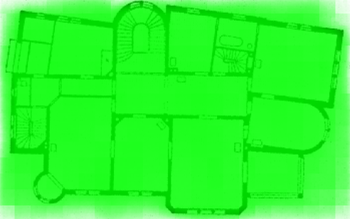}}}{}
\hfill
\jsubfigcent{{\includegraphics[width=3.5cm]{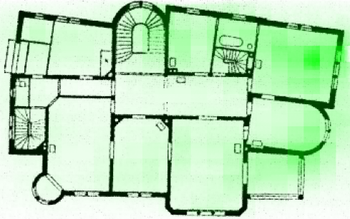}}}{}
\\
\rotatebox{90}{\hspace{-40pt}\whitetxt{room}}
\rotatebox{90}{\hspace{-30pt}\whitetxt{room}}
\jsubfigcent{{\includegraphics[width=3.5cm]{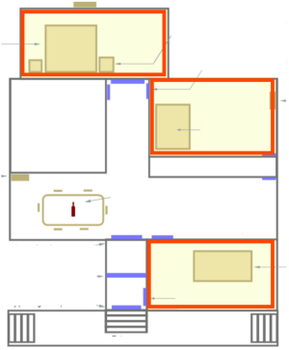}}}{Image w/ GT}
\hfill
\jsubfigcent{{\includegraphics[width=3.5cm]{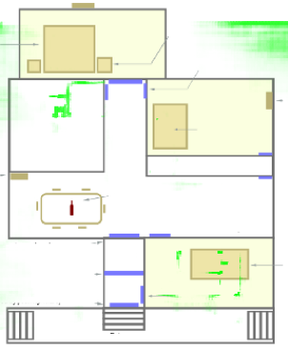}}}{CC5K \cite{kalervo2019cubicasa5k}}
\hfill
\jsubfigcent{{\includegraphics[width=3.5cm]{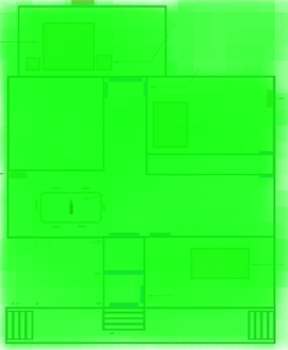}}}{CLIPSeg \cite{luddecke2022image}}
\hfill
\jsubfigcent{{\includegraphics[width=3.5cm]{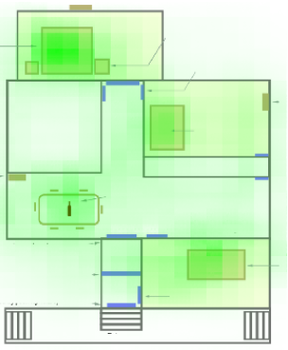}}}{Ours}
\vspace{-6pt}
{\begin{flushleft} {\scriptsize $^*$Our \emph{porch} results correspond to the CubiCasa5K \emph{outdoor} category.} \end{flushleft}}
\vspace{-12pt}
\caption{Additional comparisons of our segmentation probability map results on residential buildings with the strongly-supervised CubiCasa5K (CC5K) model~\cite{kalervo2019cubicasa5k}.}
    \label{fig:segmentation_vs_cc5k}
\end{figure*}
\vspace{-6pt}

\begin{figure*}
    \centering
    \setlength\tabcolsep{0.1pt}
    \begin{tabular}{c@{\hspace{0.3cm}}c@{\hspace{0.3cm}}ccc}
        & pretrained & \multicolumn{3}{c}{$\longleftarrow$ fine-tuned $\longrightarrow$} \\
      \rotatebox{90}{ \whitetxt{ixxxx} Cathedral}
      & \includegraphics[height=3.4cm]{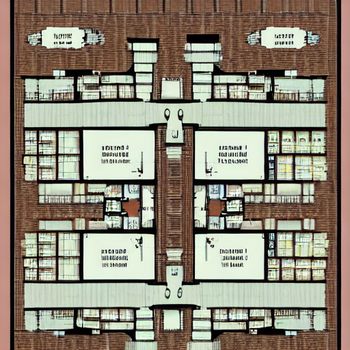} & 
        \includegraphics[height=3.4cm]{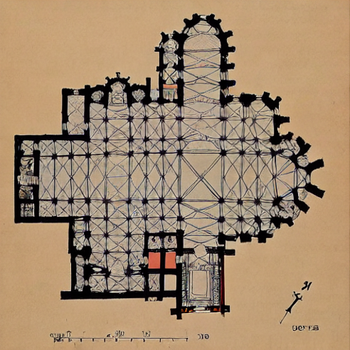} &
        \includegraphics[height=3.4cm]{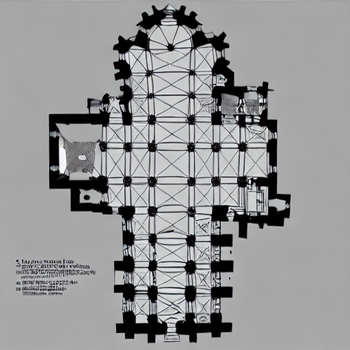} &
        \includegraphics[height=3.4cm]{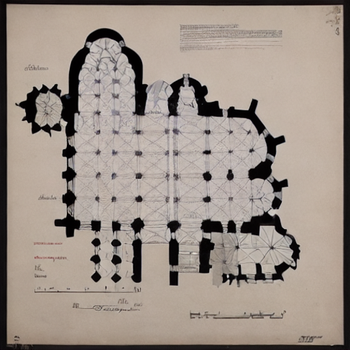} \\
        \rotatebox{90}{\whitetxt{xxxxxx} Theater} 
        & \includegraphics[height=3.4cm]{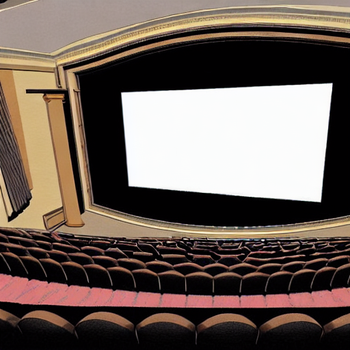} & 
        \includegraphics[height=3.4cm]{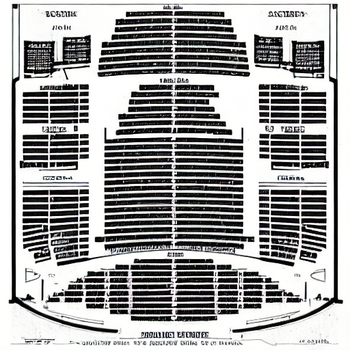} &
        \includegraphics[height=3.4cm]{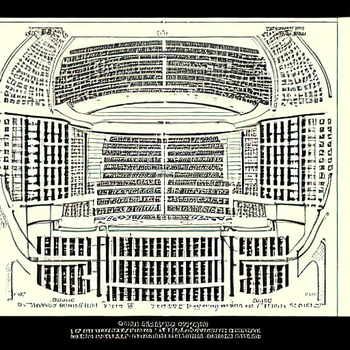} &
        \includegraphics[height=3.4cm]{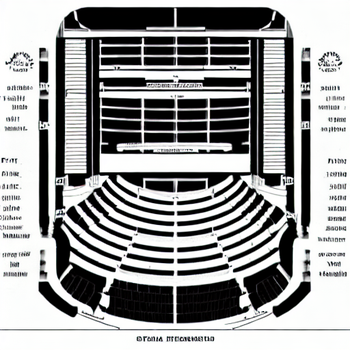} \\
        \rotatebox{90}{\whitetxt{xxxxxxx} Palace}
        & \includegraphics[height=3.4cm]{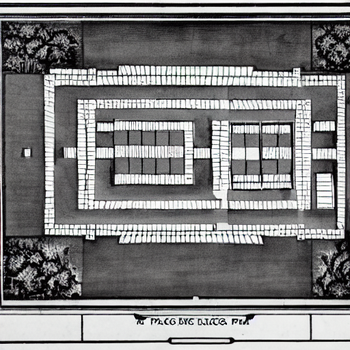} & 
        \includegraphics[height=3.4cm]{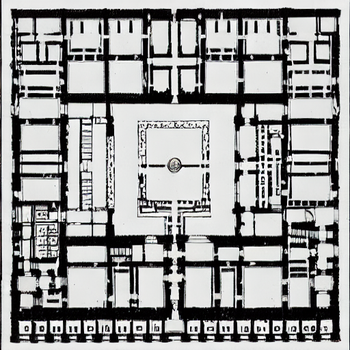} &
        \includegraphics[height=3.4cm]{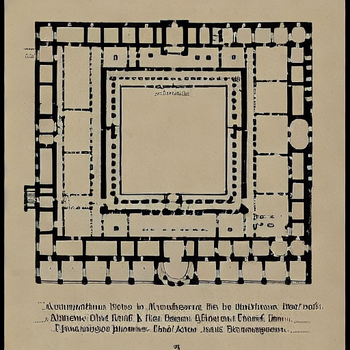} &
        \includegraphics[height=3.4cm]{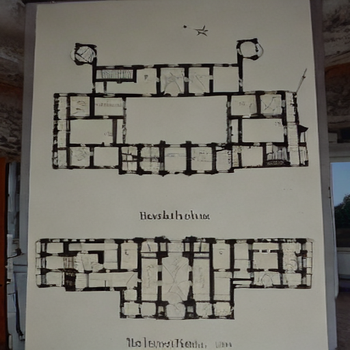} \\
        \rotatebox{90}{\whitetxt{xx}Residential Building}
        & \includegraphics[height=3.4cm]{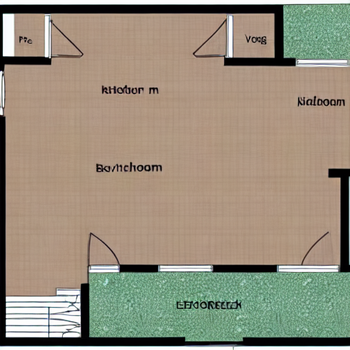} & 
        \includegraphics[height=3.4cm]{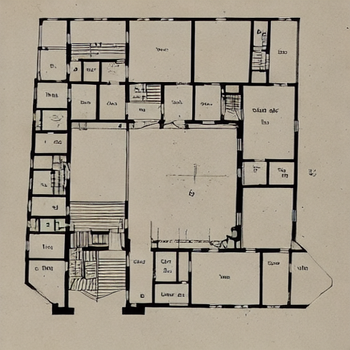} &
        \includegraphics[height=3.4cm]{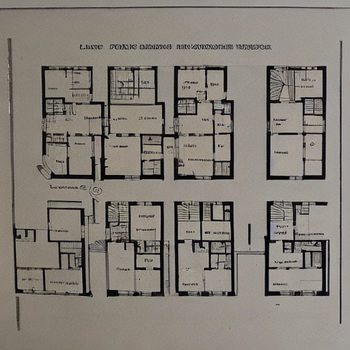} &
        \includegraphics[height=3.4cm]{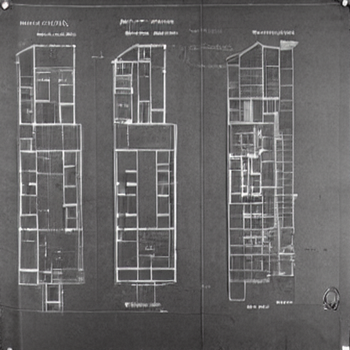} \\
        \rotatebox{90}{\whitetxt{xxxxx} Museum}
        & \includegraphics[height=3.4cm]{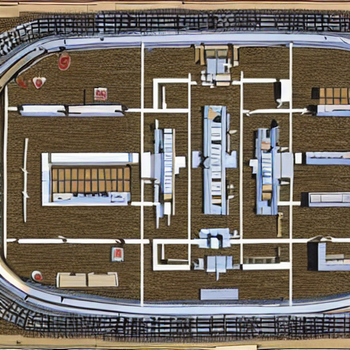} & 
        \includegraphics[height=3.4cm]{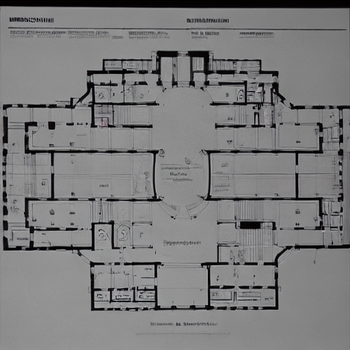} &
        \includegraphics[height=3.4cm]{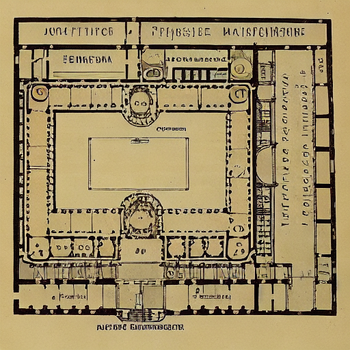} &
        \includegraphics[height=3.4cm]{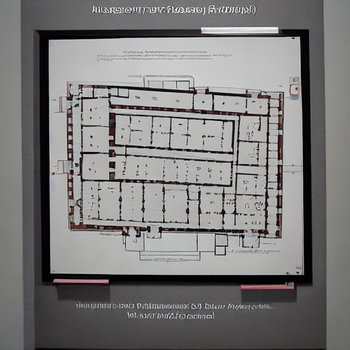} \\
        \rotatebox{90}{\whitetxt{xxxxxx} Hospital}
        & \includegraphics[height=3.4cm]{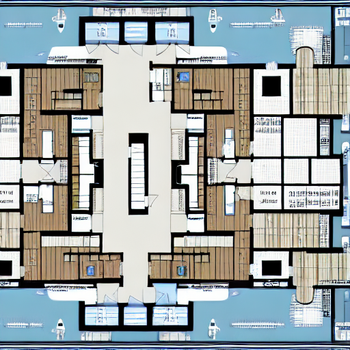} & 
        \includegraphics[height=3.4cm]{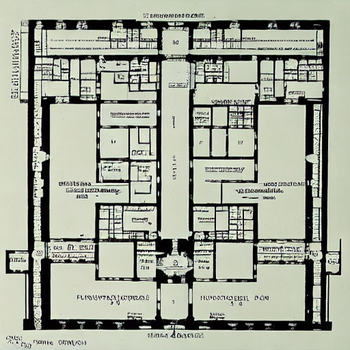} &
        \includegraphics[height=3.4cm]{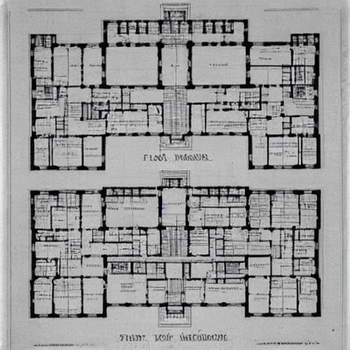} &
        \includegraphics[height=3.4cm]{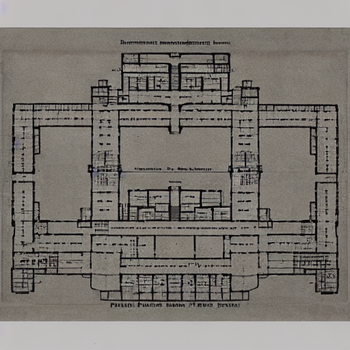} \\
        
  \end{tabular}   
\vspace{-8pt}
\caption{Additional generated floorplans, showing diverse building types (provided on the left). The left-most column shows samples from  the pretrained SD model and rest of the columns showcase the results from our fine-tuned model.}
\label{fig:generated_examples_sup}
\end{figure*}

\begin{figure*}
    \centering
    \setlength\tabcolsep{0.1pt}
    \begin{tabular}{ccccc}
        Input Images & Extracted Masks & House & Castle & Library \\
        \includegraphics[height=3.35cm]{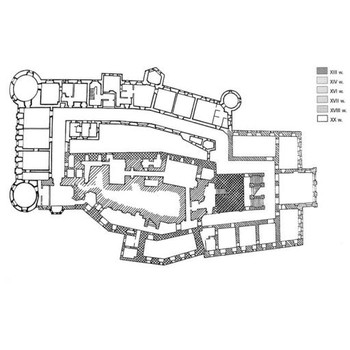} & \includegraphics[height=3.35cm]{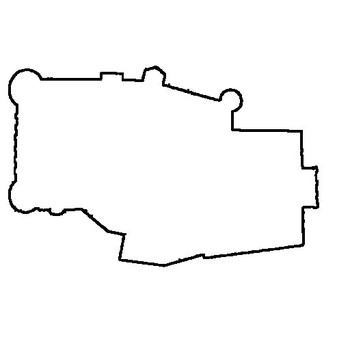} &
        \includegraphics[height=3.35cm]{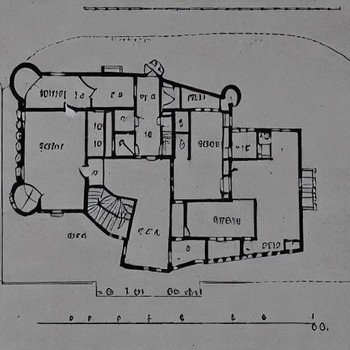} &
        \includegraphics[height=3.35cm]{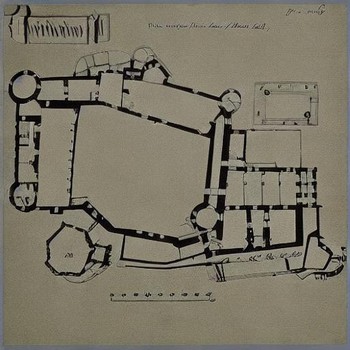} &
        \includegraphics[height=3.35cm]{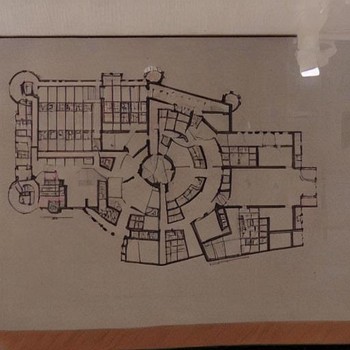} \\
        \includegraphics[height=3.35cm]{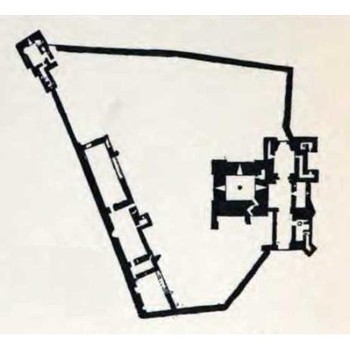} & \includegraphics[height=3.35cm]{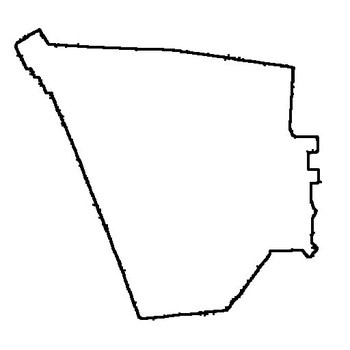} &
        \includegraphics[height=3.35cm]{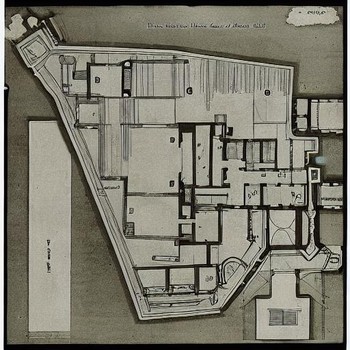} &
        \includegraphics[height=3.35cm]{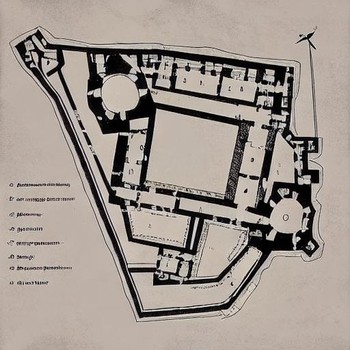} &
        \includegraphics[height=3.35cm]{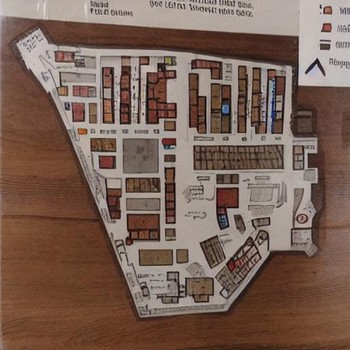}\\   \includegraphics[height=3.35cm]{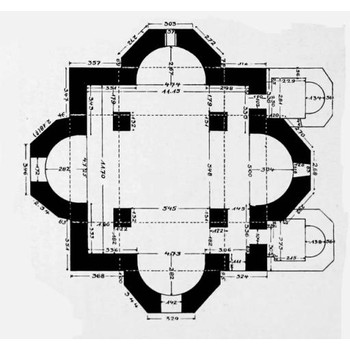} & \includegraphics[height=3.35cm]{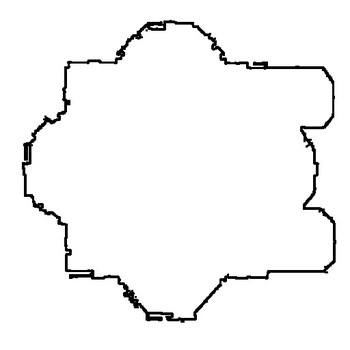} &
        \includegraphics[height=3.35cm]{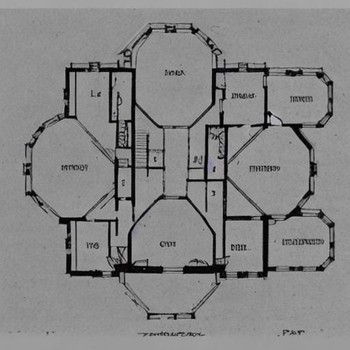} &
        \includegraphics[height=3.35cm]{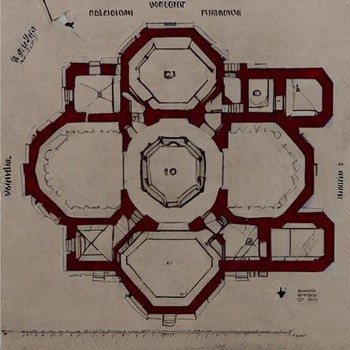} &
        \includegraphics[height=3.35cm]{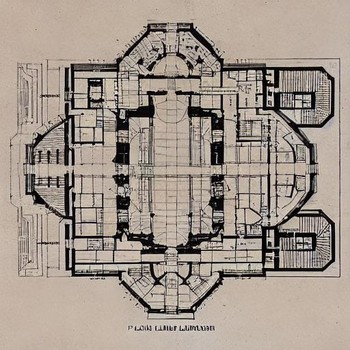}\\
        Input Images & Extracted Masks & Museum & Hospital & Hotel \\
        \includegraphics[height=3.35cm]{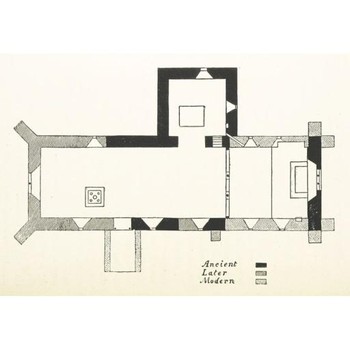} & \includegraphics[height=3.35cm]{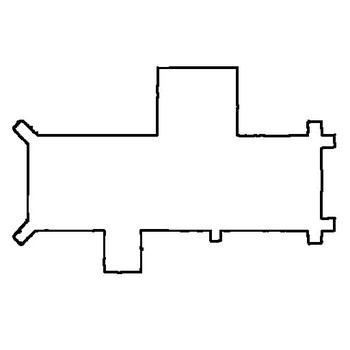} &
        \includegraphics[height=3.35cm]{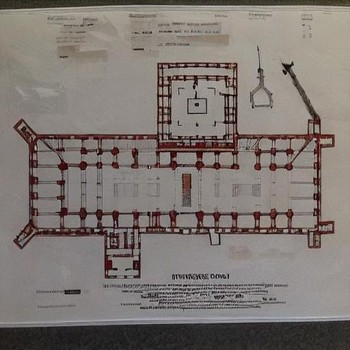} &
        \includegraphics[height=3.35cm]{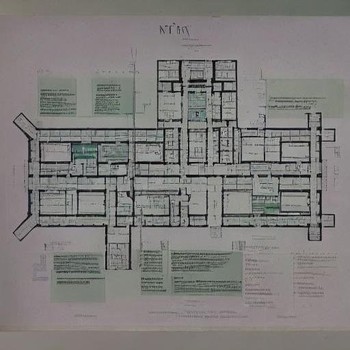} &
        \includegraphics[height=3.35cm]{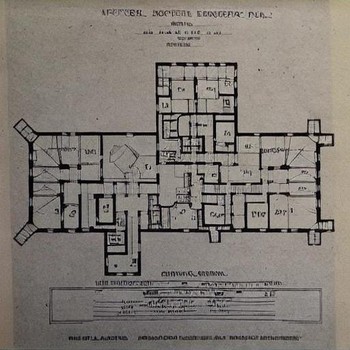} \\
        \includegraphics[height=3.35cm]{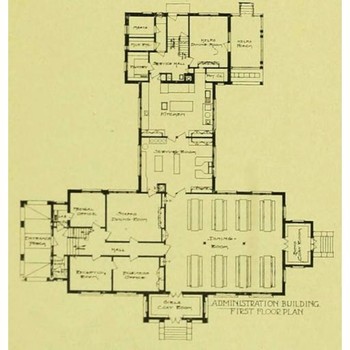} & \includegraphics[height=3.35cm]{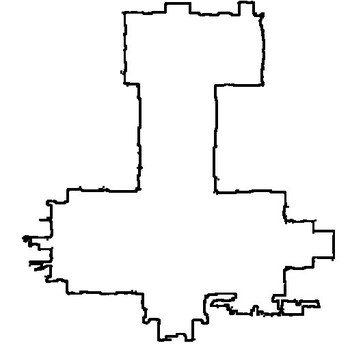} &
        \includegraphics[height=3.35cm]{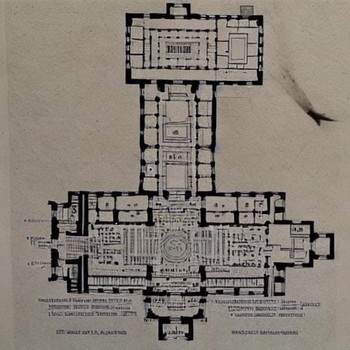} &
        \includegraphics[height=3.35cm]{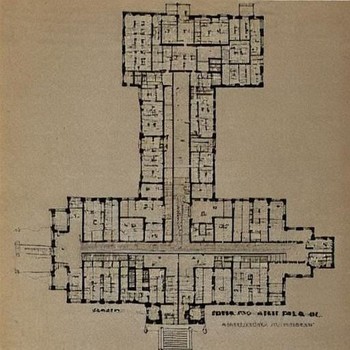} &
        \includegraphics[height=3.35cm]{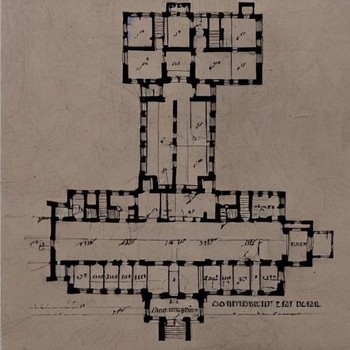} \\
        \includegraphics[height=3.35cm]{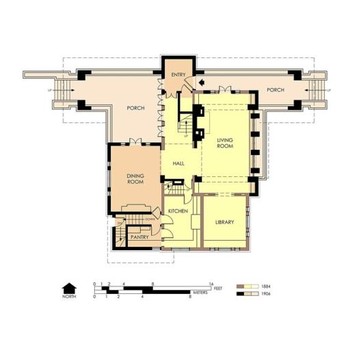} & \includegraphics[height=3.35cm]{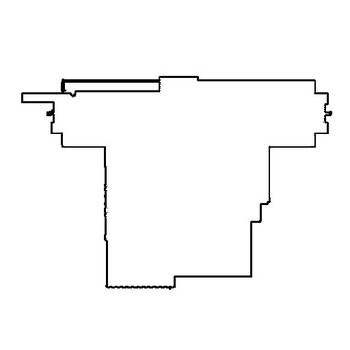} &
        \includegraphics[height=3.35cm]{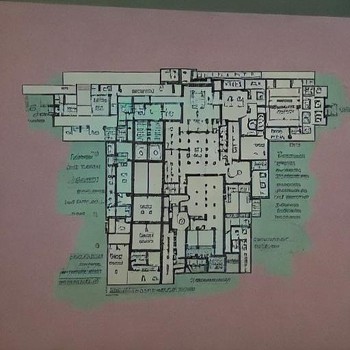} &
        \includegraphics[height=3.35cm]{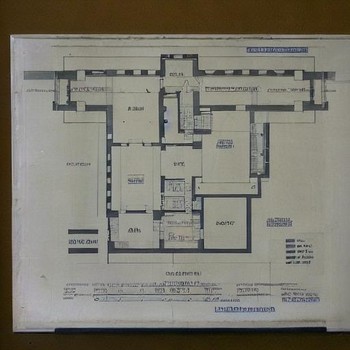} &
        \includegraphics[height=3.35cm]{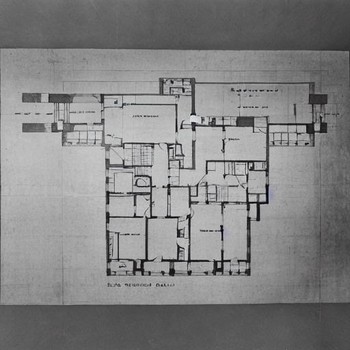} \\
        
     \end{tabular}
    \vspace{-10pt}
    \caption{Additional results for boundary-conditioned generation, showing a variety of shapes (shown on the left) and building types (shown on top).}
    \label{fig:controlnet_ex_sup}
\end{figure*}

\begin{figure*}
    \centering
    \setlength\tabcolsep{0.2pt}
    \begin{tabular}{c@{\hspace{0.3cm}}ccccc}
         & Condition & CS 0.0 & CS 0.25 & \textbf{CS 0.5} & CS 1.0 \\
        \rotatebox{90}{ \whitetxt{ixxxxxx} School} & \includegraphics[height=3.22cm]{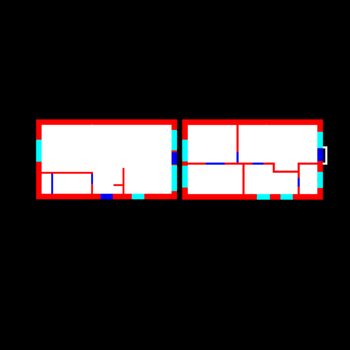} &
        \includegraphics[height=3.22cm]{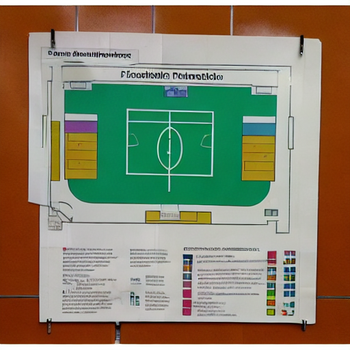} 
        &
        \includegraphics[height=3.22cm]{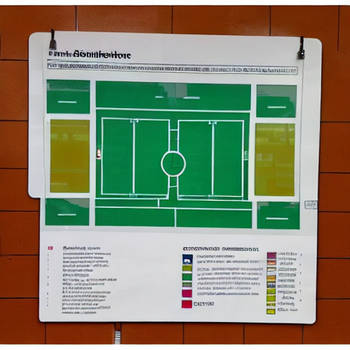} &
        \includegraphics[height=3.22cm]{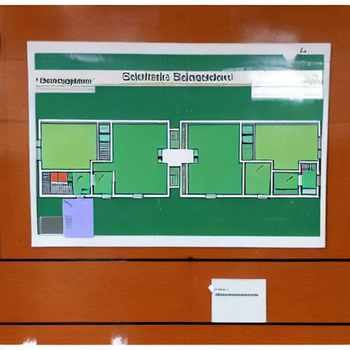} &
        \includegraphics[height=3.22cm]{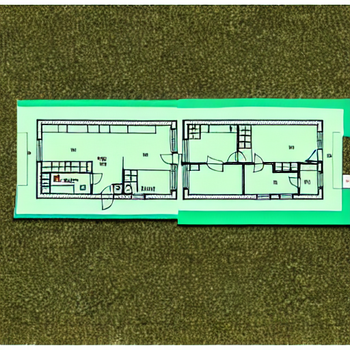} \\
        \rotatebox{90}{ \whitetxt{ixxxxx} Cathedral} & \includegraphics[height=3.22cm]{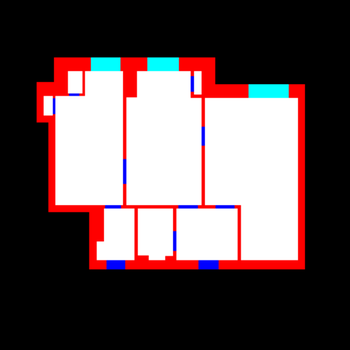} &
        \includegraphics[height=3.22cm]{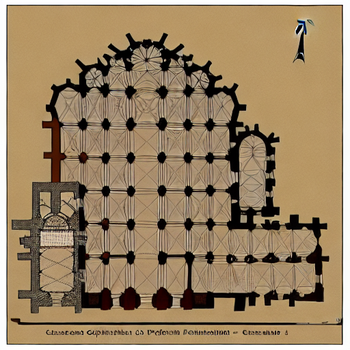} 
        &
        \includegraphics[height=3.22cm]{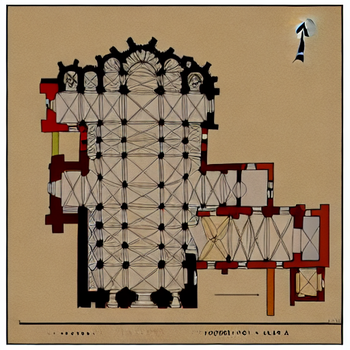} &
        \includegraphics[height=3.22cm]{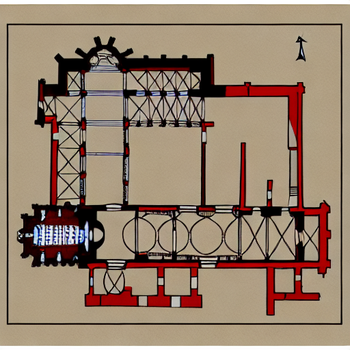} &
        \includegraphics[height=3.22cm]{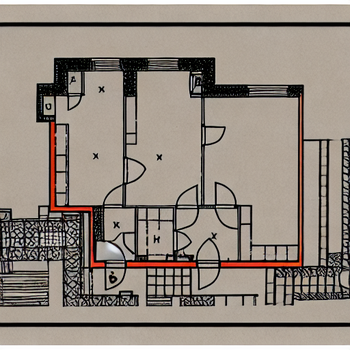} \\
        \rotatebox{90}{ \whitetxt{xxxxx} Mausoleum} & \includegraphics[height=3.22cm]{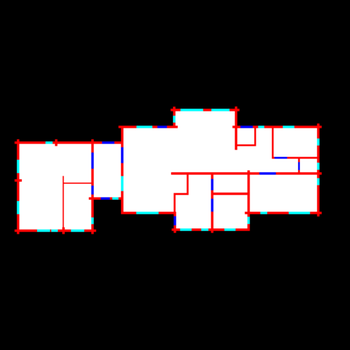} &
        \includegraphics[height=3.22cm]{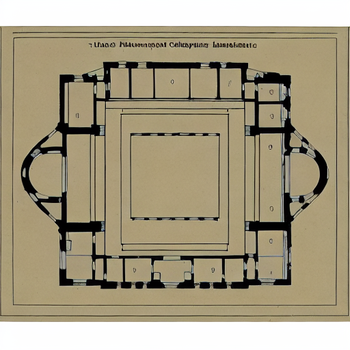} 
        &
        \includegraphics[height=3.22cm]{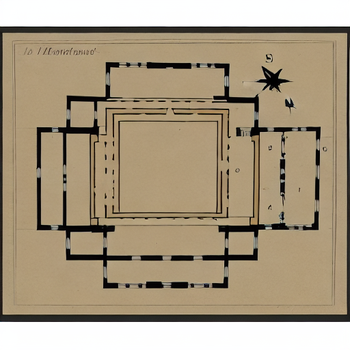} &
        \includegraphics[height=3.22cm]{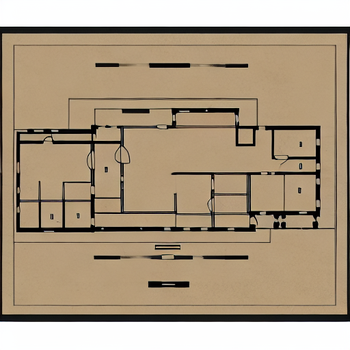} &
        \includegraphics[height=3.22cm]{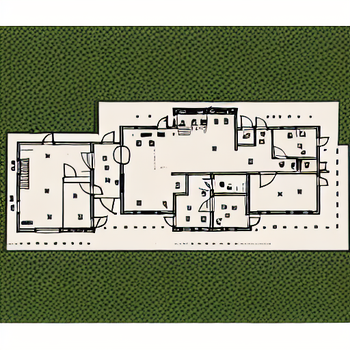} \\
        & Condition & CFG 7.5 & CFG 10 & \textbf{CFG 15} & CFG 25 \\
        \rotatebox{90}{ \whitetxt{Xxxxx} Apartment} & \includegraphics[height=3.22cm]{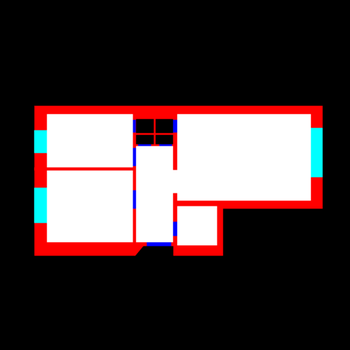} &
        \includegraphics[height=3.22cm]{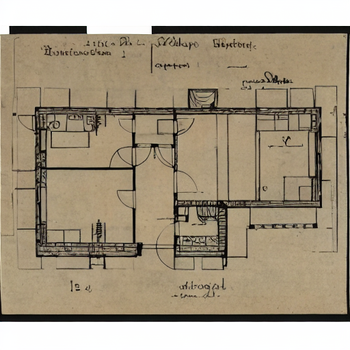} 
        &
        \includegraphics[height=3.22cm]{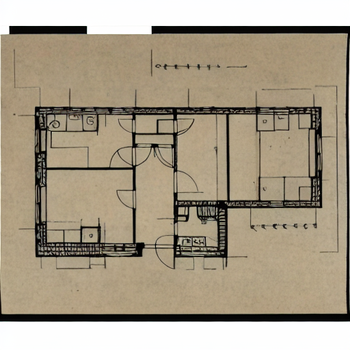} &
        \includegraphics[height=3.22cm]{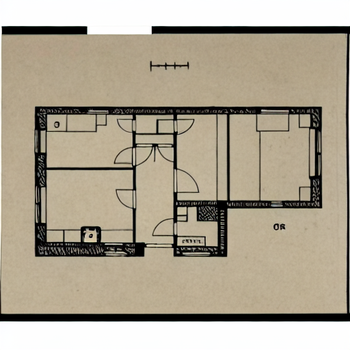} &
        \includegraphics[height=3.22cm]{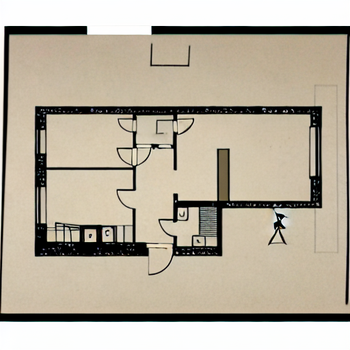} \\
        \rotatebox{90}{ \whitetxt{xxxxxxx} Castle} & \includegraphics[height=3.22cm]{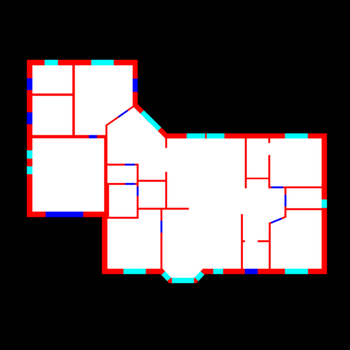} &
        \includegraphics[height=3.22cm]{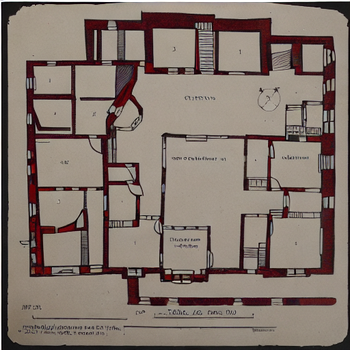} 
        &
        \includegraphics[height=3.22cm]{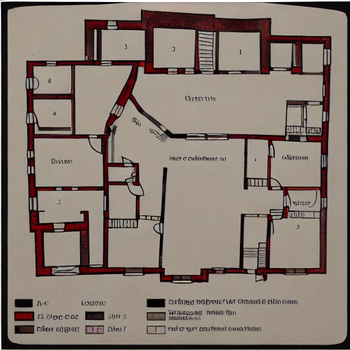} &
        \includegraphics[height=3.22cm]{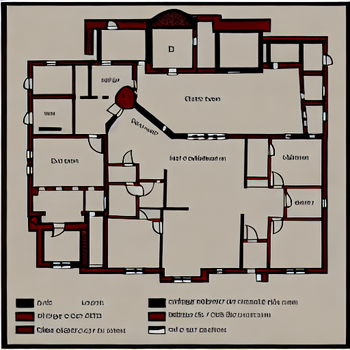} &
        \includegraphics[height=3.22cm]{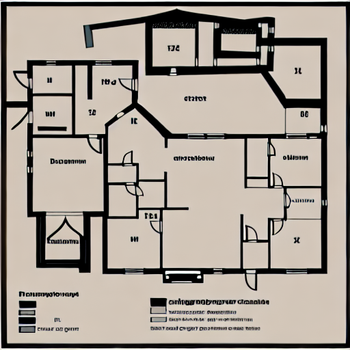} \\
        \rotatebox{90}{ \whitetxt{xxxxxx} Factory} & \includegraphics[height=3.22cm]{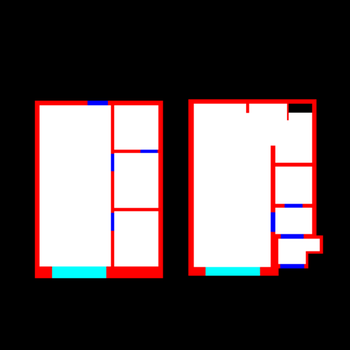} &
        \includegraphics[height=3.22cm]{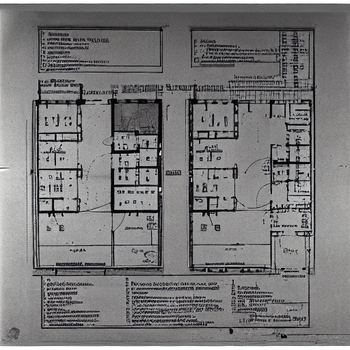} 
        &
        \includegraphics[height=3.22cm]{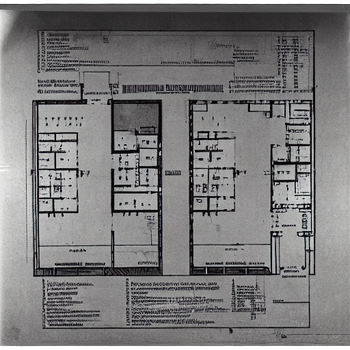} &
        \includegraphics[height=3.22cm]{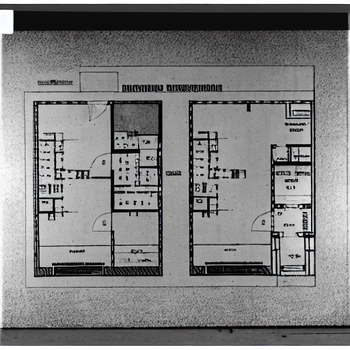} &
        \includegraphics[height=3.22cm]{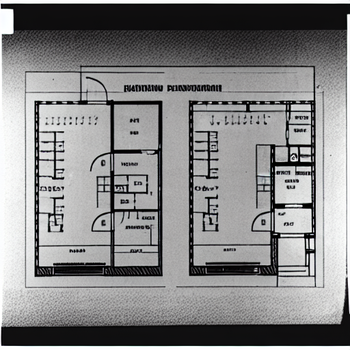} \\
     \end{tabular}
    \vspace{-10pt}
    \caption{Additional results for structure-conditioned generation, showing the effect of changing condition scale (CS) and CFG scales during inference (with a fixed seed). The condition scale controls the trade-off between adherence to the structure condition and avoiding leakage of the CubiCasa5K style which ControlNet was exposed to in fine-tuning. We also find a relatively high CFG value to improve image quality. Chosen values for inference are in \textbf{bold}.}
    \label{fig:struct_ex_sup}
\end{figure*}

\begin{figure*}
    \centering
    \setlength\tabcolsep{0.1pt}
    \begin{tabular}{c@{\hspace{0.3cm}}cc@{\hspace{0.5cm}}c@{\hspace{0.3cm}}cc}
      \rotatebox{90}{ \whitetxt{xxxxxx} Church}
      & \includegraphics[height=3.7cm, width=3.7cm]{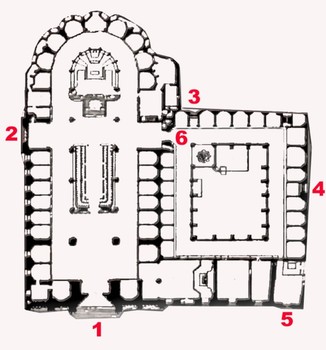} & 
        \includegraphics[height=3.7cm, width=3.7cm]{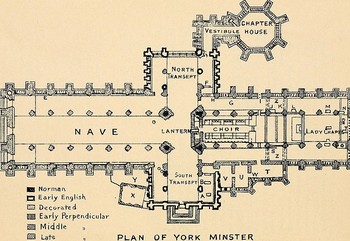} &
        \rotatebox{90}{ \whitetxt{xxxxxx} Hospital} &
        \includegraphics[height=3.7cm, width=3.7cm]{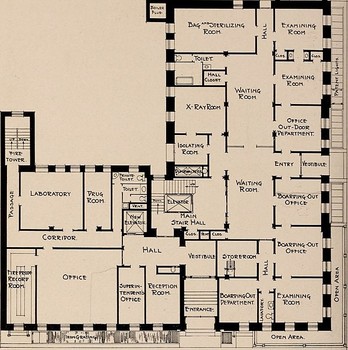} &
        \includegraphics[height=3.7cm, width=3.7cm]{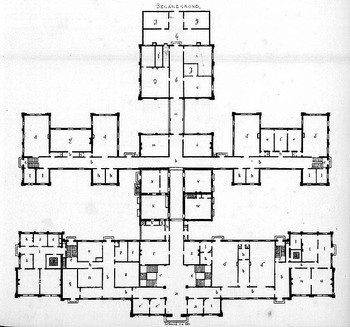} \\
        \rotatebox{90}{\whitetxt{xxxxxxx} Theater} 
        & \includegraphics[height=3.7cm, width=3.7cm]{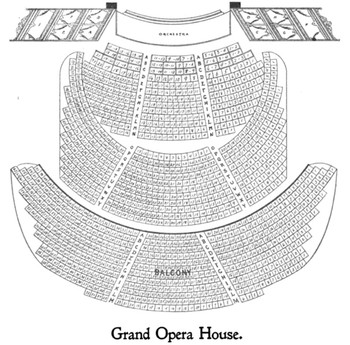} & 
        \includegraphics[height=3.7cm, width=3.7cm]{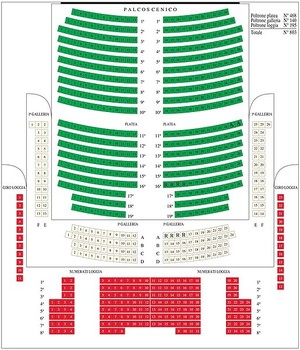} &
        \rotatebox{90}{ \whitetxt{xxxxxxx} Castle} &
        \includegraphics[height=3.7cm, width=3.7cm]{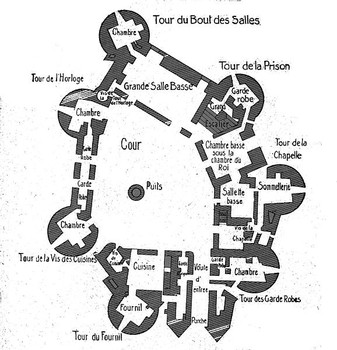} &
        \includegraphics[height=3.7cm, width=3.7cm]{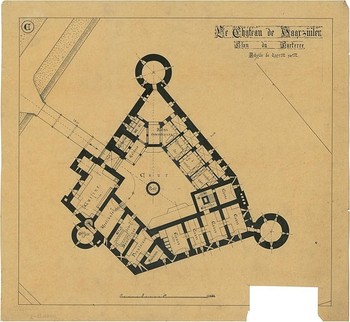} \\
        \rotatebox{90}{\whitetxt{xxxxxxxx} Library}
        & \includegraphics[height=3.7cm, width=3.7cm]{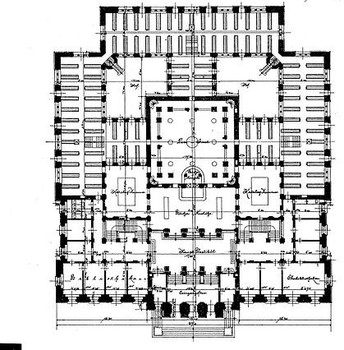} & 
        \includegraphics[height=3.7cm, width=3.7cm]{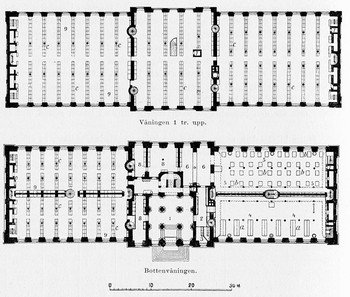} &
        \rotatebox{90}{ \whitetxt{xxxxxxx} Hotel} &
        \includegraphics[height=3.7cm, width=3.7cm]{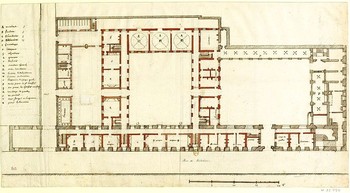} &
        \includegraphics[height=3.7cm, width=3.7cm]{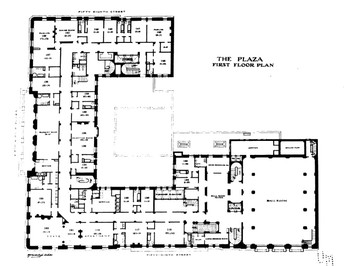} \\
        \rotatebox{90}{\whitetxt{xxx}Residential Building}
        & \includegraphics[height=3.7cm, width=3.7cm]{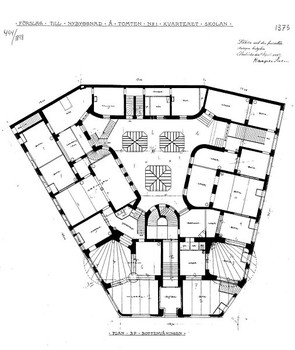} & 
        \includegraphics[height=3.7cm, width=3.7cm]{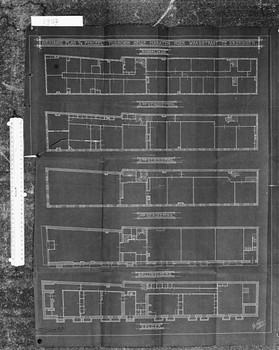} &
        \rotatebox{90}{ \whitetxt{xxxxxxx} School} &
        \includegraphics[height=3.7cm, width=3.7cm]{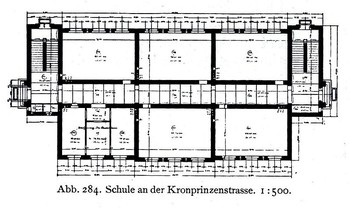} &
        \includegraphics[height=3.7cm, width=3.7cm]{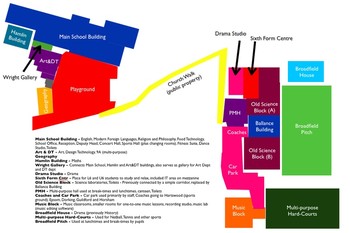} \\
        \rotatebox{90}{\whitetxt{xxxxxx} Museum}
        & \includegraphics[height=3.7cm, width=3.7cm]{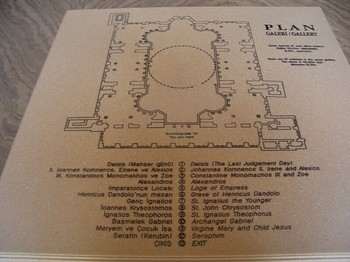} & 
        \includegraphics[height=3.7cm, width=3.7cm]{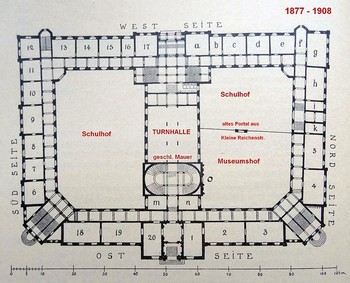} &
        \rotatebox{90}{ \whitetxt{xxxxxxx} House} &
        \includegraphics[height=3.7cm, width=3.7cm]{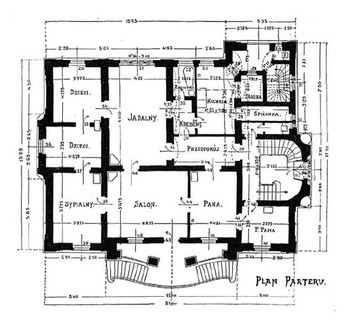} &
        \includegraphics[height=3.7cm, width=3.7cm]{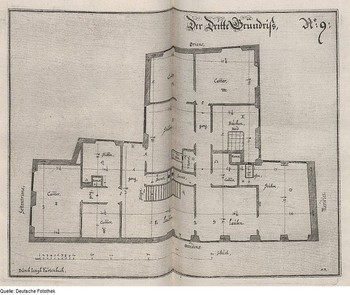}
        \\
  \end{tabular}   
\caption{Examples of images from our dataset with their building types (shown on the left)}
\label{fig:train_examples}
\end{figure*}

\section{Additional Results and Visualizations}

\subsection{\modified{ Semantic Segmentation Results}}

\modified{Figures \ref{fig:benchmark_segmentation} and \ref{fig:supp_benchmark_segmentation} contain examples of test images and annotations from our benchmark for semantic segmentation, and the results of the existing CC5K \cite{kalervo2019cubicasa5k} model on them. These figures demonstrate how the model is challenged in detecting segments it wasn’t exposed to during training, like pillars or curved walls. In Figure \ref{fig:wall_detection} we show qualitative examples of wall detection using different model architectures -- the existing ResNet-152 \cite{he2016deep} based model, and the Diffusion-based model discussed in Section \ref{sec:wall_detection}. As illustrated in the figure, using a more advanced model architecture allows for obtaining significantly cleaner wall segmentations. Table \ref{tab:wall_detection} contains a quantitative analysis of the two models on the wall segmentation prediction task.}

\subsection{Additional Open-Vocabulary Floorplan Segmentation  Results}
In Figure \ref{fig:segmentation_examples_supp}, we show additional examples of text-driven floorplan image segmentation before and after fine-tuning on our data. We see that the baseline model struggles to localize concepts inside floorplan images while our fine-tuning better concentrates probabilities in relevant regions, approaching the GT regions indicated in orange rectangles.

In Figure \ref{fig:segmentation_vs_cc5k} we visually compare the segmentation results to those of CC5K and CLIPSeg on residential buildings. We observe that the supervised CC5K model (trained on Finnish residential floorplans alone) fails to generalize to the diverse image appearances and building styles in \datasetname{}, even when they are residential buildings, while our model shows a more general understanding of semantics in such images.

\subsection{Additional Generation Results}

We show additional results for the generation task in Figure \ref{fig:generated_examples_sup} and for the spatially-conditioned generation in Figure \ref{fig:controlnet_ex_sup}. We provide multiple examples for various building types, showing that a model trained on our data learns the distinctive structural details of each building type. For example, castles have towers, libraries have long aisles, museums, hospitals, and hotels have numerous small rooms, churches have a typical cross shape, and theaters are characterized by having rows of seats facing a stage. The differences between the various types and their unique details are further shown in Figure \ref{fig:train_examples}, where we illustrate examples from our training set of various types.

In Table \ref{tab:metrics_full} we show a breakdown of the metrics for the generated images according to the most common building types in the dataset. The table compares our fine-tuned model with a base SD model, showing that for the vast majority of building types, our fine-tuned model generates images that are both more realistic and also semantically closer to the target prompt.

For structure-conditioned generation, we show additional results in Figure \ref{fig:struct_ex_sup}, where input conditions are derived from the CC5K dataset annotations as described above. In the figure, we show the effect of changing the condition and CFG scales during inference, illustrating the significance of these settings. In particular, we see that the condition scale controls the trade-off between fidelity to the layout condition and matching the building type in the prompt (rather than exclusively outputting images in the style of the CC5K fine-tuning data).

\renewcommand\cellset{\linespread{0.6}\selectfont}

\begin{figure*} {
\centering
\begin{tabular}{|p{3.88cm}|p{3.88cm}|p{3.88cm}|p{3.88cm}|}
\toprule
Image Category & Building Name & Building Type & Location Information \\
\midrule
\makecell[lt]{
\texttt{\ssmall[INST]}\\
 \\
\texttt{\ssmall
Please read the following}\\
\texttt{\ssmall (truncated) information} \\
\texttt{\ssmall extracted from Wikipedia}\\
\texttt{\ssmall related to an image:}\\
 \\
\texttt{\ssmall--- START WIKI INFO ---}\\
  \\
\texttt{\ssmall* Entity category:}\\
\whitetxt{xi}\texttt{\ssmall \textit{\{category\}}}\\
\texttt{\ssmall* Entity description:}\\
\whitetxt{xi}\texttt{\ssmall\textit{\{description\}}}\\
\texttt{\ssmall* Image filename:}\\
\whitetxt{xi}\texttt{\ssmall\textit{\{fn\}}}\\
\texttt{\ssmall* Texts that appear in the}\\
\whitetxt{xi}\texttt{\ssmall (image extracted with OCR)}\\
\whitetxt{xi}\texttt{\ssmall\textit{\{ocr\_texts\}}}\\
  \\
\texttt{\ssmall--- END WIKI INFO ---}\\
 \\
\texttt{\ssmall Now answer the following}\\
\texttt{\ssmall question in English: What is}\\
\texttt{\ssmall this file most likely a}\\
\texttt{\ssmall depiction of?}\\
\texttt{\ssmall(A) A floorplan}\\
\texttt{\ssmall(B) A building}\\
\texttt{\ssmall(C) A cross section of a}\\
\texttt{\ssmall building}\\
\texttt{\ssmall(D) A garden/park}\\
\texttt{\ssmall(E) A Map}\\
\texttt{\ssmall(F) A city/town}\\
\texttt{\ssmall(G) A physics/mathematics topic}\\
\texttt{\ssmall(H) I don't know}\\
 \\
\texttt{\ssmall Please choose one answer}\\
\texttt{\ssmall(A/B/C/D/E/F/G/H)}\\
 \\
\texttt{\ssmall[/INST]}\\
 \\
\texttt{\ssmall \textbf{The best answer is ( }}
}
&
\makecell[lt]{
\texttt{\ssmall[INST]}\\
 \\
\texttt{\ssmall
Please read the following}\\
\texttt{\ssmall (truncated) information} \\
\texttt{\ssmall extracted from Wikipedia}\\
\texttt{\ssmall related to an image of a}\\
\texttt{\ssmall building:}\\
 \\
\texttt{\ssmall--- START WIKI INFO ---}\\
  \\
\texttt{\ssmall* Entity category:}\\
\whitetxt{xi}\texttt{\ssmall\textit{\{category\}}}\\
\texttt{\ssmall* Entity description:}\\
\whitetxt{xi}\texttt{\ssmall\textit{\{description\}}}\\
\texttt{\ssmall* Image filename:}\\
\whitetxt{xi}\texttt{\ssmall\textit{\{fn\}}}\\
\texttt{\ssmall* Wiki page summary:}\\
\whitetxt{xi}\texttt{\ssmall\textit{\{wiki\_shows\}}}\\
\texttt{\ssmall* Texts that appear in the}\\
\whitetxt{xi}\texttt{\ssmall image (extracted with OCR):}\\
\whitetxt{xi}\texttt{\ssmall\textit{\{ocr\_texts\}}}\\
  \\
\texttt{\ssmall--- END WIKI INFO ---}\\
 \\
\texttt{\ssmall What is the name of the}\\
\texttt{\ssmall building depicted above?}\\
\texttt{\ssmall Write it in English,}\\
\texttt{\ssmall surrounded by brackets < >}\\
 \\
\texttt{\ssmall[/INST]}\\
 \\
\texttt{\ssmall \textbf{The name of the building}}\\
\texttt{\ssmall \textbf{discussed by the article is <}}\\
}

& 

\makecell[lt]{
\texttt{\ssmall[INST]}\\
 \\
\texttt{\ssmall
Please read the following}\\
\texttt{\ssmall (truncated) information} \\
\texttt{\ssmall extracted from Wikipedia}\\
\texttt{\ssmall related to an image of the}\\
\texttt{\ssmall building \{\textcolor{pink}{\textit{building\_name}}\}:}\\
 \\
\texttt{\ssmall--- START WIKI INFO ---}\\
  \\
\texttt{\ssmall* Entity category:}\\
\whitetxt{xi}\texttt{\ssmall\textit{\{category\}}}\\
\texttt{\ssmall* Entity description:}\\
\whitetxt{xi}\texttt{\ssmall\textit{\{description\}}}\\
\texttt{\ssmall* Image filename:}\\
\whitetxt{xi}\texttt{\ssmall\textit{\{fn\}}}\\
\texttt{\ssmall* Wiki page summary:}\\
\whitetxt{xi}\texttt{\ssmall\textit{\{wiki\_shows\}}}\\
 \\
\texttt{\ssmall--- END WIKI INFO ---}\\
 \\
\texttt{\ssmall What type or category of}\\
\texttt{\ssmall building is \textcolor{pink}{\textit{building\_name}}\}?}\\
\texttt{\ssmall Write your answer in}\\
\texttt{\ssmall English, surrounded by}\\
\texttt{\ssmall brackets < >}\\
 \\
\texttt{\ssmall[/INST]}\\
 \\
\texttt{\ssmall \textbf{The building \{\textcolor{pink}{\textit{building\_name}}\}}}\\
\texttt{\ssmall \textbf{is a <}}\\
}
 
&

\makecell[lt]{
\texttt{\ssmall[INST]}\\
 \\
\texttt{\ssmall
Please read the following}\\
\texttt{\ssmall (truncated) information} \\
\texttt{\ssmall extracted from Wikipedia}\\
\texttt{\ssmall related to an image of the}\\
\texttt{\ssmall building \{\textcolor{pink}{\textit{building\_name}}\}:}\\
 \\
\texttt{\ssmall--- START WIKI INFO ---}\\
  \\
\texttt{\ssmall* Entity category:}\\
\whitetxt{xi}\texttt{\ssmall\textit{\{category\}}}\\
\texttt{\ssmall* Entity description:}\\
\whitetxt{xi}\texttt{\ssmall\textit{\{description\}}}\\
\texttt{\ssmall* Image filename:}\\
\whitetxt{xi}\texttt{\ssmall\textit{\{fn\}}}\\
\texttt{\ssmall* Wiki page summary:}\\
\whitetxt{xi}\texttt{\ssmall\textit{\{wiki\_shows\}}}\\
 \\
\texttt{\ssmall--- END WIKI INFO ---}\\
 \\
\texttt{\ssmall Where is \{\textcolor{pink}{\textit{building\_name}}\}}\\
\texttt{\ssmall located? Write the country,}\\
\texttt{\ssmall state (if exists) and city}\\
\texttt{\ssmall surrounded by brackets < >}\\
\texttt{\ssmall and separate between them}\\
\texttt{\ssmall with a semi colon, for example:}\\
\texttt{\ssmall <City; State; Country>.}\\
\texttt{\ssmall If one of them is unknown}\\
\texttt{\ssmall write 'Unknown', for example:}\\
\texttt{\ssmall <City; Unknown; Country>,}\\
\texttt{\ssmall <Unknown; State; Country>}\\
\texttt{\ssmall etc.}\\
 \\
\texttt{\ssmall[/INST]}\\
 \\
\texttt{\ssmall \textbf{\{\textcolor{pink}{\textit{building\_name}}\} is located in <}}\\
} \\
\bottomrule
\end{tabular}} 
\caption{The prompts used for LLM-based extraction of pGTs. Each \{...\} placeholder is replaced with the respective image data. At first we only have raw data (as seen in the ``Image Category'' prompt), but once we gather pGTs we may use them in other prompts, for example \{\textcolor{pink}{\textit{building\_name}}\} as used in ``Building Type'' and ``Location Information''. We ask the LLM to return a semi-structured response (choosing an answer from a closed set; wrapping the answer in brackets etc.) so that we can easily extract the answer of interest. From left to right: The ``Image Category'' prompt is used for the initial text based filtering, where categories (A) and (B) are positive and the rest are negative. The ``Building Name'' and ``Building Type'' prompts are used for setting the building name and type respectively. The ``Location Information'' prompt extracts the country, state, and/or city (whichever of these exist). Note that the country is subsequently used for defining our test-train split.}
 \label{tab:llm_prompts}  
\end{figure*}

\renewcommand\cellset{\linespread{0.6}\selectfont}

\begin{figure*} {
\centering
\begin{tabular}{|p{3.88cm}|p{3.88cm}|p{3.88cm}|p{3.88cm}|}
\toprule
\ssmall Legend Existence (Wikipedia) & \ssmall Legend Content (Wikipedia) & \ssmall  Legend Existence (caption) & \ssmall Legend Content (caption) \\
\midrule
\makecell[lt]{
\texttt{\ssmall[INST]}\\
 \\
\texttt{\ssmall
The image "\{fn\}" is a plan of}\\
\texttt{\ssmall the building \{\textcolor{pink}{\textit{building\_name}}\} and}\\
\texttt{\ssmall it contains the following texts,}\\
\texttt{\ssmall detected by an OCR model:}\\
 \\
 \texttt{\ssmall\textit{\{ocr\_texts\}}}\\
  \\
\texttt{\ssmall Please read the following}\\
\texttt{\ssmall excerpt from an article about}\\
\texttt{\ssmall the building which contains}\\
\texttt{\ssmall this image:}\\
 \\
\texttt{\ssmall--- START EXCERPT ---}\\
  \\
\texttt{\ssmall\textit{\{snippet\}}}\\
  \\
\texttt{\ssmall--- END EXCERPT ---}\\
 \\
\texttt{\ssmall Now answer the following }\\
\texttt{\ssmall question about the excerpt:}\\
\texttt{\ssmall Does the excerpt contain a}\\
\texttt{\ssmall legend for the image "\{fn\}",}\\
\texttt{\ssmall i.e. an itemized list}\\
\texttt{\ssmall corresponding to regions marked}\\
\texttt{\ssmall by OCR labels in the image,}\\
\texttt{\ssmall explaining what each label}\\
\texttt{\ssmall signifies? Answer}\\
\texttt{\ssmall yes/no/unsure.}\\
 \\
\texttt{\ssmall[/INST]}\\
 \\
\texttt{\ssmall \textbf{The answer to the question is:}}}
&
\makecell[lt]{
\texttt{\ssmall[INST]}\\
  \\
\texttt{\ssmall
The image "\{fn\}" is a plan of}\\
\texttt{\ssmall the building \{\textcolor{pink}{\textit{building\_name}}\} and}\\
\texttt{\ssmall it contains the following texts,}\\
\texttt{\ssmall detected by an OCR model:}\\
 \\
 \texttt{\ssmall\textit{\{ocr\_texts\}}}\\
  \\
\texttt{\ssmall Please read the following}\\
\texttt{\ssmall excerpt from an article about}\\
\texttt{\ssmall the building which contains}\\
\texttt{\ssmall this image:}\\
 \\
\texttt{\ssmall--- START EXCERPT ---}\\
  \\
\texttt{\ssmall\textit{\{snippet\}}}\\
  \\
\texttt{\ssmall--- END EXCERPT ---}\\
 \\
\texttt{\ssmall The excerpt contains a legend,}\\
\texttt{\ssmall i.e. an itemized list}\\
\texttt{\ssmall corresponding to regions}\\
\texttt{\ssmall marked by labels in the image.}\\
\texttt{\ssmall Reproduce the legend below.}\\
 \\
\texttt{\ssmall[/INST]}\\
 \\
\texttt{\ssmall \textbf{Sure! Here is the legend: }}}
& 
\makecell[lt]{
\texttt{\ssmall[INST]}\\
 \\
\texttt{\ssmall
The image "\{fn\}" is a plan}\\
\texttt{\ssmall of the building \{\textcolor{pink}{\textit{building\_name}}\}} \\
\texttt{\ssmall and it has the following}\\
\texttt{\ssmall description:}\\
 \\
\texttt{\ssmall--- START IMAGE DESC. ---}\\
  \\
\texttt{\ssmall\textit{\{description\}}}\\
  \\
\texttt{\ssmall--- END IMAGE DESC. ---}\\
 \\
\texttt{\ssmall Does the description above look}\\
\texttt{\ssmall like it contains a legend for}\\
\texttt{\ssmall the image, i.e. an itemized}\\
\texttt{\ssmall list corresponding to regions}\\
\texttt{\ssmall marked by labels in the image,}\\
\texttt{\ssmall explaining what each label}\\
\texttt{\ssmall signifies?}\\
\texttt{\ssmall Write yes/no/not sure in}\\
\texttt{\ssmall English, surrounded by}\\
\texttt{\ssmall brackets < >}\\
 \\
\texttt{\ssmall[/INST]}\\
 \\
\texttt{\ssmall \textbf{<}}
}
&
\makecell[lt]{
\texttt{\ssmall[INST]}\\
 \\
\texttt{\ssmall
The image "\{fn\}" is a plan}\\
\texttt{\ssmall of the building \{\textcolor{pink}{\textit{building\_name}}\}} \\
\texttt{\ssmall and it has the following}\\
\texttt{\ssmall description:}\\
 \\
\texttt{\ssmall--- START IMAGE DESC. ---}\\
  \\
\texttt{\ssmall\textit{\{description\}}}\\
  \\
\texttt{\ssmall--- END IMAGE DESC. ---}\\
 \\
\texttt{\ssmall Does the discussed image}\\
\texttt{\ssmall contain a legend (as in a}\\
\texttt{\ssmall key/table/code for}\\
\texttt{\ssmall understanding the image)?}\\
\texttt{\ssmall If so, what are the legend's}\\
\texttt{\ssmall contents? Answer with a}\\
\texttt{\ssmall bulleted list in English of the}\\
\texttt{\ssmall legend contents. Include only}\\
\texttt{\ssmall full items and not just labels}\\
\texttt{\ssmall (for example, '1. nave' should}\\
\texttt{\ssmall be included, but '1.' alone}\\
\texttt{\ssmall shouldn't)}\\
 \\
\texttt{\ssmall[/INST]}\\
 \\
\texttt{\ssmall \textbf{Answer: The legend contains:
}}\\
\texttt{\ssmall \textbf{*
}}
} \\
\bottomrule
\end{tabular}} 
\vspace{-8pt}
\caption{The prompts used for extracting the legend contents. The two left prompts are used for extracting data from Wikipedia, and the two right ones for the image caption. In both cases this is a two-step extraction: first we ask the LLM if it thinks the text contains a legend. Only if it answers yes, we ask it for its content. This reduces hallucinations and keeps the answers accurate.}
 \label{tab:legend_prompts}  
\end{figure*}

\begin{figure} {
\centering
\begin{tabular}{|p{8cm}|}
\toprule
Legend simplification \\
\midrule
\makecell[lt]{
\texttt{\ssmall[INST]}\\
 \\
\texttt{\ssmall
The following texts contain a legend of a \{\textcolor{pink}{\textit{building\_type}}\} floor}\\
\texttt{\ssmall plan in a key:value format:}\\
 \\
\texttt{\ssmall--- START LEGEND ---}\\
  \\
\texttt{\ssmall\textit{\{\textcolor{pink}{legend}\}}}\\
  \\
\texttt{\ssmall--- END LEGEND ---}\\
 \\
\texttt{\ssmall Please rewrite the legend using simple and generic words.}\\
\\ 
\texttt{\ssmall Do:}\\
\texttt{\ssmall * Include all legend parts from the list above.}\\
\texttt{\ssmall * Keep it simple and short: summarize each row in one/two words}\\
\texttt{\ssmall * Keep the original legend keys}\\
\texttt{\ssmall * In case the features have distinct names (e.g. Chapel of the}\\
\texttt{\ssmall  Ascension) treat their type only (e.g. a chapel) and disregard}\\
\texttt{\ssmall any specific name.}\\
 \\
\texttt{\ssmall Don't:}\\
\texttt{\ssmall * Don't invent new information}\\
\texttt{\ssmall * Don't include specific names (use their type instead)}\\
\texttt{\ssmall * Don't skip any of the legend lines above}\\
 \\
 \texttt{\ssmall Write your answer in English, translating any non-English terms.}\\
 \\
\texttt{\ssmall[/INST]}\\
 \\
\texttt{\ssmall \textbf{Sure, here's a simplified and generalized version of the legend:}}\\
\texttt{\ssmall \textbf{* }}\\
} \\
\bottomrule
\end{tabular}} 
\caption{The prompt used for legend simplification, serving to clean up the original legends for the image grounding process. The goal is to obtain a list of keys to architectural features. We aim to shorten long descriptions, remove names, and translate any non-English text.}
 \label{tab:legend_simplification_prompt}  
\end{figure}

\renewcommand\cellset{\linespread{0.6}\selectfont}

\begin{figure*} {
\centering
\begin{tabular}{|p{3.88cm}|p{3.88cm}|p{3.88cm}|p{3.88cm}|}
\toprule
\ssmall Legend Existence & \ssmall Legend Content & \ssmall  Arc-Feats Existence & \ssmall Arc-Feats Content \\
\midrule
\makecell[lt]{
\texttt{\ssmall[INST]}\\
 \\
\texttt{\ssmall
The image "\{fn\}" is a plan}\\
\texttt{\ssmall of the building \{\textcolor{pink}{\textit{building\_name}}\}} \\
\texttt{\ssmall and it contains the following}\\
\texttt{\ssmall texts, detected by an OCR model:}\\
 \\
 \texttt{\ssmall--- START IMAGE TEXTS ---}\\
 \\
 \texttt{\ssmall\textit{\{ocr\_legend\_candidate\}}}\\
 \\
 \texttt{\ssmall--- END IMAGE TEXTS ---}\\
 \\
\texttt{\ssmall Do the OCR detections above }\\
\texttt{\ssmall look like they contain a legend}\\
\texttt{\ssmall for the image, i.e. an}\\
\texttt{\ssmall itemized list corresponding to}\\
\texttt{\ssmall regions marked by OCR labels in}\\
\texttt{\ssmall the image, explaining what each}\\
\texttt{\ssmall label signifies?}\\\\
\texttt{\ssmall Write yes/no/not sure in}\\
\texttt{\ssmall English, surrounded by}\\
\texttt{\ssmall brackets < >}\\
 \\
\texttt{\ssmall[/INST]}\\
 \\
\texttt{\ssmall \textbf{<}}}
&
\makecell[lt]{
\texttt{\ssmall[INST]}\\
 \\
\texttt{\ssmall
The image "\{fn\}" is a plan}\\
\texttt{\ssmall of the building \{\textcolor{pink}{\textit{building\_name}}\}} \\
\texttt{\ssmall and it contains the following}\\
\texttt{\ssmall texts, detected by an OCR model:}\\
 \\
 \texttt{\ssmall--- START IMAGE TEXTS ---}\\
 \\
 \texttt{\ssmall\textit{\{ocr\_legend\_candidate\}}}\\
 \\
 \texttt{\ssmall--- END IMAGE TEXTS ---}\\
 \\
\texttt{\ssmall The above texts may contain,}\\
\texttt{\ssmall among other things, the content}\\
\texttt{\ssmall  of an image legend (as in a}\\
\texttt{\ssmall key/table/code for}\\
\texttt{\ssmall understanding the image).}\\
\texttt{\ssmall Can you extract the legend}\\
\texttt{\ssmall contents from the above texts?}\\
\texttt{\ssmall Answer with a bulleted list of}\\
\texttt{\ssmall the legend contents.}\\
\texttt{\ssmall Include only full items and}\\
\texttt{\ssmall not just keys/labels (for}\\
\texttt{\ssmall example, '1. nave' can be}\\
\texttt{\ssmall included, but '1.' or 'nave'}\\
\texttt{\ssmall alone shouldn't).}\\
\texttt{\ssmall Disregard text that doesn't}\\
\texttt{\ssmall seem like it's part of the}\\
\texttt{\ssmall legend.}\\
\texttt{\ssmall Include the original}\\
\texttt{\ssmall keys/labels and don't invent}\\
\texttt{\ssmall new ones. If you can't deduce}\\
\texttt{\ssmall a legend return "I don't know".}\\
 \\
\texttt{\ssmall[/INST]}\\
 \\
\texttt{\ssmall \textbf{Sure! Here are the legend}}\\
\texttt{\ssmall \textbf{contents: }}}
& 
\makecell[lt]{
\texttt{\ssmall[INST]}\\
 \\
\texttt{\ssmall
The image "\{fn\}" is a plan}\\
\texttt{\ssmall of the building \{\textcolor{pink}{\textit{building\_name}}\}} \\
\texttt{\ssmall and it contains the following}\\
\texttt{\ssmall texts, detected by an OCR model:}\\
 \\
 \texttt{\ssmall--- START IMAGE TEXTS ---}\\
 \\
 \texttt{\ssmall\textit{\{ocr\_legend\_candidate\}}}\\
 \\
 \texttt{\ssmall--- END IMAGE TEXTS ---}\\
 \\
\texttt{\ssmall Do the OCR detections above}\\
\texttt{\ssmall look like they contain words}\\
\texttt{\ssmall that represent architectural}\\
\texttt{\ssmall feature labels?}\\
\texttt{\ssmall Disregard anything that looks}\\
\texttt{\ssmall like a symbol or a key (like}\\
\texttt{\ssmall numbers), and any words that}\\
\texttt{\ssmall represent direction (e.g. }\\
\texttt{\ssmall north, east, etc.)}\\
\texttt{\ssmall Write yes/no/not sure in}\\
\texttt{\ssmall English, surrounded by}\\
\texttt{\ssmall brackets < >}\\
 \\
\texttt{\ssmall[/INST]}\\
 \\
\texttt{\ssmall \textbf{<}}}
&
\makecell[lt]{
\texttt{\ssmall[INST]}\\
 \\
\texttt{\ssmall
The image "\{fn\}" is a plan}\\
\texttt{\ssmall of the building \{\textcolor{pink}{\textit{building\_name}}\}} \\
\texttt{\ssmall and it contains the following}\\
\texttt{\ssmall texts, detected by an OCR model:}\\
 \\
 \texttt{\ssmall--- START IMAGE TEXTS ---}\\
 \\
 \texttt{\ssmall\textit{\{ocr\_legend\_candidate\}}}\\
 \\
 \texttt{\ssmall--- END IMAGE TEXTS ---}\\
 \\
\texttt{\ssmall The above texts may contain,}\\
\texttt{\ssmall among other things,}\\
\texttt{\ssmall architectural features marked}\\
\texttt{\ssmall on the floorplan.}\\
\texttt{\ssmall Out of the texts above, can you}\\
\texttt{\ssmall extract those that represent}\\
\texttt{\ssmall architectural features?}\\
\texttt{\ssmall Like room types, halls, porches,}\\
\texttt{\ssmall etc.}\\
\texttt{\ssmall Don't include anything that}\\
\texttt{\ssmall looks like a symbol or a key}\\
\texttt{\ssmall (like numbers). Don't include}\\
\texttt{\ssmall any words that represent}\\
\texttt{\ssmall direction (e.g. north, east,}\\
\texttt{\ssmall etc.). }\\
\texttt{\ssmall Disregard text that isn't}\\
\texttt{\ssmall related to architectural}\\
\texttt{\ssmall features.}\\
\texttt{\ssmall Answer with a bulleted list}\\
\texttt{\ssmall of the architectural features.}\\
\texttt{\ssmall Use the original text, do not}\\
\texttt{\ssmall modify, translate, or add}\\
\texttt{\ssmall extensions to the text you}\\
\texttt{\ssmall chose to add.}\\
\texttt{\ssmall If you can't deduce any}\\
\texttt{\ssmall architectural features return}\\
\texttt{\ssmall "I don't know".}\\
 \\
\texttt{\ssmall[/INST]}\\
 \\
\texttt{\ssmall \textbf{Sure! Here are the}}\\
\texttt{\ssmall \textbf{architectural features that}}\\
\texttt{\ssmall \textbf{appear in the texts you}}\\
\texttt{\ssmall \textbf{provided:}}} \\
\bottomrule
\end{tabular}} 
\vspace{-8pt}
\caption{The prompts used for extracting legends and architectural features from OCR detections. The two left prompts are used for extracting legends, and the two right ones for architectural feature labels. In both cases this is a two-step extraction: first we ask the LLM if it thinks the text contains a legend. Only if it answers yes, we ask it for its content. This reduces hallucinations and keeps the answers accurate.}
 \label{tab:legend_arc_ocr_prompts}  
\end{figure*}

\clearpage

\end{document}